\def\year{2022}\relax
\documentclass[letterpaper]{article} 
\usepackage{aaai22}  
\usepackage{times}  
\usepackage{helvet}  
\usepackage{courier}  
\usepackage[hyphens]{url}  
\usepackage{graphicx} 
\usepackage{natbib}  
\usepackage{caption} 
\DeclareCaptionStyle{ruled}{labelfont=normalfont,labelsep=colon,strut=off} 
\usepackage{algorithm}
\usepackage{algorithmic}

\usepackage{newfloat}
\usepackage{listings}
\floatstyle{ruled}
\newfloat{listing}{tb}{lst}{}
\floatname{listing}{Listing}

\author{
    Hao Chen\textsuperscript{\rm 1}, Feihong Shen
}
\affiliations{
    \textsuperscript{\rm 1}School of Computer Science, Southeast University, China \\
    \{haochen303, feihongshen\}@seu.edu.cn

}

\usepackage{color}
\usepackage{booktabs}
\usepackage{multirow}
\usepackage{makecell}
\usepackage{bm}
\usepackage{subfig}
\usepackage{bm}

\title{Hierarchical Cross-modal Transformer for RGB-D Salient Object Detection}

\usepackage{bibentry}

\begin{document}

\maketitle

\begin{abstract}
Most of existing RGB-D salient object detection (SOD) methods follow the CNN-based paradigm, which is unable to model long-range dependencies across space and modalities due to the natural locality of CNNs. Here we propose the Hierarchical Cross-modal Transformer (HCT), a new multi-modal transformer, to tackle this problem. Unlike previous multi-modal transformers that directly connecting all patches from two modalities, we explore the cross-modal complementarity hierarchically to respect the modality gap and spatial discrepancy in unaligned regions. Specifically, we propose to use intra-modal self-attention to explore complementary global contexts, and measure spatial-aligned inter-modal attention locally to capture cross-modal correlations. In addition, we present a Feature Pyramid module for Transformer (FPT) to boost informative cross-scale integration as well as a consistency-complementarity module to disentangle the multi-modal integration path and improve the fusion adaptivity. Comprehensive experiments on a large variety of public datasets verify the efficacy of our designs and the consistent improvement over state-of-the-art models.  

\end{abstract}

\section{Introduction}

Salient object detection (SOD), which aims to simulate human visual systems to identify the most attractive objects in a scene, has benefited a large variety of computer vision tasks such as object detection \cite{jiao2019survey}, image recognition \cite{sukanya2016survey} and tracking \cite{ciaparrone2020deep}.

\captionsetup[subfigure]{textfont=scriptsize}
\begin{figure}[ht]
	\centering
	\subfloat[Global cross-attention in VST]{\includegraphics[width = 0.15\textwidth]{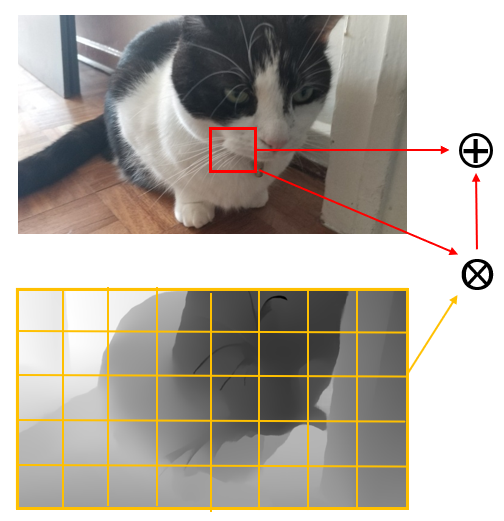}}
	\hspace{1mm}
	\subfloat[Global self-attention]{\includegraphics[width = 0.15\textwidth]{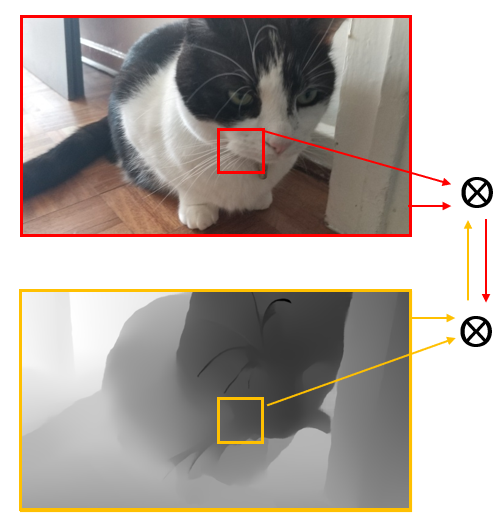}}
	\hspace{1mm}
	\subfloat[Local-aligned cross-attention.]{\includegraphics[width = 0.15\textwidth]{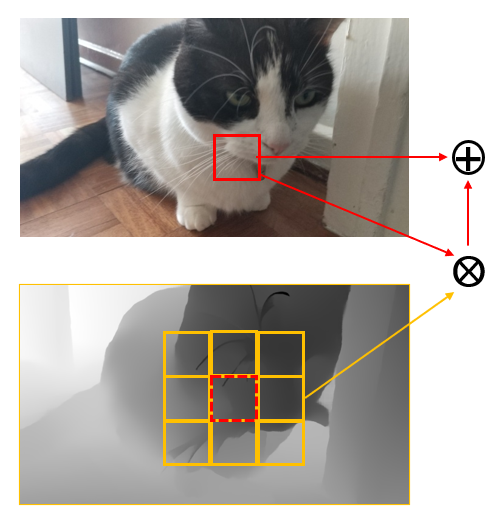}}
	
	\caption{Examples to show the advantages of our proposed cross-modal attention scheme, formed by (b) and (c).}
	\label{fig:similarity}
\end{figure}

Recently, RGB-D SOD has attracted increasing attention for the additional spatial structure cues from depth to complement the RGB inference on challenging cases such as when the background and the foreground hold similar appearance. Most of existing RGB-D SOD methods \cite{cheng2014depth,cong2016saliency,shigematsu2017learning,sun2021deep,zhao2022self} follow the CNN-based paradigm. With powerful CNN backbones \cite{simonyan2014very,he2016deep} to extract feature hierarchies, their most efforts focus on designing various cross-modal cross-level interaction and fusion paths \cite{chen2019multi,zhao2019contrast,sun2021deep} to explore the heterogeneous feature complementarity, and diverse strategies such as attention modules \cite{zhou2018semi,chen2019three}, dynamic convolution \cite{chen2020dynamic}, feature disentanglement \cite{9165931} and knowledge distillation \cite{HaoChen2021CNNBasedRS} to boost the adaptivity in selecting complementary cues. 
These methods, although greatly advance the RGB-D SOD community, hold an intrinsic limitation in capturing global contexts as the natural locality of convolutions. However, it has been widely acknowledged in \cite{goferman2011context, zhao2015saliency, liu2018picanet} that global contexts are dominated to correctly localize the SOD. Even some strategies try to enhance the global understanding by appending fully connected \cite{NianLiu2016DHSNetDH} or global pooling layers \cite{NianLiu2017PiCANetLP} on restricted layers, they still struggle with large computational cost or limited capability in modelling global correlations.

\par Recently, Transformer \cite{vaswani2017attention}, which experts in capturing long-range dependencies, overcomes the limitation in CNNs, thus carrying great potential in modelling complex cross-modal complementarity and studying global contexts to infer the SOD. Given this, the VST \cite{liu2021visual} model, using the transformer as the backbone, has been proposed for RGB(D) SOD. Specifically, VST adopts T2T ViT \cite{t2tvit} as the encoder and the cross-modal fusion problem is solved by a cross-modal attention module, formed by the similarity between a query from one modality and all the keys from the paired modality. Compared to the CNN-based counterpart, VST catches long range dependencies within/cross modalities, thus achieves state-of-the-art results on RGB-D SOD.

\begin{figure*}
	\centering
	\includegraphics[width=0.87\linewidth]{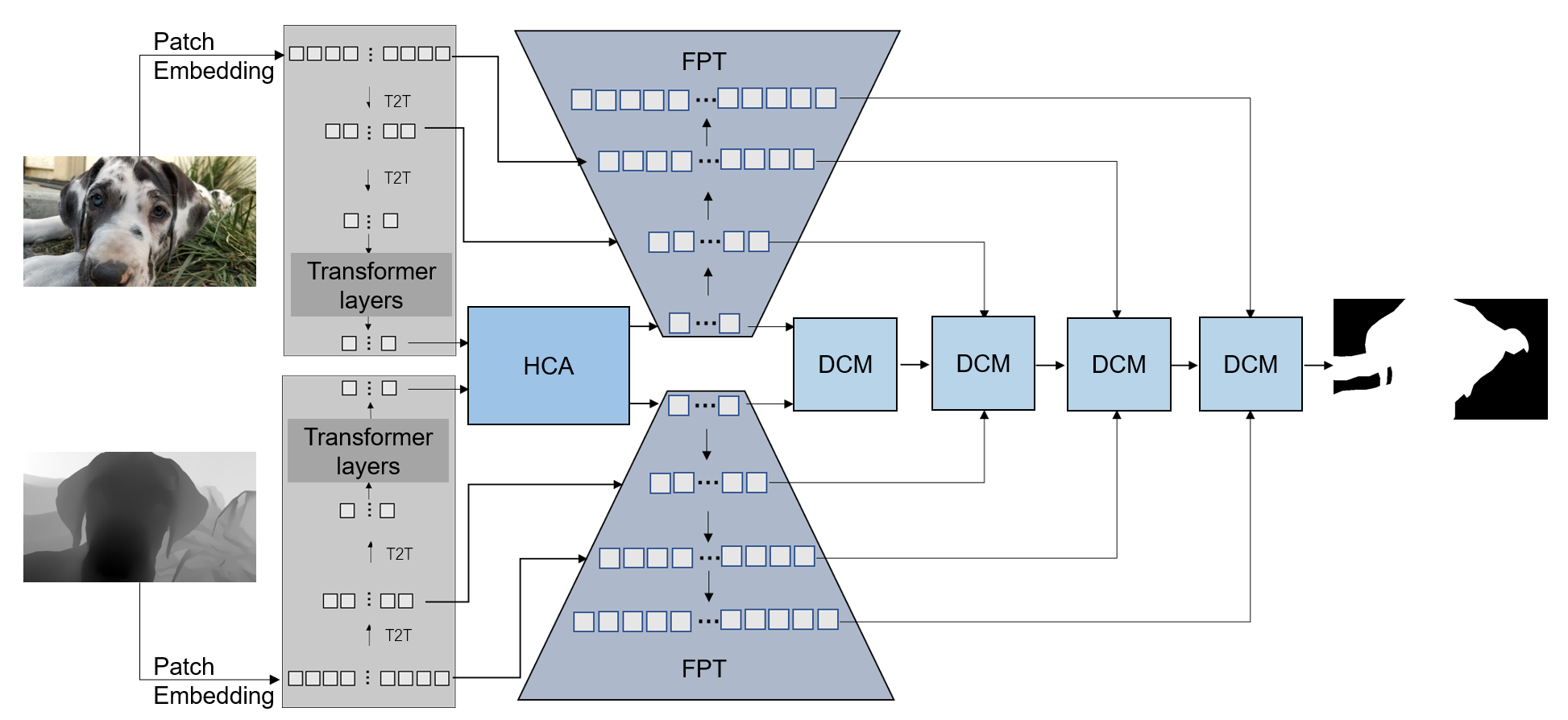}
	\caption{Overall architecture of HCT. 
 }
	\label{fig:model}
\end{figure*}

\par However, two key questions in selecting cross-modal and cross-level complements are still remaining. 
\par 1. How to combine cross-modal cues is the key to multi-modal understanding. VST answers this question by involving the depth patches to form a multi-modal token pool to model the cross-modal long-range dependencies. As shown in Fig. \ref{fig:similarity} (a), each patch in RGB will be compared to all regions in the paired depth image  and vice verse to generate cross-modal dependencies. However, measuring the dependencies between cross-modal tokens that lie distant in space (e.g., the cat in RGB and the floor in depth) makes little sense due to the severe cross-modal representation gap. Hence, this strategy is difficult to model the cross-modal global contexts and the improvement is quite limited. Additionally, such global cross-modal fusion overlooks the inborn spatial-alignment in RGB-D pairs and the similarity between cross-modal spatial-distant patches will introduce noise to the query from the keys in other areas. Hence, we argue that the modality gap and spatial discrepancy should not concurrent when measuring cross-modal dependencies for complementing. Based on this insight, we propose a \textbf{H}ierarchical \textbf{C}ross-modal \textbf{T}ransformer (HCT), which customizes the cross-modal complementarity from two views: I) Intra-modal global contexts. As illustrated in Fig. \ref{fig:similarity} (b), The within-modal self-attention reveals global contrasts and contexts in each modality. Combining this global understanding from two modalities will facilitate the discrimination between foreground from background a lot. II) Cross-modal local spatial-aligned enhancement. As a local region aligned in two modalities carry the same local context (Fig. \ref{fig:similarity} (c)), the cross-modal attention restricted in the same small aligned region can well capture the cross-modal correlation and mapping, thereby bridging the cross-modal gap and easing multi-modal feature fusion. With the complements from above two perspectives, our HCT can well exploit the cross-modal global/local contexts for complementary global reasoning and local enhancement. On top of this, we draw inspiration from \cite{zhao2022self} to disentangle heterogeneous multi-modal complements and boost the fusion by explicitly constraining the consistency and complementarity between two transformer streams.

\par 2. As different transformer layers characterize an object with varying scales, directly concatenating such heterogenous cross-level representations without explicit selecting and adaptive weighting will result in deficient feature integration. \cite{li2022exploring} finds deep semantic features from a transformer-based encoder dominate the dense prediction tasks. However, the VST model treats all features from different scales equally during the fusion process. To tackle this issue, we design a Feature Pyramid for Transformer (FPT) to adaptively propagate the high-level semantics to progressively guide the selection and integration of shallow features.
\par By solving the above two problems, our HCT enjoys long-range dependencies, adaptive cross-scale integration, hierarchical cross-modal contextual complements, fine-grained cross-modal interactions, and explicit disentanglement of complex cross-modal relationships. Extensive experiments verify the efficacy of our designs and the large improvement over state-of-the-art on 6 benchmark datasets.

\begin{figure*}[ht]
	\centering
	\subfloat[Hierarchical cross-modal attention]{\includegraphics[width = 0.315\textwidth]{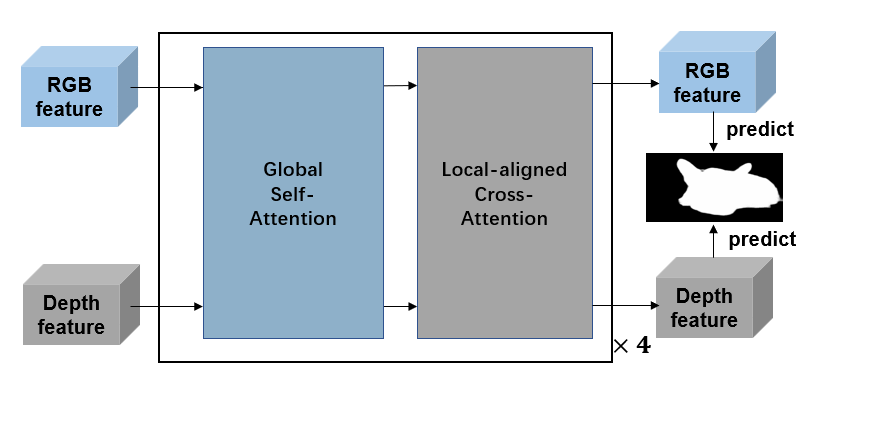}}
        \hspace{1.5mm}
	\subfloat[Global self-attention]{\includegraphics[width = 0.315\textwidth]{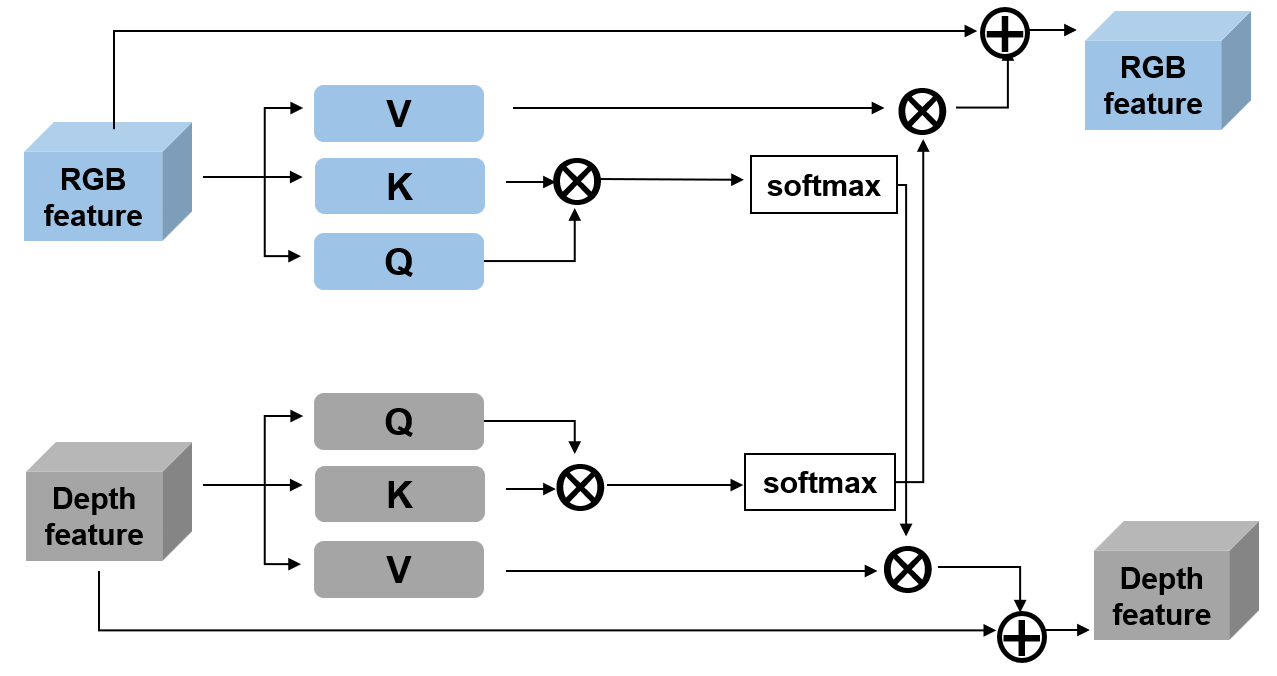}}
	\hspace{1.5mm}
	\subfloat[Local-aligned cross-attention]{\includegraphics[width = 0.315\textwidth]{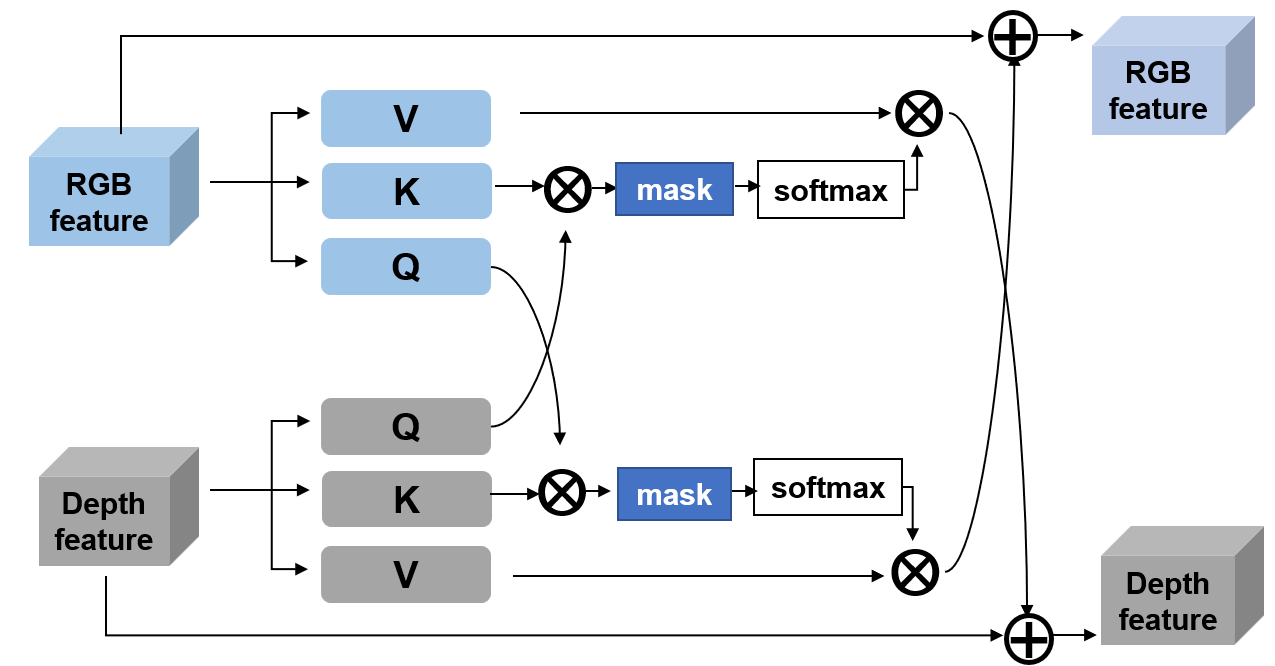}}
	
	\caption{Architecture the hierarchical cross-modal attention module and its components: global self-attention and local-aligned cross-attention.}
	\label{fig:crossattention}
\end{figure*}

\section{Related Work}
\subsection{RGB-D Salient Object Detection}

As handcrafted saliency cues \cite{ciptadi2013depth,peng2014rgbd,cong2016saliency} are weak in learning global contexts and hold limited generalization ability, recent RGB-D SOD models mainly focus on designing CNN architectures and cross-modal fusion paths to better study the complements. 
For example, \citet{qu2017rgbd} integrate handcrafted cues from two modalities as a joint input to train a shared CNN. PCF \cite{chen2018progressively} proposes a progressive multi-scale fusion strategy and TANet \cite{chen2019three} introduces a three-stream architecture to explicitly select complementary cues in each level. 

Apart from the basic fusion architectures (i.e., single stream, two-stream, and three-stream), some other works introduce various feature combination strategies \cite{JunweiHan2018CNNsBasedRS}, cross-modal cross-level interaction paths \cite{NianLiu2020LearningSS}, and other strategies such as knowledge distillation \cite{HaoChen2021CNNBasedRS} and dynamic convolution \cite{pang2020hierarchical} to boost the fusion sufficiency.

In summary, the prior CNN-based RGB-D SOD community has achieved noticable advances. Nonetheless, the locality nature of CNNs make them weak in learning global contexts. Different from previous remediation strategies such as using global pooling \cite{NianLiu2017PiCANetLP} or fully connected layers \cite{NianLiu2016DHSNetDH}, \citet{liu2021visual} eschew this intrinsic limitation by designing a transformer-based architecture to extract intra/inter-modal long-rang dependencies. They use ViTs as encoders for each modality and stitch the tokens from two modalities for cross-modal fusion. Benefit from the long-range self-attention mechanism, VST achieves large improvement over previous CNN-based methods. Whereas, it ignores the large cross-modal gap. Consequently, directly computing the cross-modal dependency between distant regions contributes little contexts and even tends to introduce noises. Also, simply concatenating features in the decoder without differentiating neglects the varying contributions of different layers in inference.
Differently, we customize a hierarchical cross-modal fusion module to carefully model the cross-modal transformer complements from two views, as well as a feature pyramid module to enable adaptive cross-level integration.

\subsection{Transformer in Computer Vision}
The transformer \cite{vaswani2017attention} is first proposed for machine translation and achieves impressive results in various natural language processing tasks with its powerful ability in modeling global contexts. 
Inspired by its great potential, increasing vision transformers have been put forward in the computer vision community. \citet{dosovitskiy2020image} successfully introduce transformer (ViT) into image classification by splitting an image into patches. Since then, various transformers are introduced, e.g., T2T ViT \cite{t2tvit} aggregates adjacent tokens to model local structure and reduce the computational cost, PVT \cite{wang2021pyramid}  proposes spatial-reduction attention to learn multi-scale feature maps flexibly. Swin transformer \cite{liu2021swin} fully takes advantages of displacement invariance and size invariance and integrates them into an transformer. 
These transformers are widely applied in other computer vision tasks as backbones, including detection \cite{carion2020end}, segmentation \cite{kirillov2019panoptic}, video processing \cite{sun2019videobert}, etc.

Transformers are also widely used in multi-modal learning for its flexible input. For instance, \citet{sun2019videobert} straightly project the video and audio into patches and throws them into a shared transformer backbone. \citet{zheng2021fused} design a transformer structure to handle embeded acoustic input and text simultaneously. However, these methods concatenate embeddings straightly and achieve intra-modal and inter-modal combinations at the same time, leading to uninformative fusion and even introducing noisy features. The VST \cite{liu2021visual} model tackles this issue by extracting RGB and depth features individually and adopting a cross attention module for cross-modal fusion. However, the cross-modal attention module in VST introduces cross-modal gap and spatial discrepancy simultaneously, making it difficult to explore the heterogeneous cross-modal complements clearly. Hence, we decouple the complex cross-modal transformer complements and propose a hierarchical cross-modal attention to control the modality/spatial gaps and progressively incorporate cross-modal complements.

\section{The Proposed Method}
In this section, we first introduce the overall architecture of our model. And then, we will detail our key designs, including the hierarchical cross-modal attention module (HCA), the feature pyramids for transformer (FPT) and the disentangled complementing module (DCM).

\subsection{Overall Architecture}
Fig. \ref{fig:model} shows the architecture of our proposed model. Two T2T ViT backbones pretrained on ImageNet are applied to extract RGB and depth features, respectively. And then the HCA module is tailored to explicitly take advantage of the cross-modal spatial alignment to boost the global/local contextual fusion. Next, the feature pyramid is constructed to selectively assimilate shallower layers with the guidance of the deep semantics. Finally, the DCM is introduced to bifurcate the heterogeneous complementary cues to modal-shared and modal-specific ones to boost the fusion sufficiency and their combination is used to generate the final joint prediction.

\subsection{Hierarchical Cross-Modal Attention}
\begin{figure}
	\centering
	\includegraphics[width=0.98\linewidth]{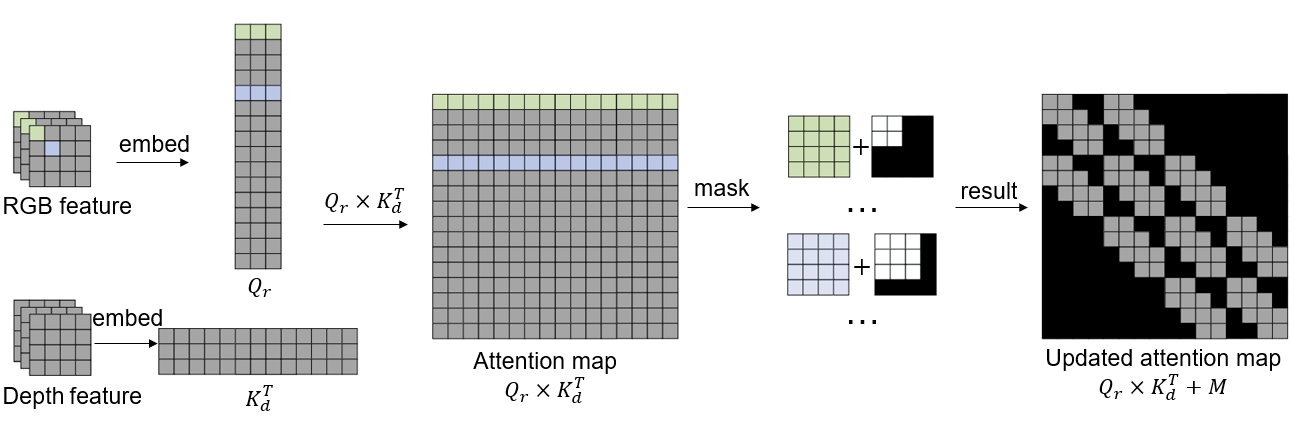}
	\caption{The masking process of local-aligned cross-attention in HCA. }
	\label{fig:mask}
\end{figure}
As shown in Fig. \ref{fig:crossattention} (a), our HCA consists of two stages: Global self-attention (GSA) and Local-aligned cross-attention (LCA) as illustrated in Fig. \ref{fig:crossattention} (b) and Fig. \ref{fig:crossattention} (c), respectively. In the GSA stage, RGB and depth features will calculate their intra-modal self-attention maps individually, and then swap their global self-attention maps to achieve complementary global contexts. The attention scores are calculated by the following formulas:
\begin{equation}\label{eq:complementarycross}
GSA(Q_r,K_r,V_d)=softmax(\frac{Q_rK_r^T}{\sqrt{d}}V_d),
\end{equation}
\begin{equation}\label{eq:complementarycross2}
GSA(Q_d,K_d,V_r)=softmax(\frac{Q_dK_d^T}{\sqrt{d}}V_r),
\end{equation}
where $Q_r$, $K_r$, $V_r$ denotes query, key and value generated from RGB features, and $Q_d$, $K_d$, $V_d$ come from depth features. Unlike self-attention, we swap the attention map $Q\times K$ which describes the within-modal self-relation globally. The $GSA(Q_d,K_d,V_r)$ and $GSA(Q_r,K_r,V_d)$ will finally add back to RGB and depth features respectively as a residual part.

After swapping the structural information, we adopt a local cross-attention module to reinforce the fusion of cross-modal local semantics. The main difference between LCA and traditional cross attention lies in a mask operation, which will keep attention similarity $Q\times K^T$  in adjacent area and set similarity to 0 in remote areas. Specifically, the global cross-attention map $Q\times K^T$ will add with a mask matrix $M$ where 0 denotes adjacent area and -100 represents remote area, after softmax the attention scores from remote area are close to 0, thus remote areas make no contribution to the further fusion process.
This process can be formulated as follows:
\begin{equation}\label{eq:cross}
LCA(Q_r,K_d,V_d)=softmax(\frac{Q_rK_d^T+M}{\sqrt{d}}V_d) , 
\end{equation}
\begin{equation}\label{eq:cross2}
LCA(Q_d,K_r,V_r)=softmax(\frac{Q_dK_r^T+M}{\sqrt{d}}V_r) . 
\end{equation}

Fig. \ref{fig:mask} details the process of mask generation. After embedding RGB and depth features into $Q$, $K$ and $V$ forms, we multiply $Q_r$ with $K_d^T$ to obtain a unified attention map. For example, the first line in attention map depicts the similarity between the first patch in RGB features and all the patches in the paired depth. Then, we design a mask for each line according to its distance to the target patch and get the flitted attention map. The masked attention map, carrying complementary local contexts and cross-modal correlation, is then multiplied with the $V_d$ to generate the final depth features. Note that a symmetrical interaction line is also performed to enhance RGB features. To further encourage the extraction and selection of cross-modal complementary cues,  updated features will be forced to predict the saliency maps by optimizing: 
\begin{equation}\label{eq:loss1}
loss1 = -[y\cdot \log \sigma(x_r)+(1-y)\cdot \log(1-\sigma(x_r))],
\end{equation}
\begin{equation}\label{eq:loss2}
loss2 = -[y\cdot \log \sigma(x_d)+(1-y)\cdot \log(1-\sigma(x_d))],
\end{equation}
where $\sigma$ means sigmoid function, y denotes the groundtruth labels, $x_r$ and $x_d$ denote the predicted results from RGB feature and depth feature, respectively.

\subsection{Feature Pyramid for Transformer}

After we have obtained the complemented features from the HCA module, we draw inspiration from the Feature Pyramid Networks (FPN) \cite{lin2017feature} to combine the transformer features from different levels to get multi-level multi-modal representations. 
Unlike previous methods \cite{chen2019multi,liu2021visual,zhao2022self} that construct feature pyramid using CNN features and distribute the features to each level equally, we highlight the contribution of the deepest level in the pyramid with a larger proportion and use the deep semantics to guide the selection of shallow features.  

\begin{figure}
	\centering
	\includegraphics[width=0.9\linewidth]{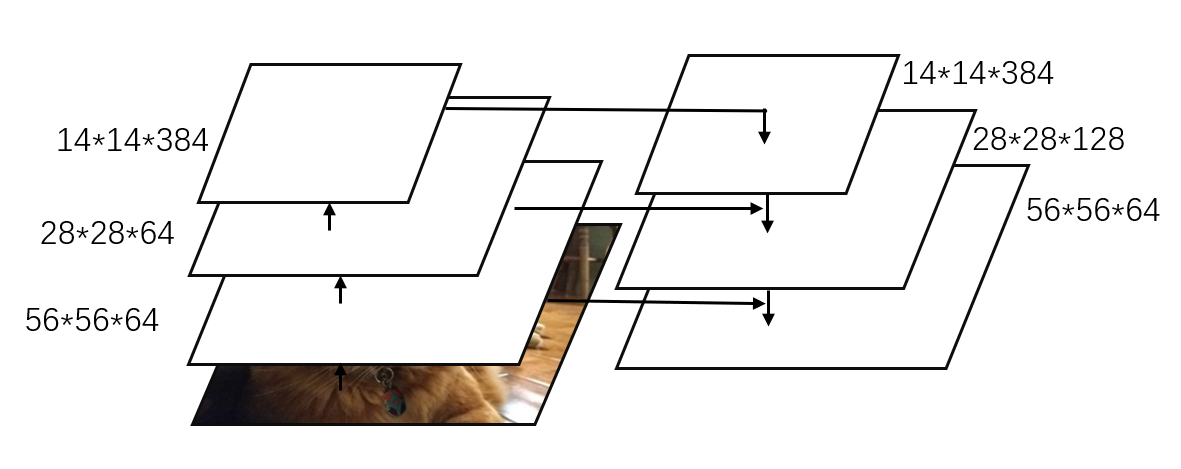}
	\caption{Details of the feature pyramid for transformer}
	\label{fig:fpn}
\end{figure}

The architecture of the FPT is shown in Fig. \ref{fig:fpn}. The features (14*14*384, 28*28*64, 56*56*64) from two T2T ViT encoder are aligned by progressively upsampling and concatenated with channel ratios 6:1, 2:1 and 1:1, respectively.

\subsection{Disentangled Complementing Module}

\begin{figure}
	\centering
	\includegraphics[width=0.9\linewidth]{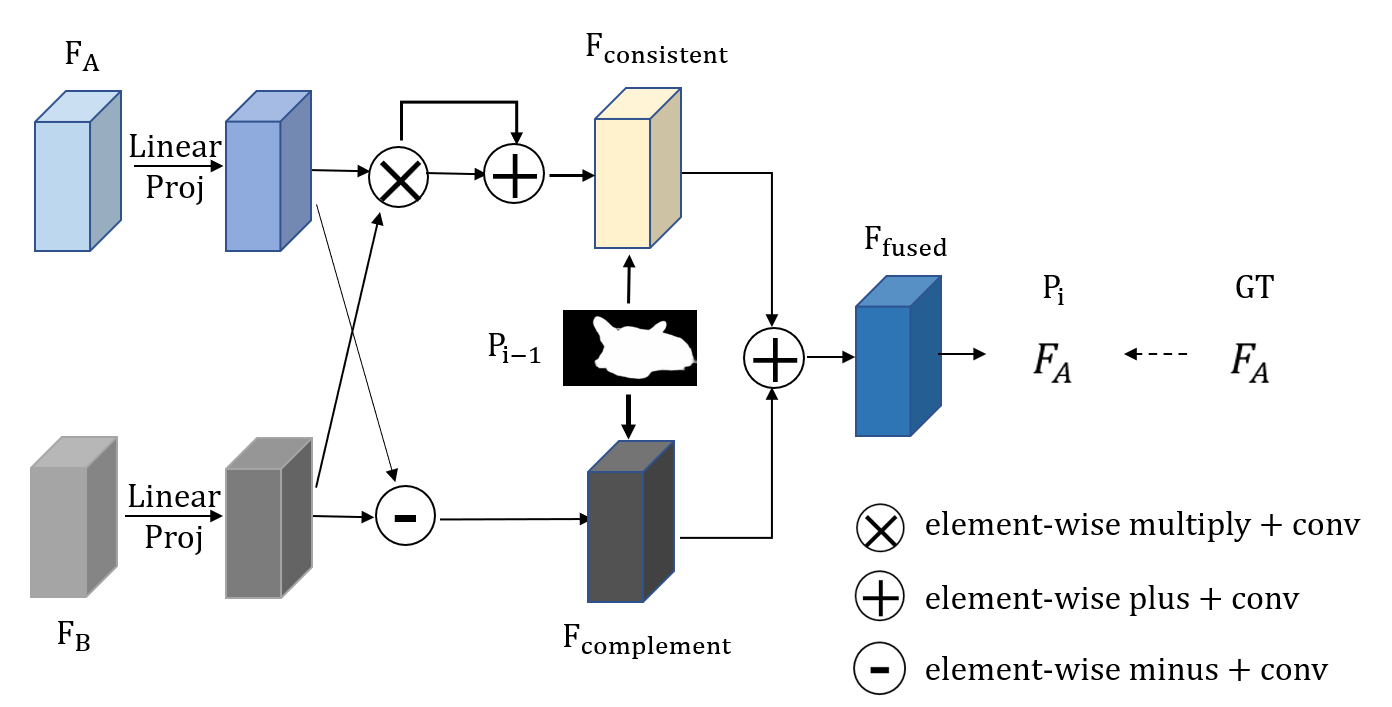}
	\caption{Details of the disentangled complementing module}
	\label{fig:ccm}
\end{figure}

Considering the complementarity among cross-modal features are heterogeneous, we draw inspiration from \cite{zhao2022self,9165931} to disentangle the cross-modal complements into consistent and complementary ones to improve the fusion adaptivity. Specifically shown in the DCM in Fig. \ref{fig:ccm}, the input RGB features $F_A$ and depth features $F_B$ will first be mapped to the same feature space by a linear projection and fed to two branches to extract consistent and complementary features, respectively. The branch above explores the consistency by point multiplication and a residual connection:
\begin{equation}\label{eq:consistency}
F_{consistent} = Conv(F_A^{map} \otimes F_B^{map} \oplus F_A^{map}), 
\end{equation}
where $F_A^{map}$ and $F_B^{map}$ mean the RGB features and depth features after mapping, $\otimes$ depicts element-wise multiplication, $\oplus$ means element-wise addition and $ Conv $ is the convolution layer. The branch below explores the complementarity by substraction: 
\begin{equation}\label{eq:complementarity}
F_{complement} = Conv(\left| F_A^{map} \ominus F_B^{map} \right| ) ,
\end{equation}
where $\ominus$ depicts element-wise subtraction.

To encourage the disentanglement of complements, we append the saliency groundtruth mask to supervise the prediction of the fused feature $F_{fused}$. Specifically, the saliency map $P_{i-1}$, predicted by the former DCM, carries important global contexts and localization cues. Hence, we use $P_{i-1}$ to guide this disentanglement in the i-th level by point multiplication with $F_{consistent}$ and $F_{complement}$ respectively to avoid noisy backgrounds in shallower layers. The constrained features are calculated as following formulas:
\begin{equation}\label{eq:constrain}
F_{consistent}' = Conv(F_{consistent}\otimes P_{i-1} ) ,
\end{equation}
\begin{equation}\label{eq:constrain2}
F_{complement}' = Conv(F_{complement}\otimes P_{i-1} ) .
\end{equation}
Finally, $F_{consistent}'$ and $F_{complement}'$ will be combined with an addition and a convolution operation:
\begin{equation}\label{eq:fused}
F_{fused} = Conv(F_{consistent}'\oplus F_{complement}'   ) .
\end{equation}
The prediction loss in the i-th DCM is 
\begin{equation}\label{eq:loss3}
loss_i = -[y\cdot \log \sigma(P_i)+(1-y)\cdot \log(1-\sigma(P_i))],
\end{equation}
where $P_i$ is the saliency map generated by the i-th DCM.

Compared with undifferentiated concatenation, our proposed DCM will explicitly diversify the complex cross-modal complements into the consistent and complementary elements.

The final loss includes 2 losses from the HCA module and 4 losses from DCMs:
\begin{equation}\label{eq:totalloss}
loss_{final} = loss_r + loss_d + \sum_{i=1}^{4} loss_i.
\end{equation}

\section{Experiments}

\begin{table*}[h]
	\centering
	\caption{Comparison with state-of-the-art RGB-D SOD methods. $\uparrow$ and $\downarrow$ mean that the larger and smaller ones are better respectively. The best results are labeled in \textcolor[rgb]{1,0,0}{red}. "-" indicates the code or result is not available. "Base" means VST model trained without using the edge detection task.}
	\setlength{\tabcolsep}{2mm}{
		\begin{tabular}{@{}lc|ccccccccc|cc@{}}
			\toprule
			Dataset & Metric&\makecell[c]{CMW}&\makecell[c]{Cas-Gnn}&\makecell[c]{HDFNet} &\makecell[c]{CoNet} &\makecell[c]{BBS-Net} & \makecell[c]{SSP}  & \makecell[c]{VST} &\makecell[c]{SPSN} &\makecell[c]{MVSalNet} & Base  &   HCT   \\ 
			\hline
			\multirow{4}{*}{\makecell[l]{NJUD}}
			&$S_m \uparrow $  &0.870 &0.911&0.908 &0.896 &0.921  &0.909 & 0.922 &0.918 &0.910 &0.925 &\textcolor[rgb]{1,0,0}{0.933}  \\
			&$maxF\uparrow $  &0.871&0.916&0.911&0.893 &0.919   &0.923 & 0.920 &- &0.922 &0.924 &\textcolor[rgb]{1,0,0}{0.932} \\
			&$E^{max}\uparrow$ &0.927&0.948&0.944&0.937 & 0.949   & 0.951 &0.939 &0.950 &0.939 &0.956 &\textcolor[rgb]{1,0,0}{0.960}\\
			&$MAE\downarrow$  &0.061&0.036&0.039&0.046 & 0.035   &0.039 & 0.035 &0.032 &0.035 &0.037 &\textcolor[rgb]{1,0,0}{0.030}\\
			\hline
			\multirow{4}{*}{\makecell[l]{NLPR}} 
			&$S_m \uparrow $  &0.917&0.919&0.923&0.912 &0.931   &0.922 & 0.932 &0.923 &0.927 &0.925    &\textcolor[rgb]{1,0,0}{0.934}  \\
			&$maxF\uparrow $  &0.903&0.906&0.917&0.893 &0.918  &0.889 & 0.920 &- &\textcolor[rgb]{1,0,0}{0.929} &0.913 &\textcolor[rgb]{0,0,1}{0.924} \\
			&$E^{max}\uparrow$ &0.951&0.955&\textcolor[rgb]{1,0,0}{0.963}&0.948 &0.961  &0.960 & 0.962 &0.958 &0.959 &0.957 &\textcolor[rgb]{1,0,0}{0.963}\\
			&$MAE\downarrow$  &0.027&0.028&0.027&0.027 &\textcolor[rgb]{0,0,1}{0.023}  &0.025 &\textcolor[rgb]{0,0,1}{0.023} & 0.024 &\textcolor[rgb]{1,0,0}{0.021} &0.030 &\textcolor[rgb]{0,0,1}{0.023}\\
			\hline
			\multirow{4}{*}{\makecell[l]{DUTLF}} 
			&$S_m \uparrow $  &0.797&0.920&0.908&0.923 &0.882 &0.929 & 0.943 &- &- &0.939  &\textcolor[rgb]{1,0,0}{0.947} \\
			&$maxF\uparrow $  &0.779&0.926&0.915&0.932 &0.870  &0.947 & 0.948 &- &- &0.945 &\textcolor[rgb]{1,0,0}{0.950}\\
			&$E^{max}\uparrow$ &0.864&0.953&0.945&0.959 & 0.912  &0.958 & \textcolor[rgb]{1,0,0}{0.969} &- &- &0.964 &\textcolor[rgb]{1,0,0}{0.969}\\
			&$MAE\downarrow$  &0.098&0.030&0.041&0.029 &0.058  &0.029 & 0.024 &- &- &0.030 &\textcolor[rgb]{1,0,0}{0.023}\\
			\hline
			\multirow{4}{*}{\makecell[l]{STERE}} 
			&$S_m \uparrow $ &0.852&0.899&0.900&0.905  &0.908 &0.904  & 0.913 &0.907 &0.911 &0.914 &\textcolor[rgb]{1,0,0}{0.923}    \\
			&$maxF\uparrow $ &0.837&0.901&0.900&0.901 &0.903  &0.914 & 0.907 &- &\textcolor[rgb]{1,0,0}{0.920} &0.907 &\textcolor[rgb]{0,0,1}{0.918}\\
			&$E^{max}\uparrow$ &0.907&0.944&0.943&0.947 &0.942  &0.939 & 0.951 &0.943 &0.946 &0.949 &\textcolor[rgb]{1,0,0}{0.955}\\
			&$MAE\downarrow$ &0.067&0.039&0.042&0.037 &0.041 &0.039  & 0.038 &0.035 &0.035 &0.042 &\textcolor[rgb]{1,0,0}{0.034}\\
			\hline
			\multirow{4}{*}{\makecell[l]{RGBD135}} 
			&$S_m \uparrow $  &0.934&0.894&0.926&0.914 &0.934 &0.936 & 0.943 &- &0.931 &0.934 &\textcolor[rgb]{1,0,0}{0.946} \\
			&$maxF\uparrow $  &0.931&0.894&0.921&0.902 &0.928  &0.944 & 0.940 &- &0.934 &0.922 &\textcolor[rgb]{1,0,0}{0.945}\\
			&$E^{max}\uparrow$ &0.99&0.937&0.970 &0.948 &0.966  &0.978 & 0.978 &- &0.971 &0.968 &\textcolor[rgb]{1,0,0}{0.978}\\
			&$MAE\downarrow$  &0.022&0.028&0.022&0.024 &0.021  &0.017 & 0.017 &- &0.019 &0.022 &\textcolor[rgb]{1,0,0}{0.016}\\
			\hline
			\multirow{4}{*}{\makecell[l]{SIP}} 
			&$S_m \uparrow $  &0.705&-&0.886&0.860 &0.879  &0.888& 0.904 &0.892 &- &0.918 &\textcolor[rgb]{1,0,0}{0.922}\\
			&$maxF\uparrow $  &0.677&-&0.894&0.873 &0.884   &0.909& 0.915 &- &- &0.930 &\textcolor[rgb]{1,0,0}{0.935}\\
			&$E^{max}\uparrow$ &0.804&-&0.930&0.917  &0.922   &0.927& 0.944 &0.934 &- &0.955 &\textcolor[rgb]{1,0,0}{0.958}\\
			&$MAE\downarrow$  &0.141&-&0.048&0.048 &0.055   &0.046& 0.040 &0.042 &- &0.037 &\textcolor[rgb]{1,0,0}{0.031}\\
			\bottomrule
	\end{tabular}}
	\label{tab:rgbd_sod}
\end{table*}

\subsection{Datasets}
We evaluate the proposed model on seven public RGB-D SOD datasets which are NJUD \cite{ju2014depth} (1985 image pairs), NLPR \cite{peng2014rgbd} (1000 image pairs), DUTLF \cite{piao2019depth}(1200 image pairs), STERE \cite{niu2012leveraging} (1000 image pairs),  RGBD135 \cite{cheng2014depth} (135 image pairs), SIP \cite{fan2020rethinking} (929 image pairs) and COME15K \cite{zhang2021rgb} (15625 image pairs). We follow the consistent setting in previous works \cite{liu2021visual,zhang2021rgb} and choose 1485 image pairs in NJUD, 700 image pairs in NLPR, 800 image pairs in DUTLF and 8025 pairs image in COME15K as the training set and the remaining are for testing. Similarly. we also adopt some data augmentation techniques such as resize, random crop and random flipping to avoid overfitting. 

\subsection{Evaluation Metrics}
We adopt four widely used evaluation metrics to evaluate our model. Specifically,
Structure-measure $S_m$ \cite{fan2017structure} evaluates region-aware and
object-aware structural similarity. 
E-measure $E^{max}_\rho$ \cite{fan2018enhanced}
simultaneously considers pixel-level errors and
image-level errors.
Maximum F-measure \cite{achanta2009frequency} jointly considers precision and recall under the optimal threshold.   
Mean Absolute Error (MAE) computes pixel-wise average absolute error.

\subsection{Implementation Details}
We implement our model on the base of VST \cite{liu2021visual} using Pytorch and train it on a RTX 3090. The training parameters are set as follows: batch size is 8, epoch is 50.
For the optimizer, we use Adam with the learning rate gradually decaying from $10^{-4}$ to $10^{-6}$.

\subsection{Comparisons with State-of-the-art}

To quantitatively measure our model, we compare it with 6 SOTA RGB-D SOD methods, including CMW \cite{li2020cross}, Cas-Gnn \cite{luo2020cascade}, HDFNet \cite{pang2020hierarchical}, CoNet \cite{ji2020accurate}, BBS-Net \cite{fan2020bbs} and VST \cite{liu2021visual}. Tab. \ref{tab:rgbd_sod} shows the comparison in terms of the S-measure, F-measure, E-measure and MAE scores. To fairly demonstrate the advantages of our cross-modal fusion scheme, we select the VST without the edge detection task and retrain it with our training set as the baseline model (denoted by "base"), as we argue that edge detection is a trick requiring additional edge labels and not related to multimodal fusion. The quantitative results illustrate that VST achieves consistent improvement over CNN-based methods, denoting the superiority of the transformer. Our model outperforms previous RGB-D SOD models, including VST on all datasets, demonstrating the advantages of our cross-modal fusion scheme and designs.

\begin{figure}[!ht] 
	\centering 
        \captionsetup[subfloat]{labelsep=none,format=plain,labelformat=empty}
	\begin{minipage}[b]{0.99\linewidth} 
		\subfloat[RGB]{
			\begin{minipage}[b]{0.13\linewidth} 
				\centering
				\includegraphics[width=\linewidth]{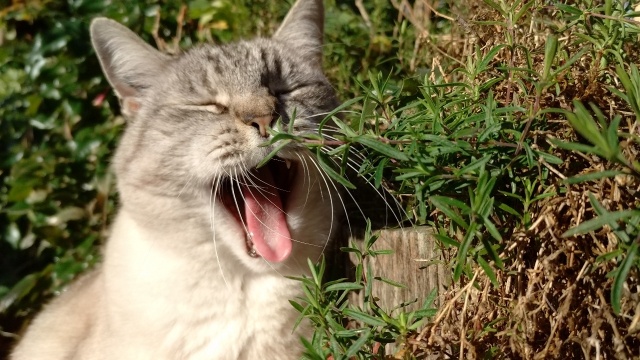}\vspace{1pt}
				\includegraphics[width=\linewidth]{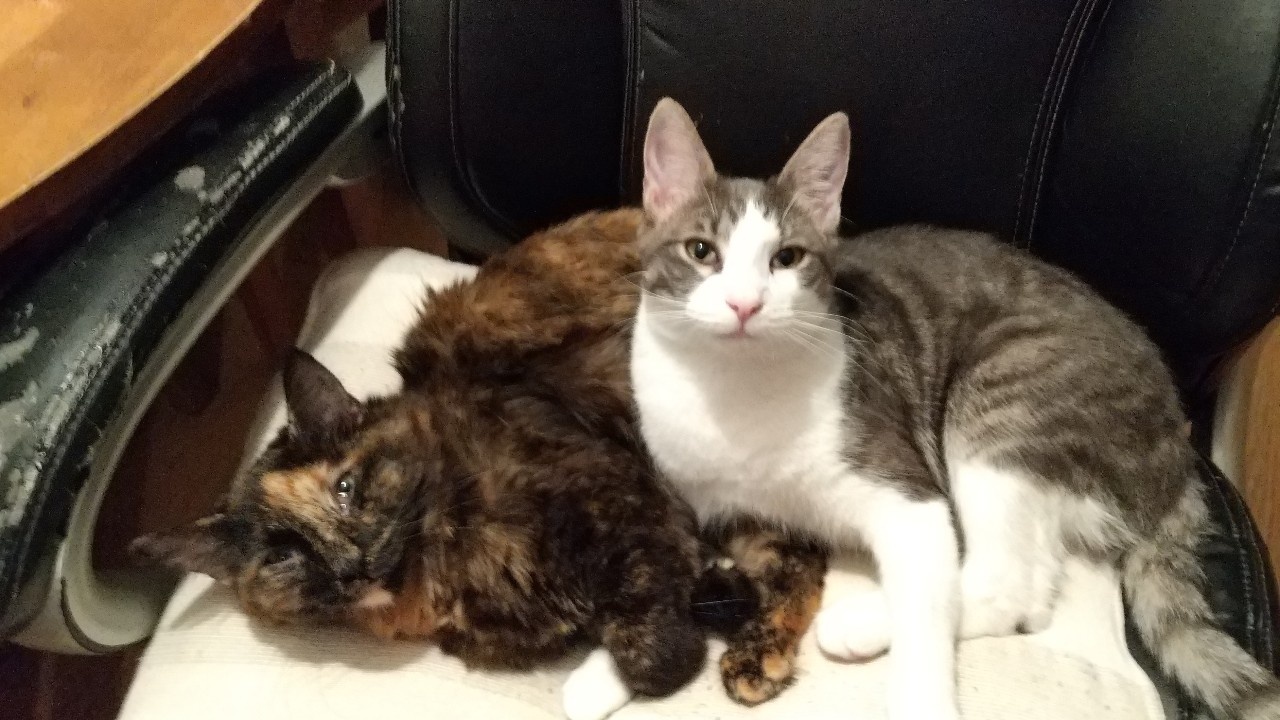}\vspace{1pt}
				\includegraphics[width=\linewidth]{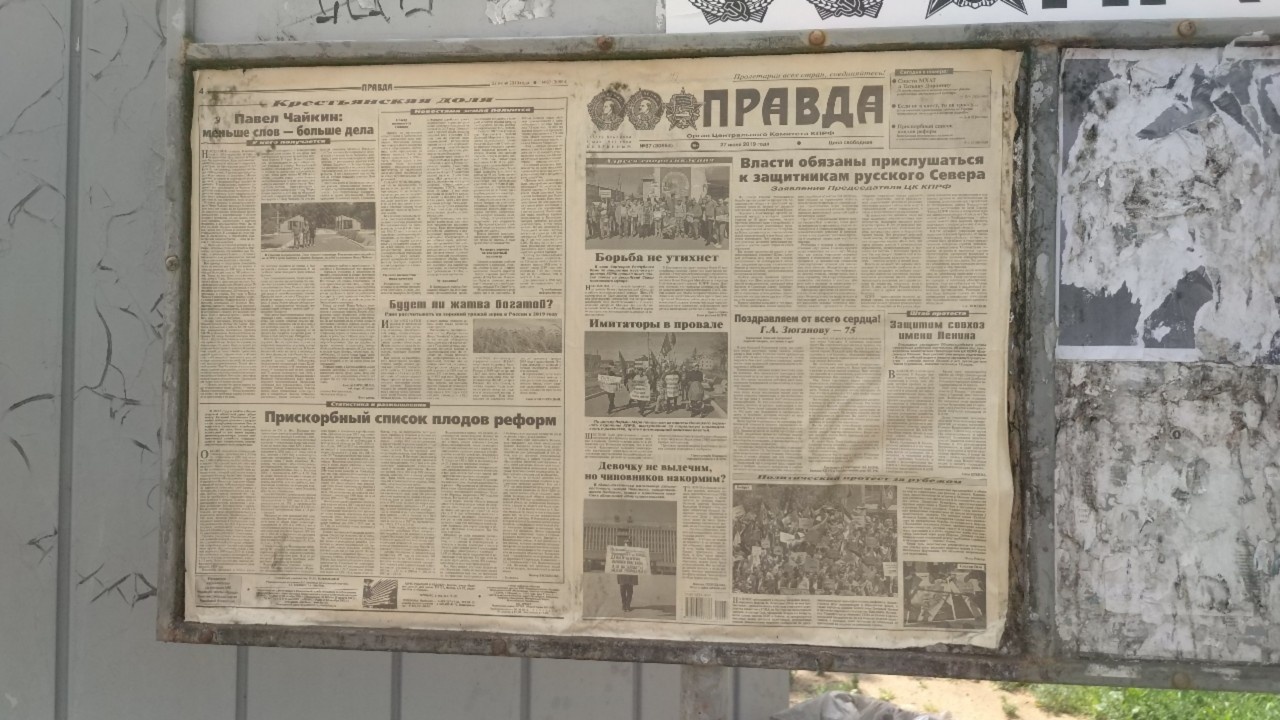}\vspace{1pt}
				\includegraphics[width=\linewidth]{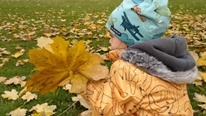}\vspace{1pt}
				\includegraphics[width=\linewidth]{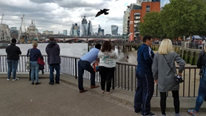}\vspace{1pt}
				\includegraphics[width=\linewidth]{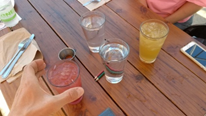}
			\end{minipage}
		}\hspace{-5pt}
		\subfloat[Depth]{
			\begin{minipage}[b]{0.13\linewidth}
				\centering
				\includegraphics[width=\linewidth]{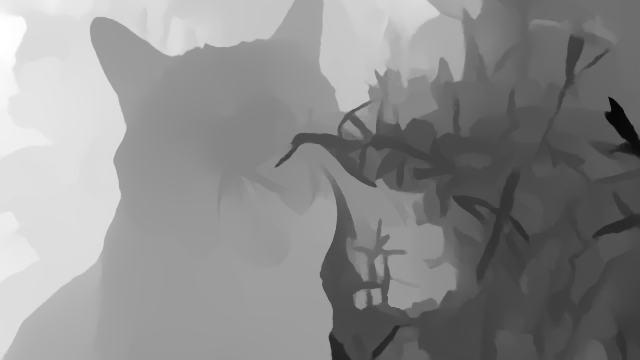}\vspace{1pt}
				\includegraphics[width=\linewidth]{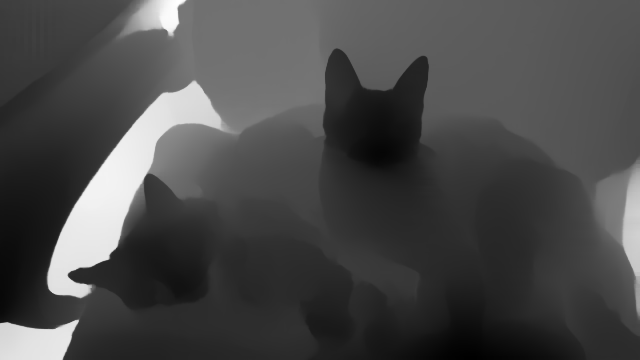}\vspace{1pt}
				\includegraphics[width=\linewidth]{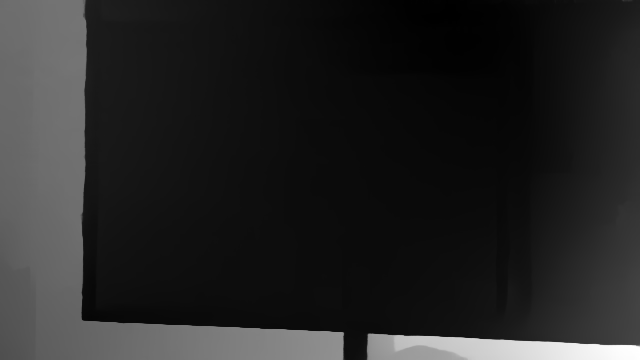}\vspace{1pt}
				\includegraphics[width=\linewidth]{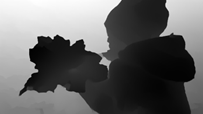}\vspace{1pt}
				\includegraphics[width=\linewidth]{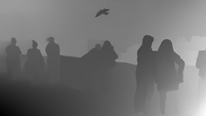}\vspace{1pt}
				\includegraphics[width=\linewidth]{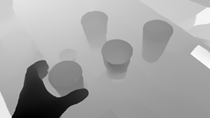}
			\end{minipage}
		}\hspace{-5pt}
		\subfloat[GT]{
			\begin{minipage}[b]{0.13\linewidth}
				\centering
				\includegraphics[width=\linewidth]{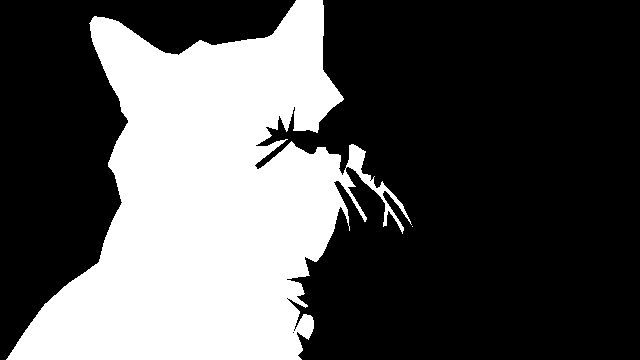}\vspace{1pt}
				\includegraphics[width=\linewidth]{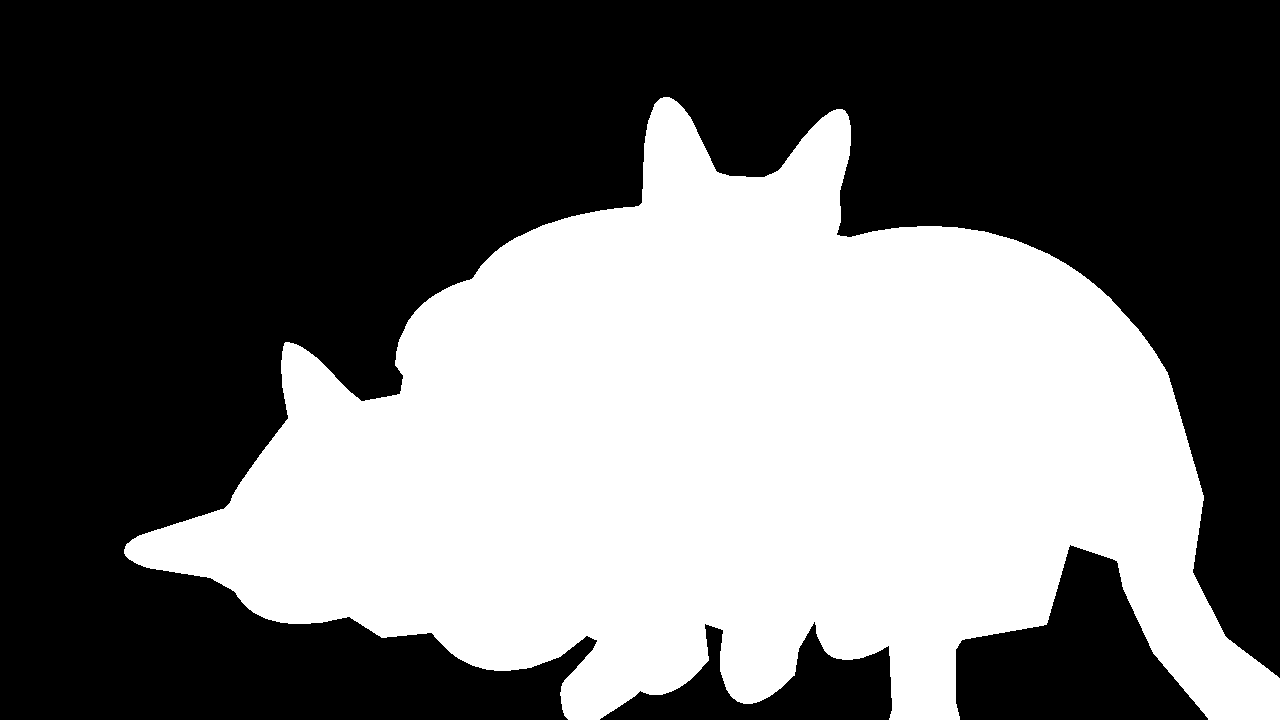}\vspace{1pt}
				\includegraphics[width=\linewidth]{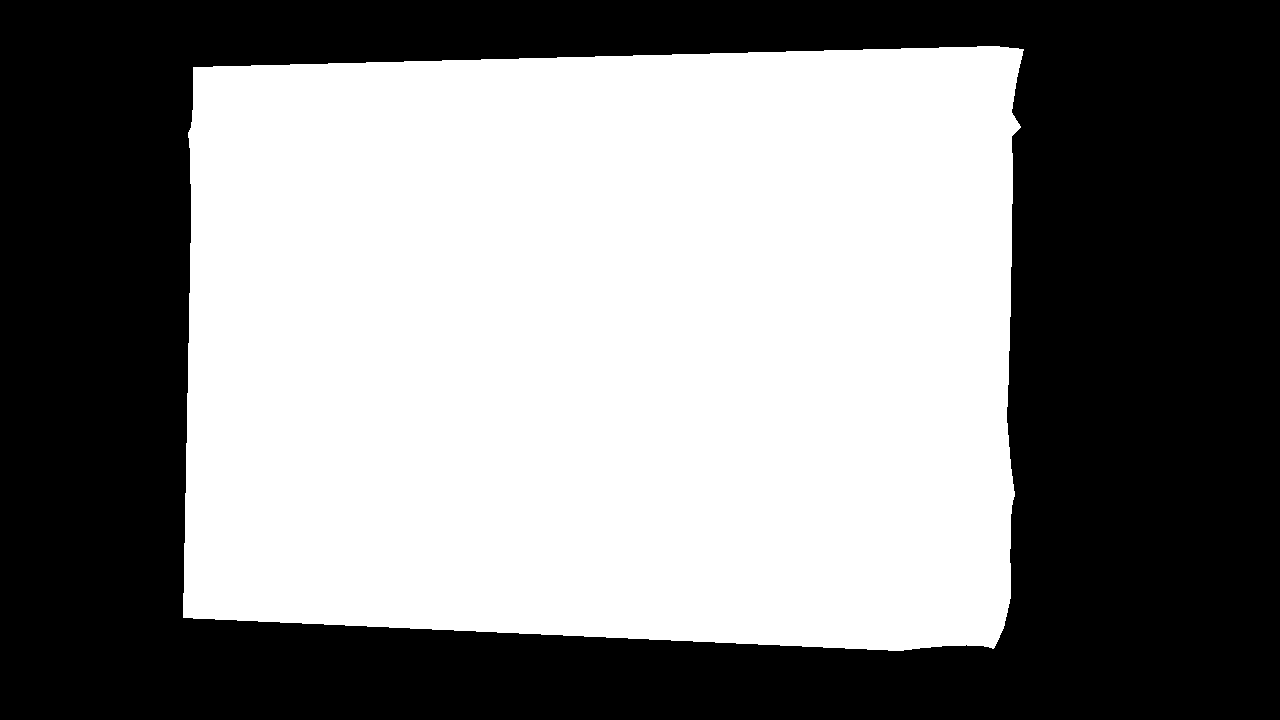}\vspace{1pt}
				\includegraphics[width=\linewidth]{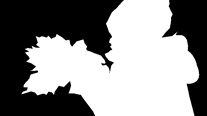}\vspace{1pt}
				\includegraphics[width=\linewidth]{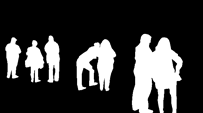}\vspace{1pt}
				\includegraphics[width=\linewidth]{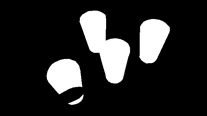}
			\end{minipage}
		}\hspace{-5pt}
		\subfloat[CoNet]{
			\begin{minipage}[b]{0.13\linewidth}
				\centering
				\includegraphics[width=\linewidth]{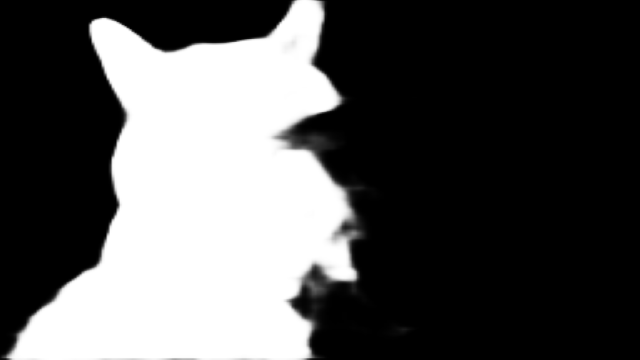}\vspace{1pt}
				\includegraphics[width=\linewidth]{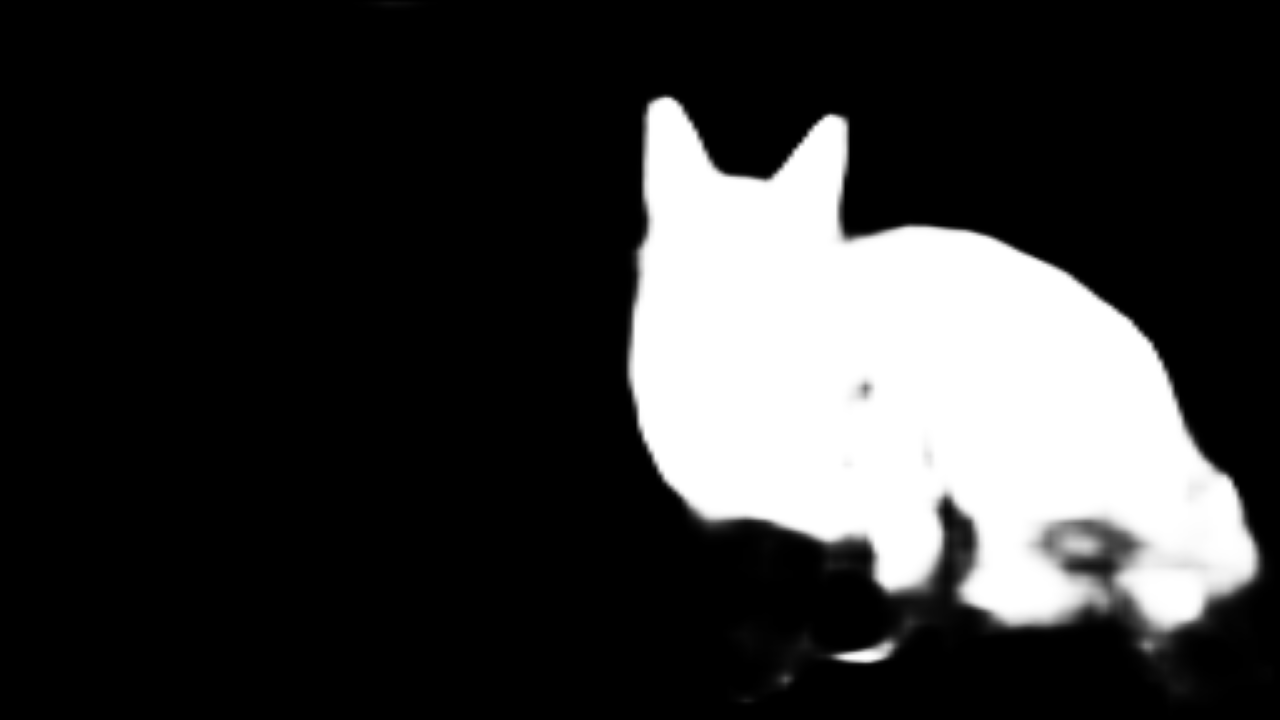}\vspace{1pt}
				\includegraphics[width=\linewidth]{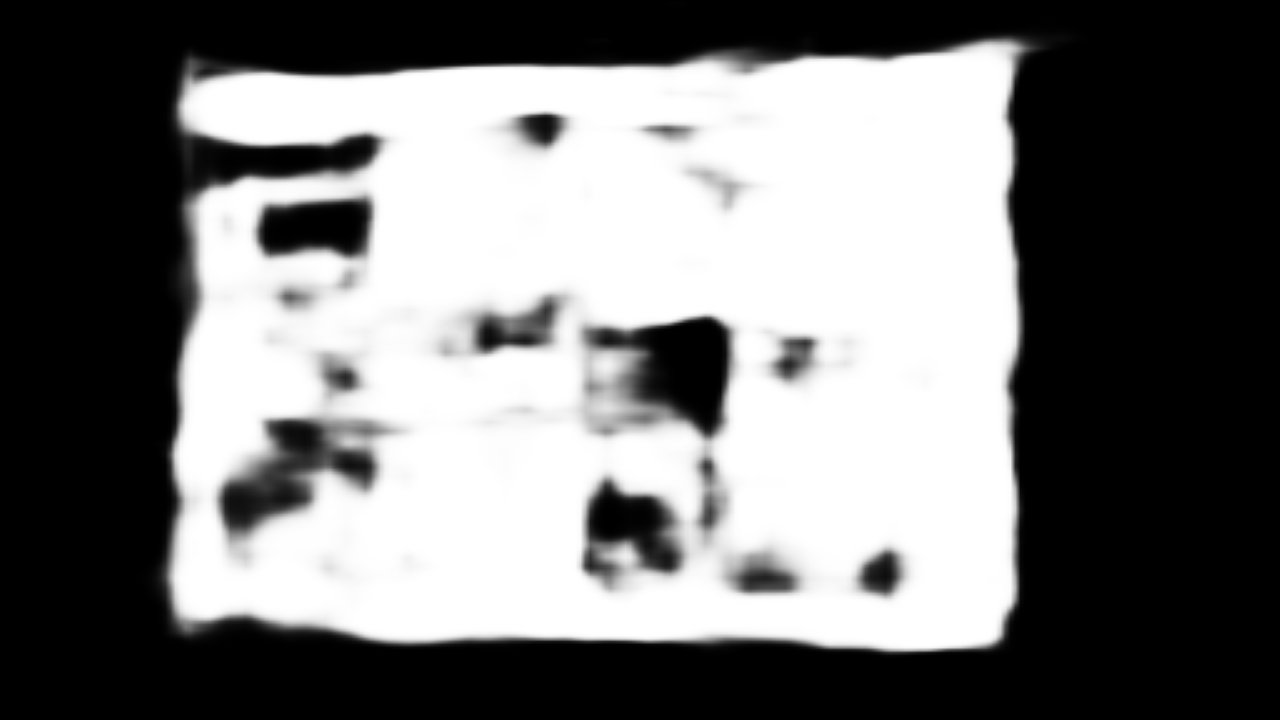}\vspace{1pt}
				\includegraphics[width=\linewidth]{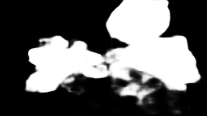}\vspace{1pt}
				\includegraphics[width=\linewidth]{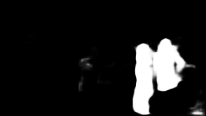}\vspace{1pt}
				\includegraphics[width=\linewidth]{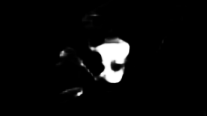}

			\end{minipage}
		}\hspace{-5pt}
		\subfloat[BBS-Net]{
			\begin{minipage}[b]{0.13\linewidth}
				\centering
				\includegraphics[width=\linewidth]{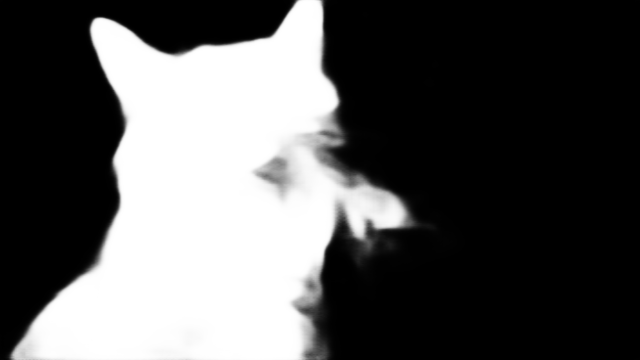}\vspace{1pt}
				\includegraphics[width=\linewidth]{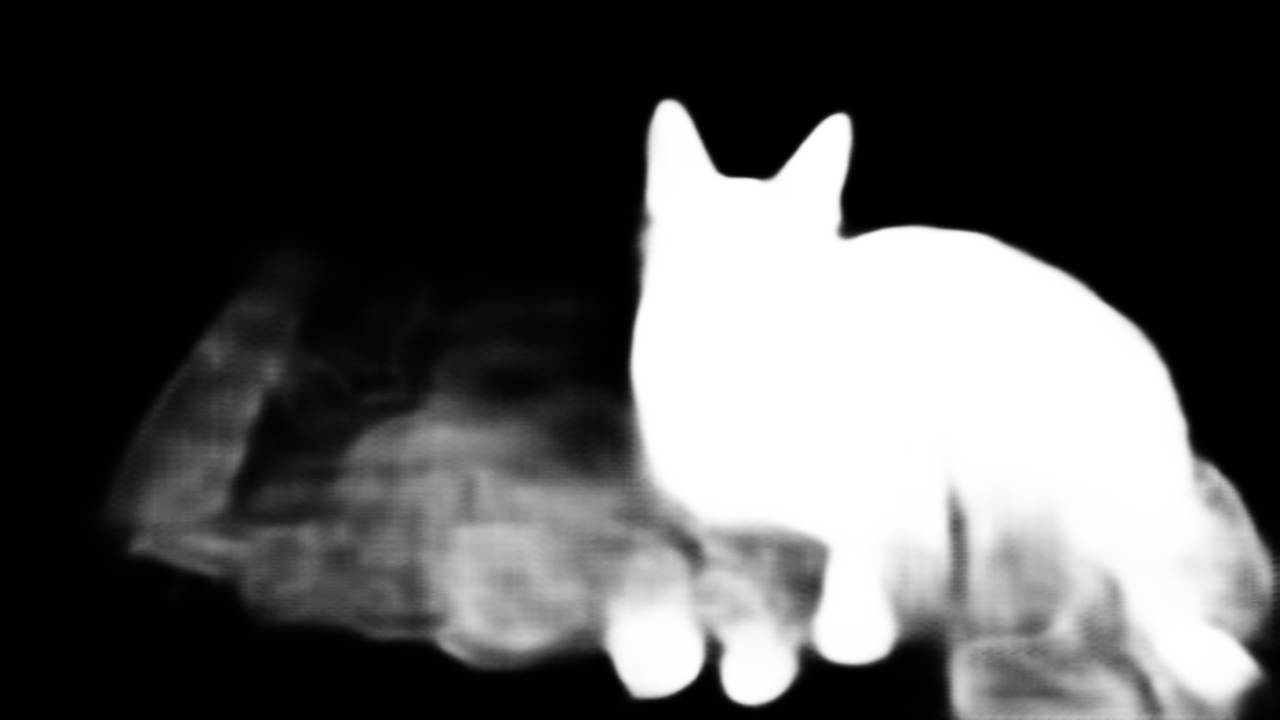}\vspace{1pt}
				\includegraphics[width=\linewidth]{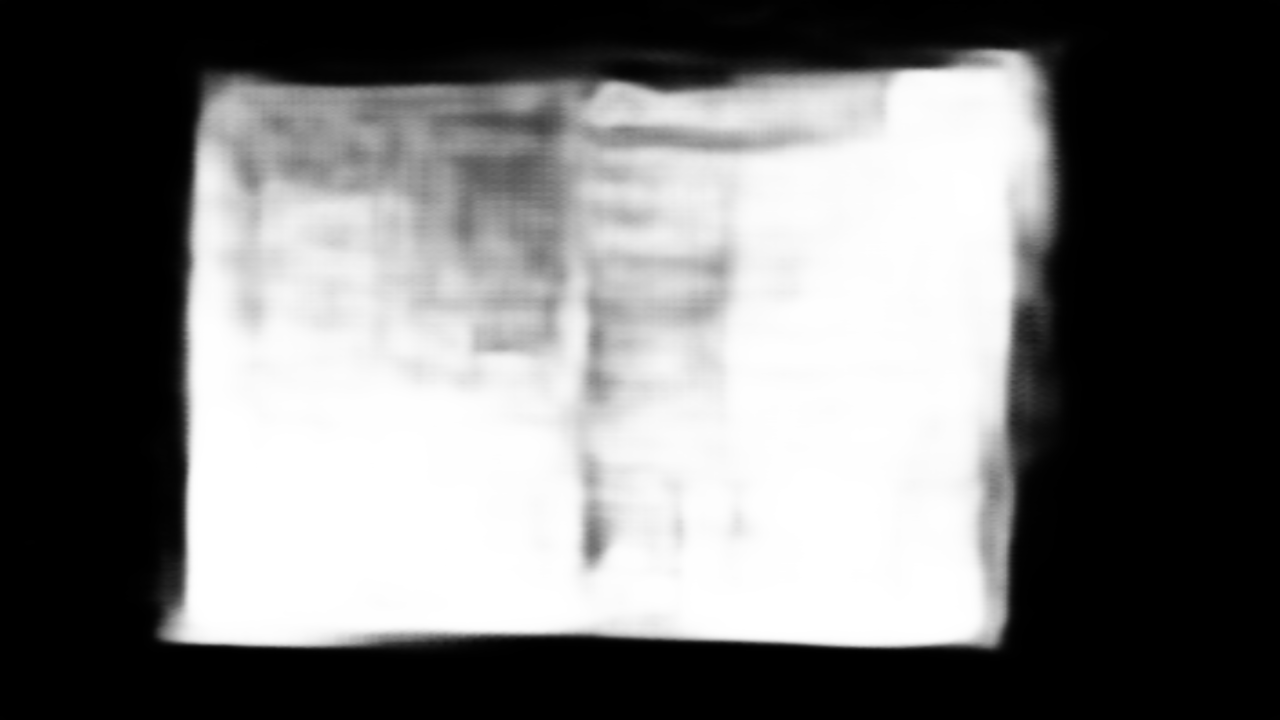}\vspace{1pt}
				\includegraphics[width=\linewidth]{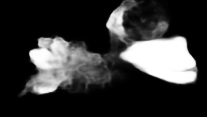}\vspace{1pt}
				\includegraphics[width=\linewidth]{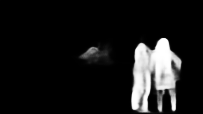}\vspace{1pt}
				\includegraphics[width=\linewidth]{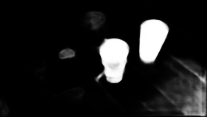}
			\end{minipage}
		}\hspace{-5pt}
		\subfloat[Base]{
			\begin{minipage}[b]{0.13\linewidth}
				\centering
				\includegraphics[width=\linewidth]{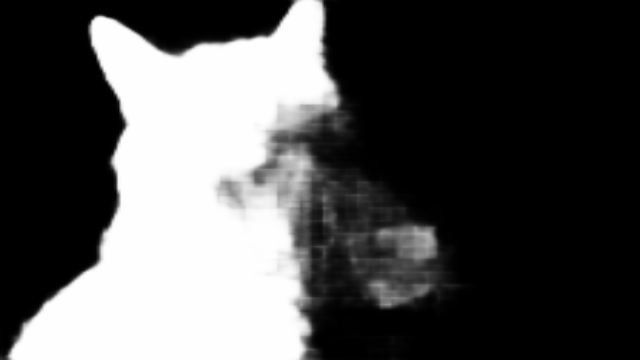}\vspace{1pt}
				\includegraphics[width=\linewidth]{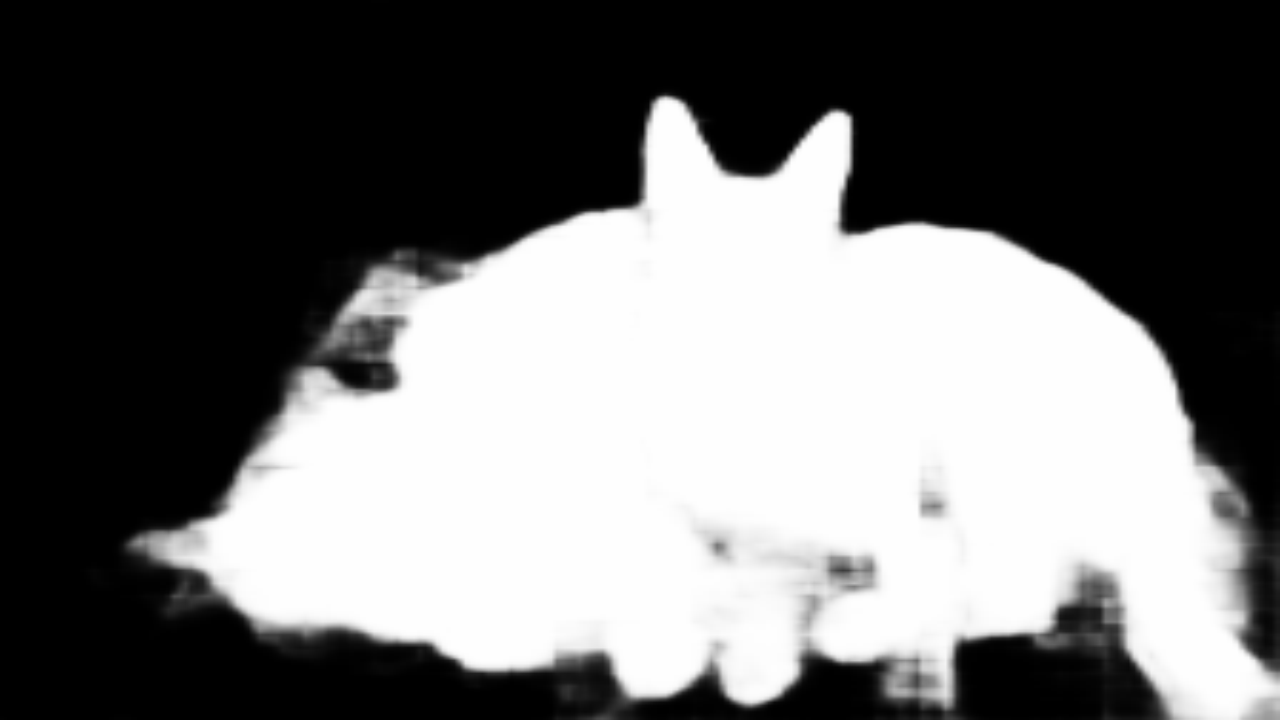}\vspace{1pt}
				\includegraphics[width=\linewidth]{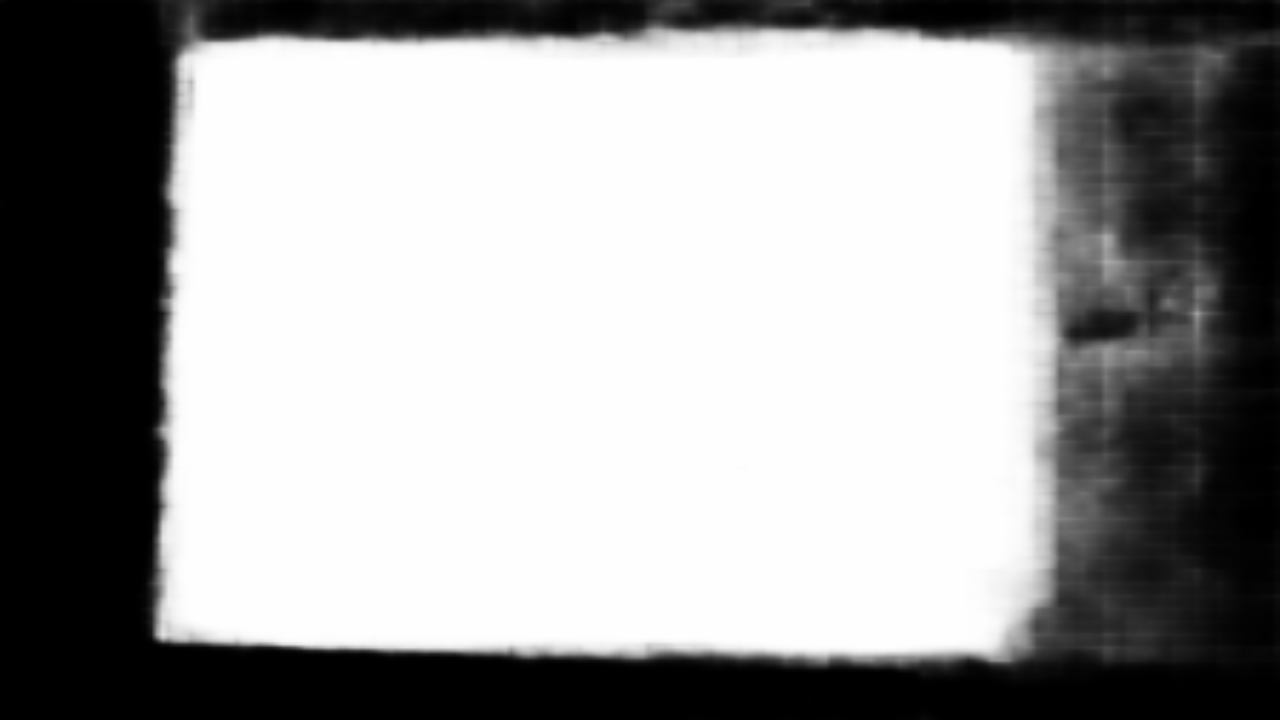}\vspace{1pt}
				\includegraphics[width=\linewidth]{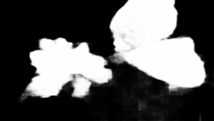}\vspace{1pt}
				\includegraphics[width=\linewidth]{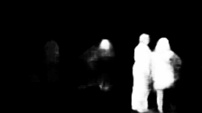}\vspace{1pt}
				\includegraphics[width=\linewidth]{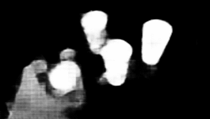}

			\end{minipage}
		}\hspace{-5pt}
		\subfloat[HCT]{
			\begin{minipage}[b]{0.13\linewidth}
				\centering
				\includegraphics[width=\linewidth]{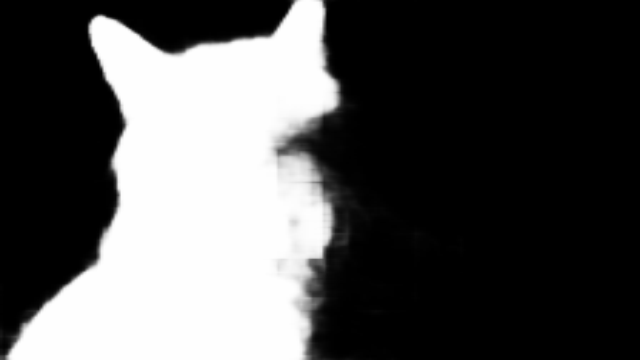}\vspace{1pt}
				\includegraphics[width=\linewidth]{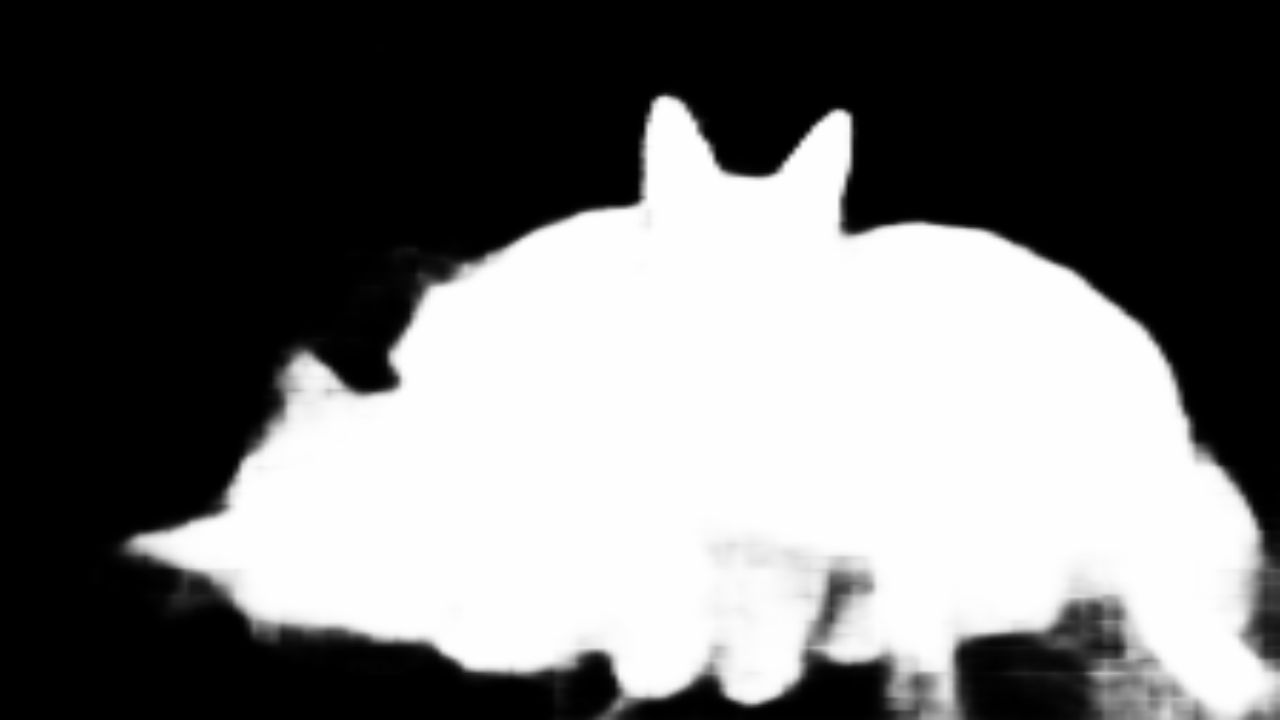}\vspace{1pt}
				\includegraphics[width=\linewidth]{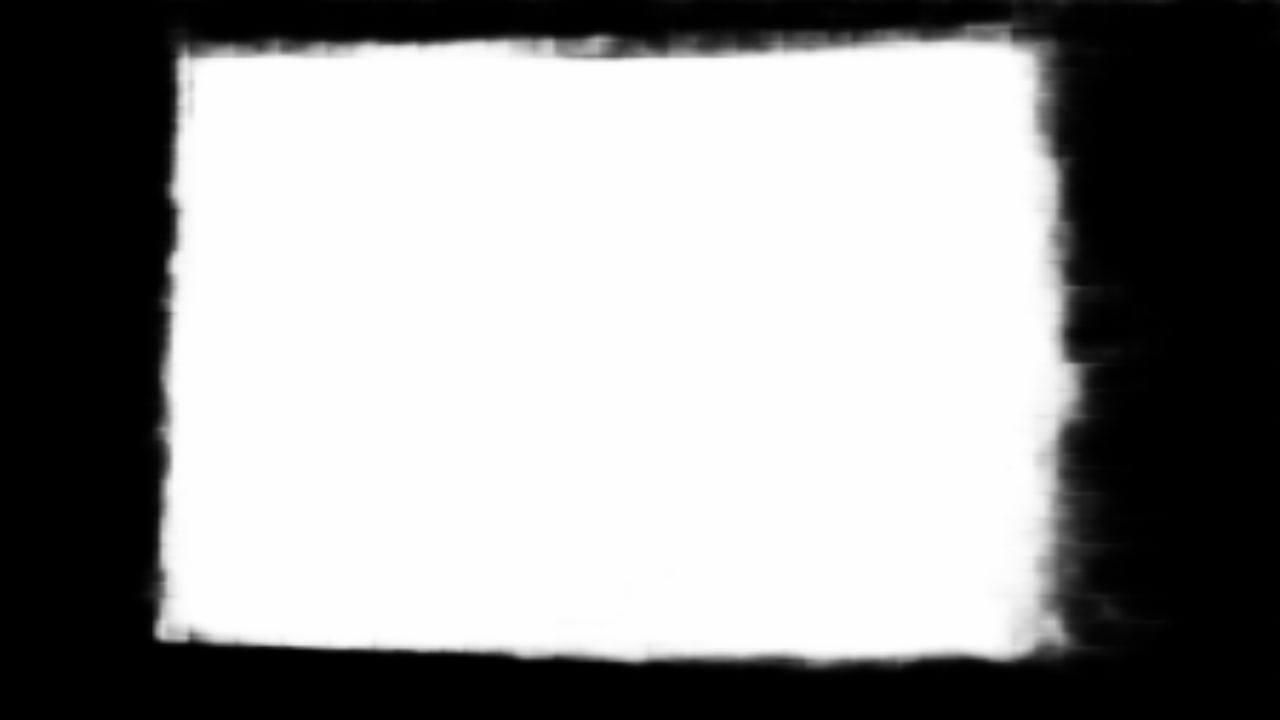}\vspace{1pt}
				\includegraphics[width=\linewidth]{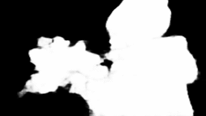}\vspace{1pt}
				\includegraphics[width=\linewidth]{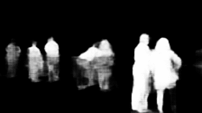}\vspace{1pt}
				\includegraphics[width=\linewidth]{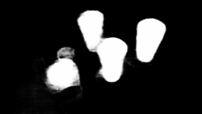}
			\end{minipage}
		}
	\end{minipage}
	\vfill
	\caption{Visual comparison to SOTA RGB-D SOD methods}
	\label{fig:result}
\end{figure}

To visually measure our model, we compare it with 3 recent SOTA RGB-D SOD methods, including CoNet \cite{ji2020accurate}, BBS-Net \cite{fan2020bbs} and the Base\cite{liu2021visual}. The visualized results on some representative challenging scenes are illustrated in Fig. \ref{fig:result}. In the case of large intra-difference in the foreground (see the $1^{st} $,  $2^{nd}$ and $4^{th}$ rows), previous models often fail to completely detect the correct salient objects, while our model can achieve more accurate and uniform detection. When the foreground and background hold similar appearance or depth (see the $1^{st} $, $3^{rd}$ and $6^{th}$ rows), previous methods may mistake some background areas as foreground, while our model successfully handles this confusion and well leverage the discriminating modality. For scenes having multiple salient objects (see the $2^{nd} $, $5^{th}$ and $6^{th}$ rows), other models tends to overlook some regions, while our method can highlight all salient objects. 
The success on these difficult cases demonstrates our advantages in modeling within/cross-modal long-range dependencies and local correlations.

\subsection{Ablation Study}

\begin{table*}[h]
	\centering
	\caption{Ablation study on our designs. 'Base-CA' denotes the variant by removing cross-modal attention in VST. HCA means hierarchical cross-modal attention module, FPT means the feature pyramid for transformer and DCM denotes the disentangled complementing module.} 
        \setlength{\tabcolsep}{1.5mm}{
		\begin{tabular}{@{}l|ccccc|cccc@{}}
			\hline
			Settings & \multicolumn{4}{c}{COME-E \cite{zhang2021rgb}} & & \multicolumn{4}{c}{COME-H \cite{zhang2021rgb}}   \\
			& $S_m \uparrow $& $maxF\uparrow $&$E^{max}\uparrow$&$MAE\downarrow$ &
			& $S_m \uparrow $& $maxF\uparrow $&$E^{max}\uparrow$&$MAE\downarrow$ \\
			\hline
			Base          &0.902 &0.899 &0.940 &0.042 &&0.865 &0.865 &0.903& 0.067 \\
                Base-CA          &0.904  &0.901  & 0.940 & 0.039 && 0.866 & 0.866 & 0.903 & 0.065 \\
			+HCA              &0.906 &0.904 &0.942 &0.038 &&0.868 &0.869 &0.907 &0.062  \\
			+HCA+FPT         &0.908 &0.906 &0.943 &0.037 &&0.871 &0.871 &0.908 &0.061 \\
			+HCA+FPT+DCM   &\textcolor[rgb]{1,0,0}{0.911} &\textcolor[rgb]{1,0,0}{0.910} &\textcolor[rgb]{1,0,0}{0.945} &\textcolor[rgb]{1,0,0}{0.035} &&\textcolor[rgb]{1,0,0}{0.873} &\textcolor[rgb]{1,0,0}{0.872} &\textcolor[rgb]{1,0,0}{0.910} &\textcolor[rgb]{1,0,0}{0.059} \\
			
			\midrule
	\end{tabular}}
	\label{tab:ablation}
\end{table*}

In this section, we will verify the effectiveness of each proposed structure by comprehensive ablation experiments. The experiments are implemented on 2 largest datasets, i.e., COME-EASY and COME-HARD. We select the Base as the baseline model (noted as "Base").

\textbf{Effectiveness of hierarchical cross-modal attention.}
As shown in Tab. \ref{tab:ablation}, compared to the original VST baseline (denoted by "Base"), removing the global cross-attention layer in VST (denoted by "Base-CA") surprisingly improves the performance, suggesting the cross-modal gap and the irrationality of using spatial-distant cross-modal dependencies as complementary contexts. In contrast, our hierarchical cross-modal attention scheme (denoted by "+HCA") shows noticeable improvement, which well supports our motivation that cross-modal gap and spatial discrepancy should not be concurrent when measuring cross-modal correlations.

\begin{figure}[!ht] 
	\centering 
        \captionsetup[subfloat]{labelsep=none,format=plain,labelformat=empty}
	\begin{minipage}[b]{0.99\linewidth} 
		\subfloat[RGB]{
			\begin{minipage}[b]{0.15\linewidth} 
				\centering
				\includegraphics[width=\linewidth]{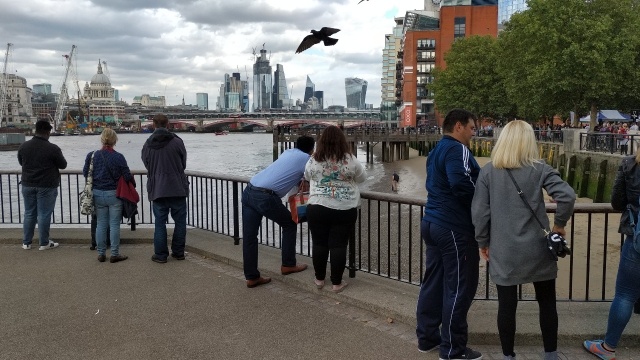}\vspace{1pt}
				\includegraphics[width=\linewidth]{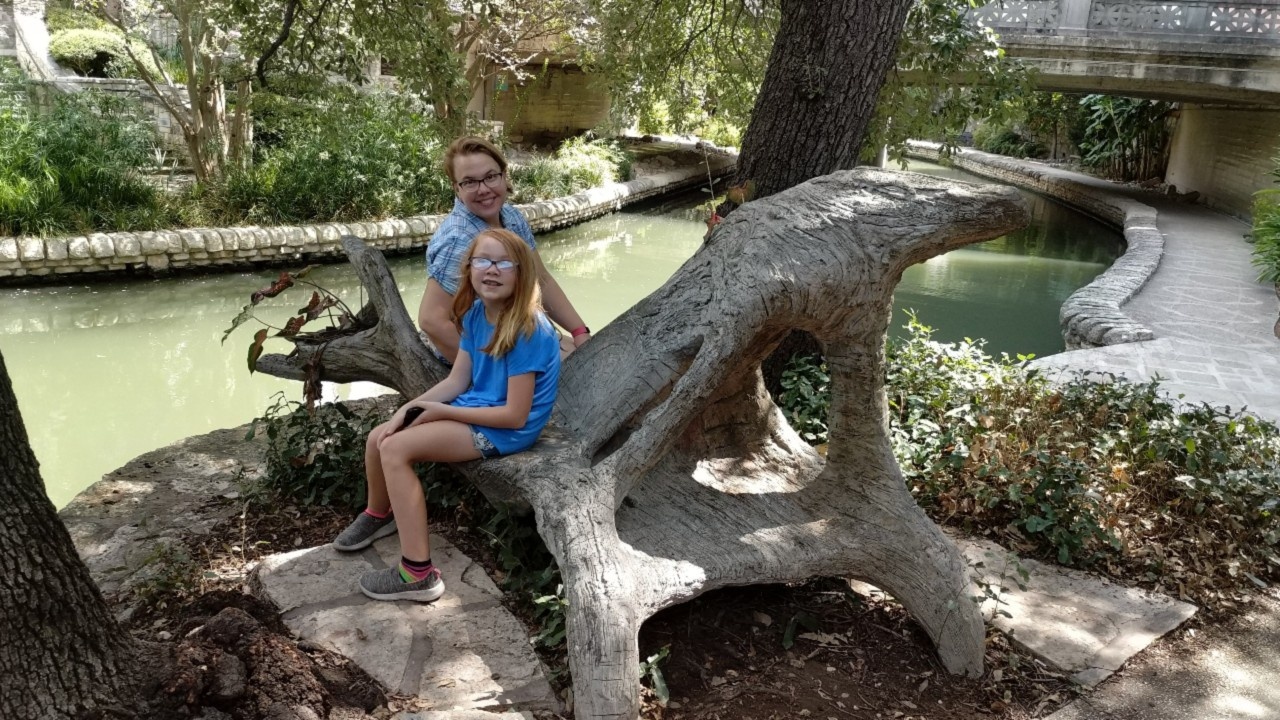}\vspace{1pt}
				\includegraphics[width=\linewidth]{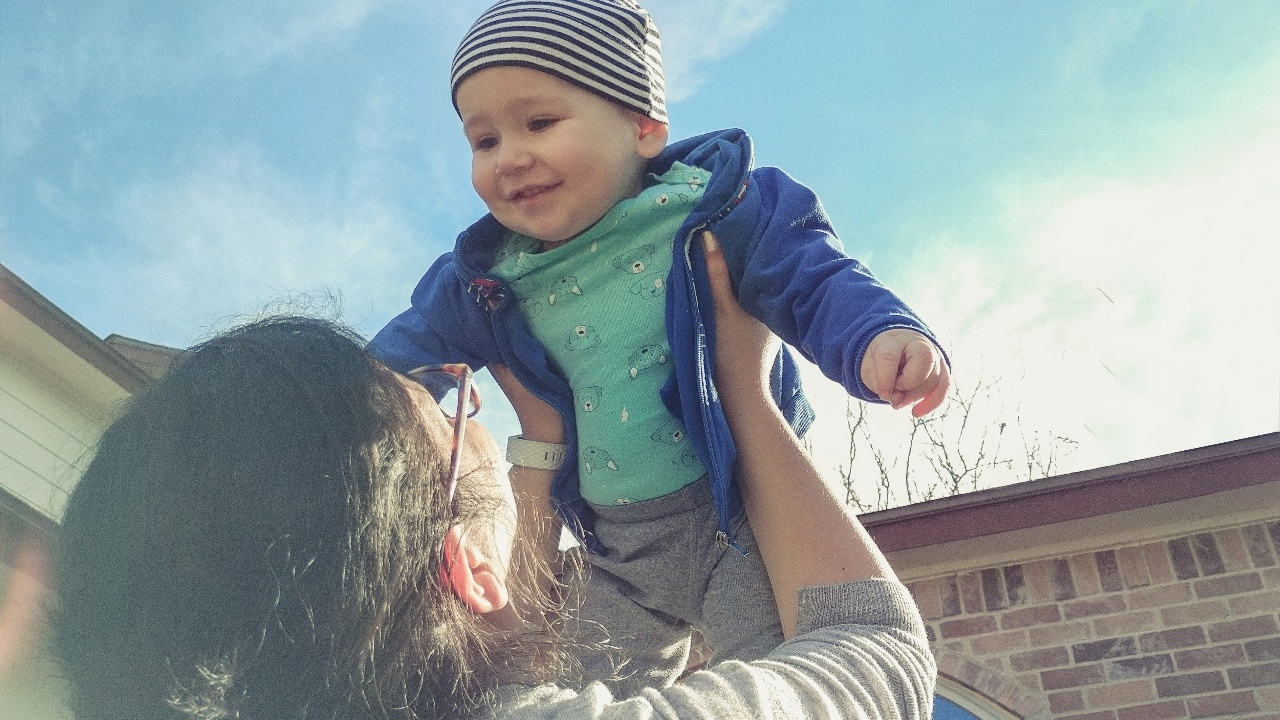}\vspace{1pt}
				\includegraphics[width=\linewidth]{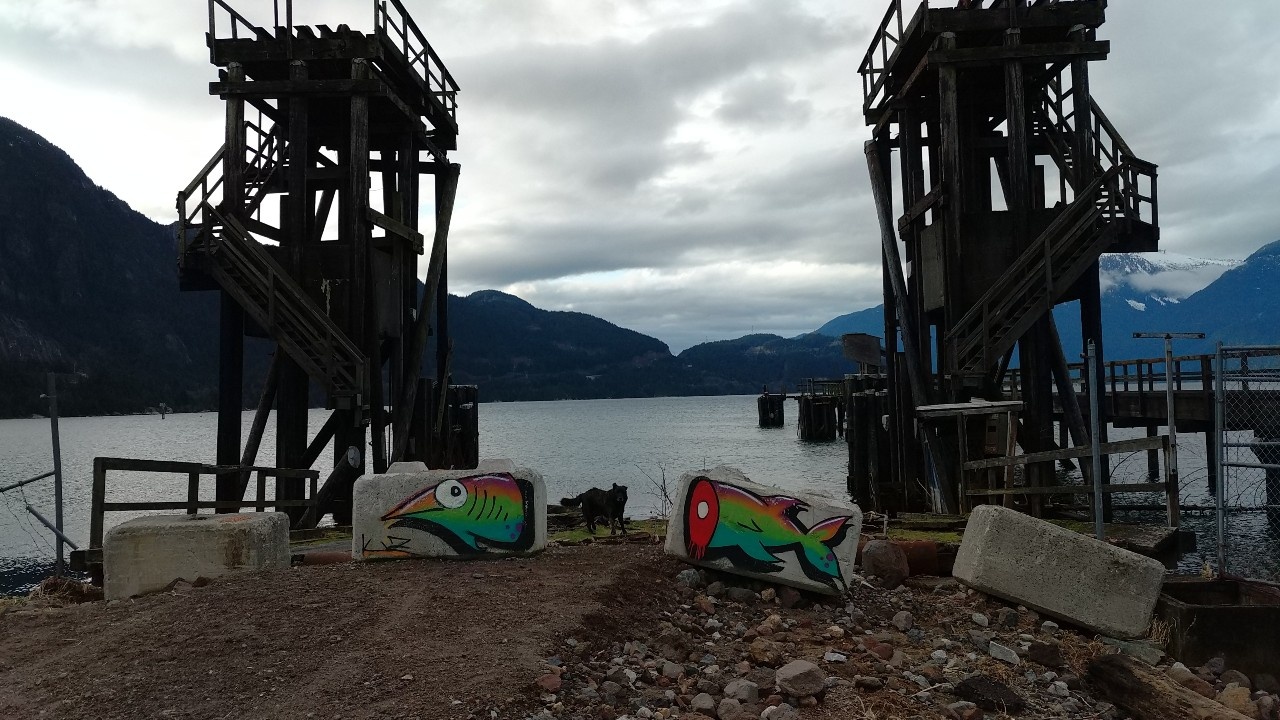}
				
			\end{minipage}
		}\hspace{-4pt}
		\subfloat[Depth]{
			\begin{minipage}[b]{0.15\linewidth}
				\centering
				\includegraphics[width=\linewidth]{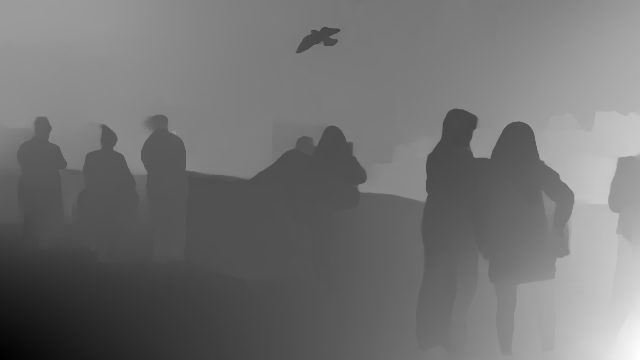}\vspace{1pt}
				\includegraphics[width=\linewidth]{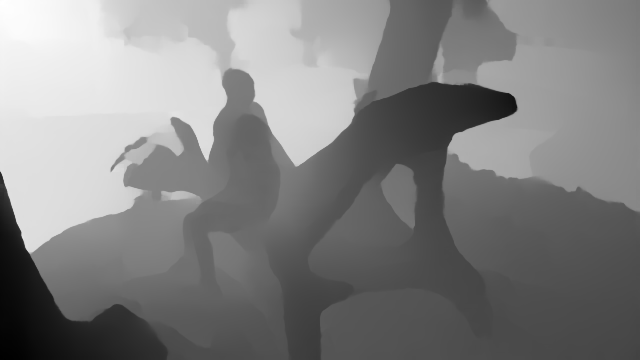}\vspace{1pt}
				\includegraphics[width=\linewidth]{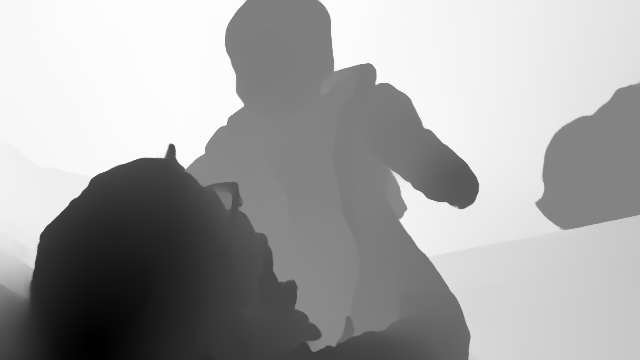}\vspace{1pt}
				\includegraphics[width=\linewidth]{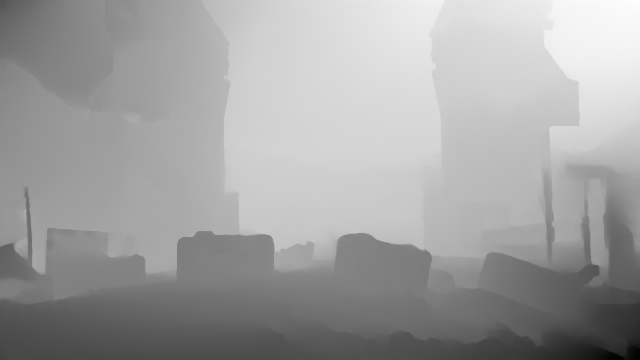}
				
			\end{minipage}
		}\hspace{-4pt}
		\subfloat[GT]{
			\begin{minipage}[b]{0.15\linewidth}
				\centering
				\includegraphics[width=\linewidth]{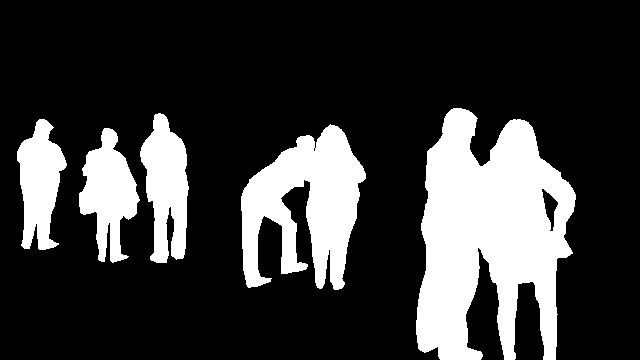}\vspace{1pt}
				\includegraphics[width=\linewidth]{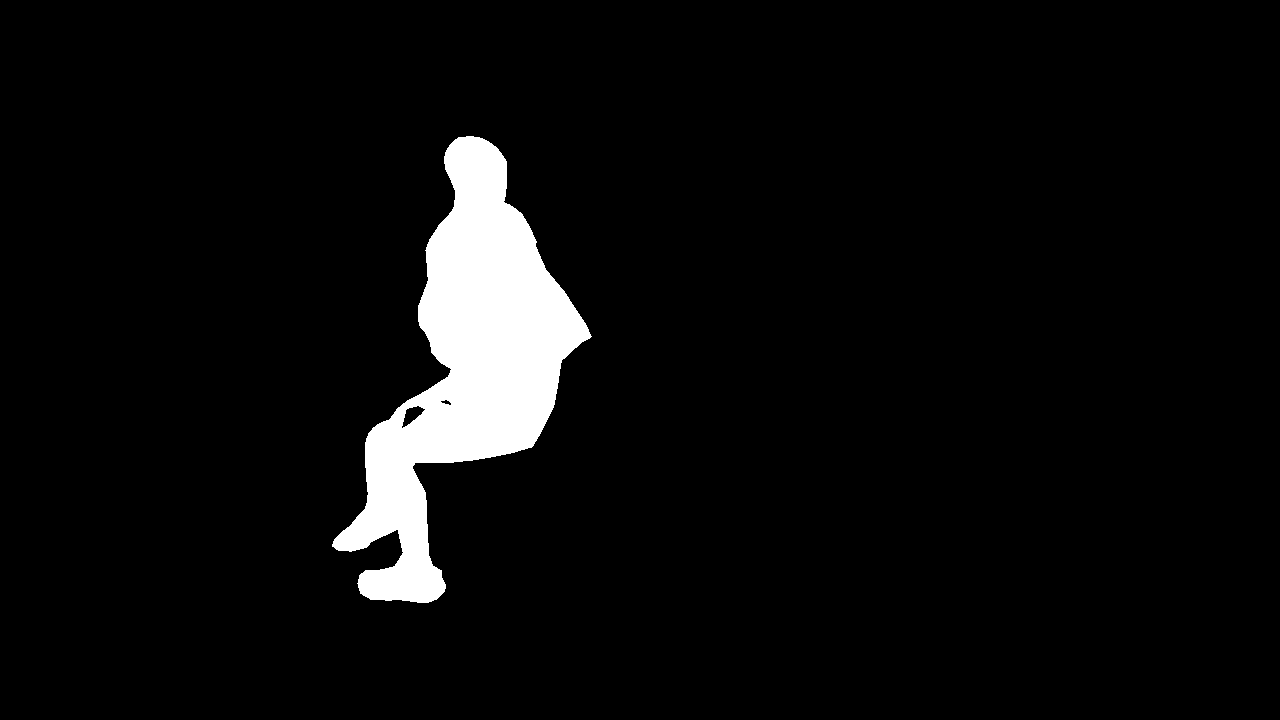}\vspace{1pt}
				\includegraphics[width=\linewidth]{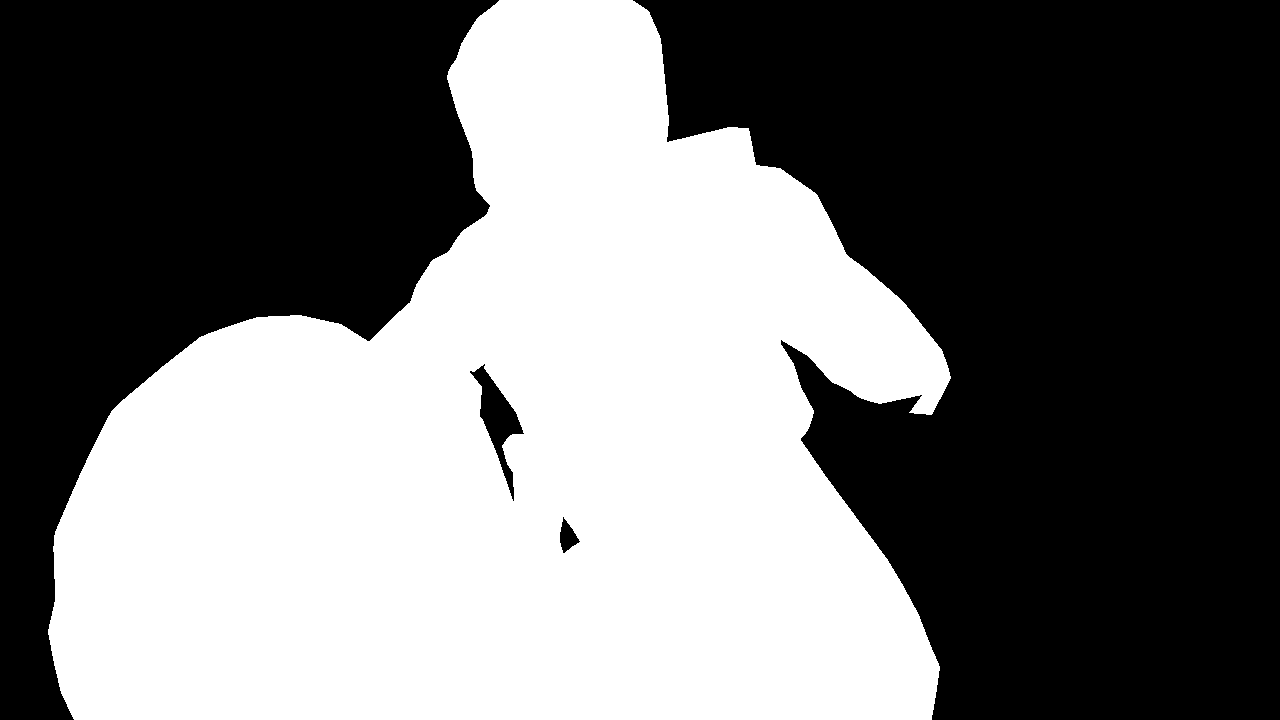}\vspace{1pt}
				\includegraphics[width=\linewidth]{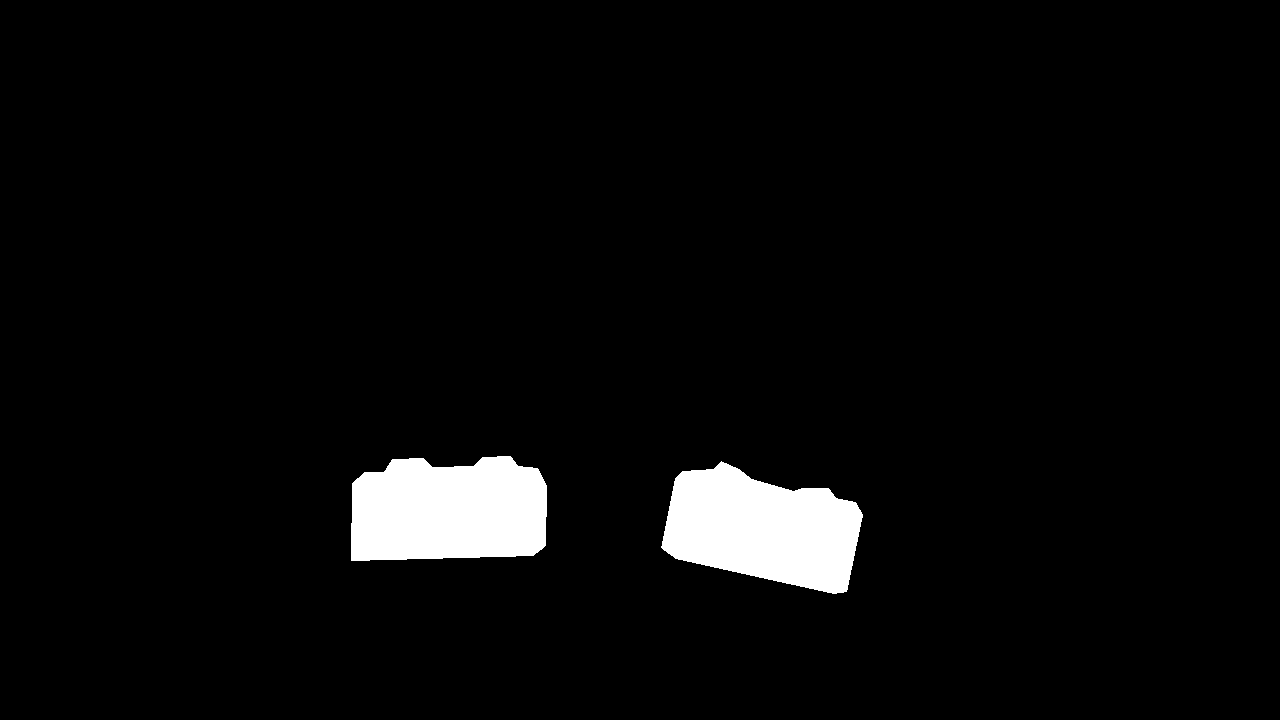}
			\end{minipage}
		}\hspace{-4pt}
		\subfloat[Base]{
			\begin{minipage}[b]{0.15\linewidth}
				\centering
				\includegraphics[width=\linewidth]{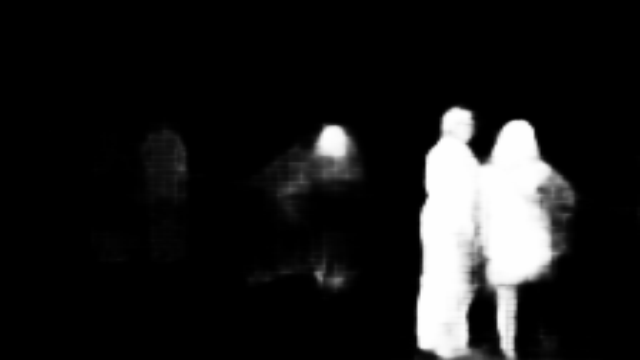}\vspace{1pt}
				\includegraphics[width=\linewidth]{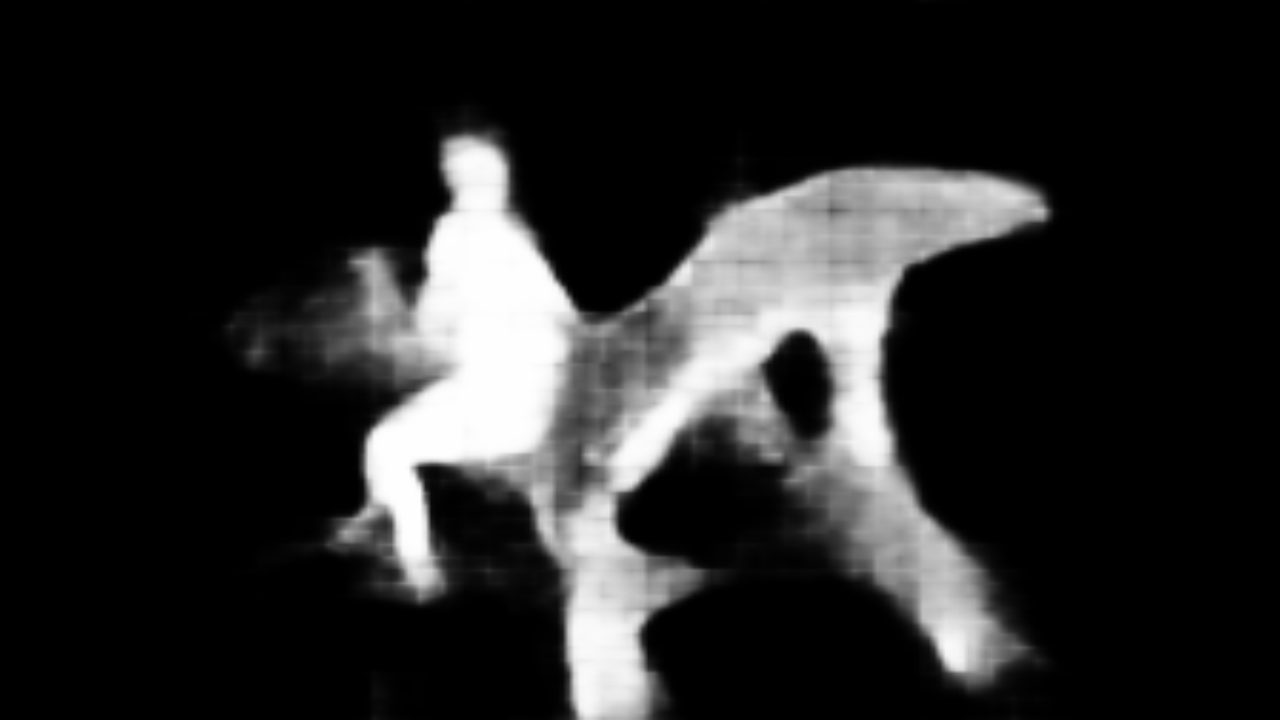}\vspace{1pt}
				\includegraphics[width=\linewidth]{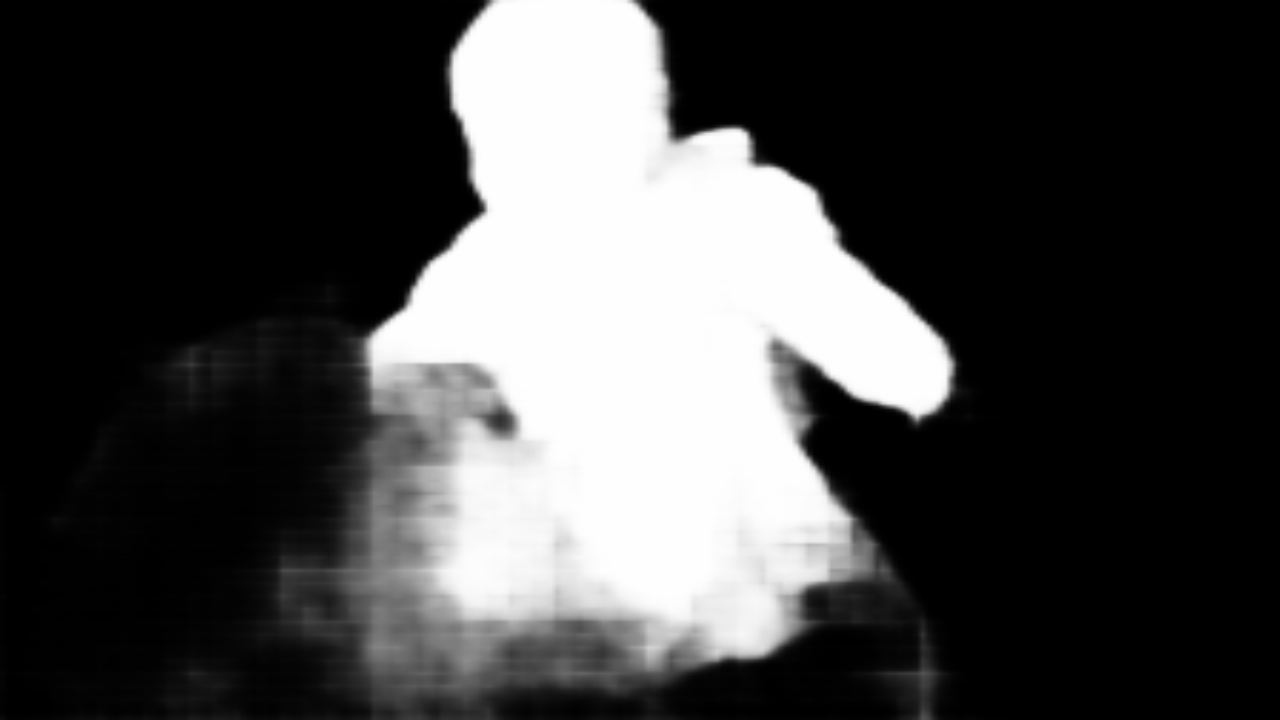}\vspace{1pt}
				\includegraphics[width=\linewidth]{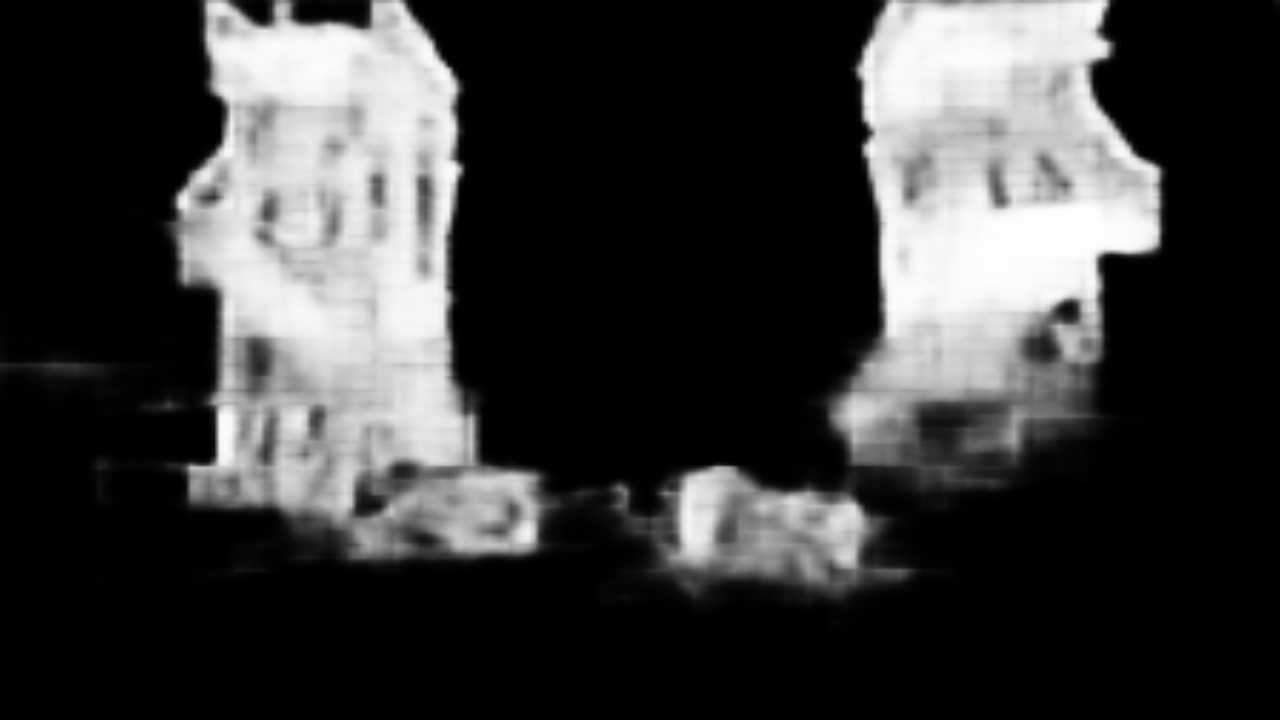}
				
			\end{minipage}
		}\hspace{-4pt}
		\subfloat[+GSA]{
			\begin{minipage}[b]{0.15\linewidth}
				\centering
				\includegraphics[width=\linewidth]{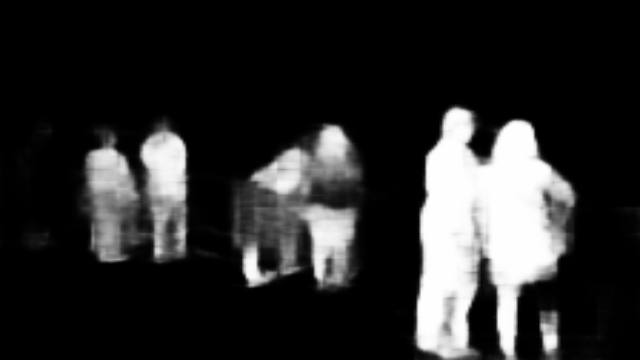}\vspace{1pt}
				\includegraphics[width=\linewidth]{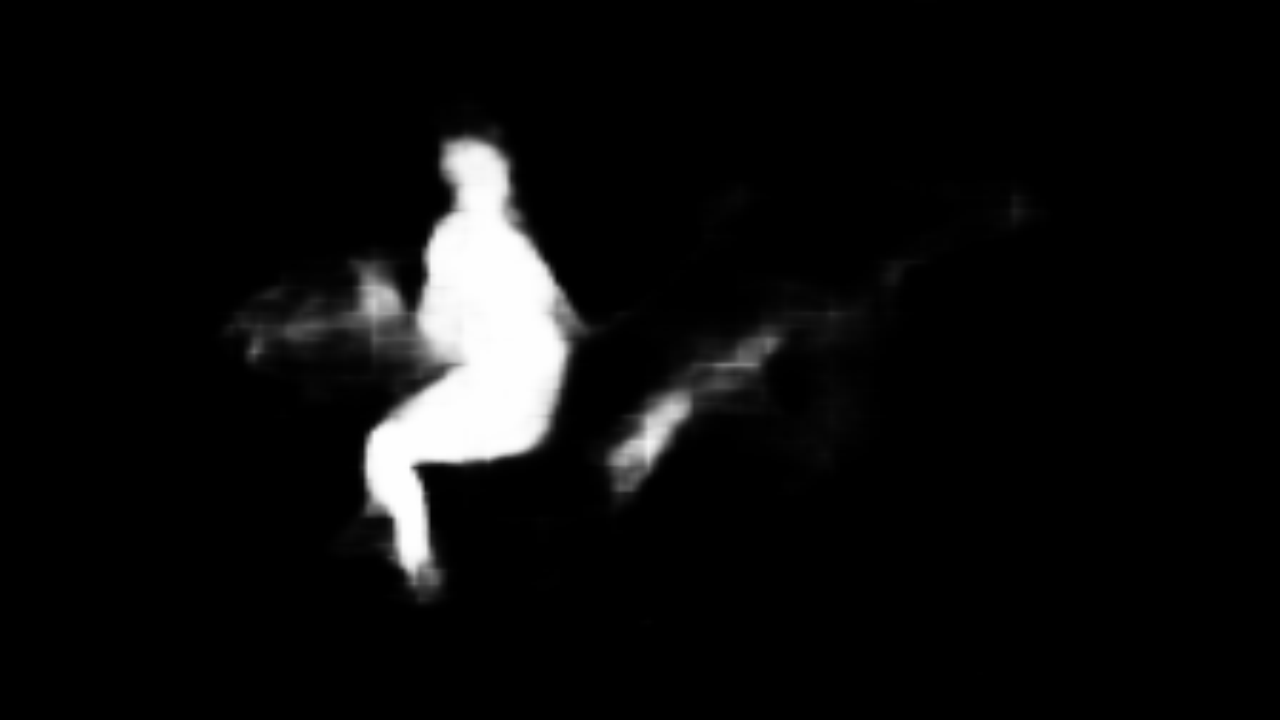}\vspace{1pt}
				\includegraphics[width=\linewidth]{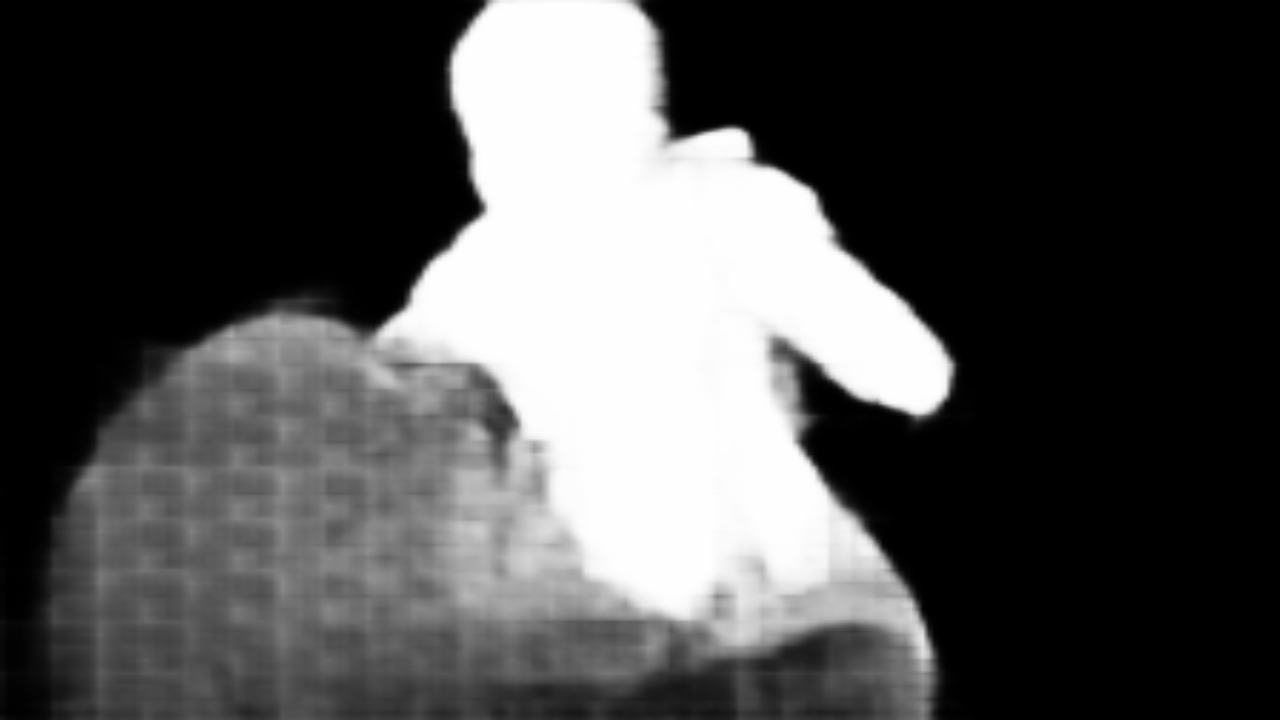}\vspace{1pt}
				\includegraphics[width=\linewidth]{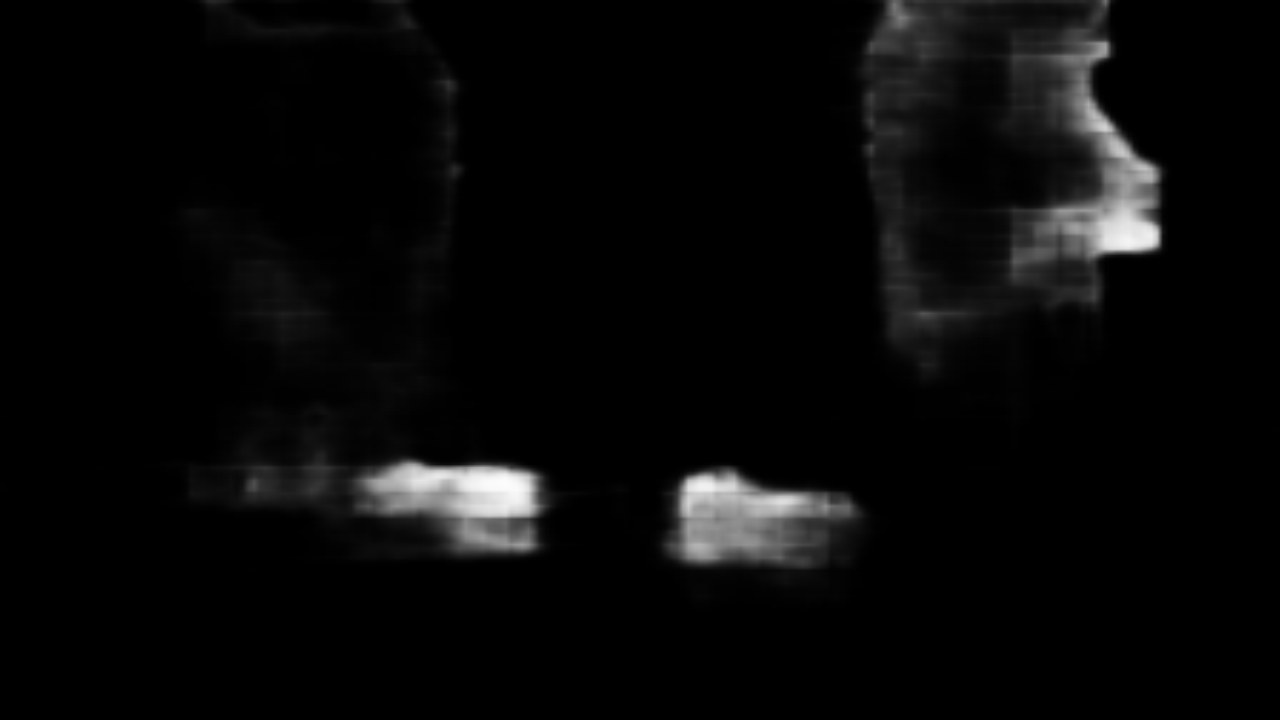}
			\end{minipage}
		}\hspace{-4pt}
		\subfloat[+HCA]{
			\begin{minipage}[b]{0.15\linewidth}
				\centering
				\includegraphics[width=\linewidth]{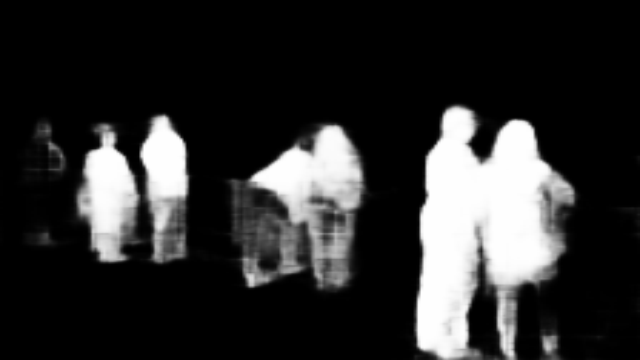}\vspace{1pt}
				\includegraphics[width=\linewidth]{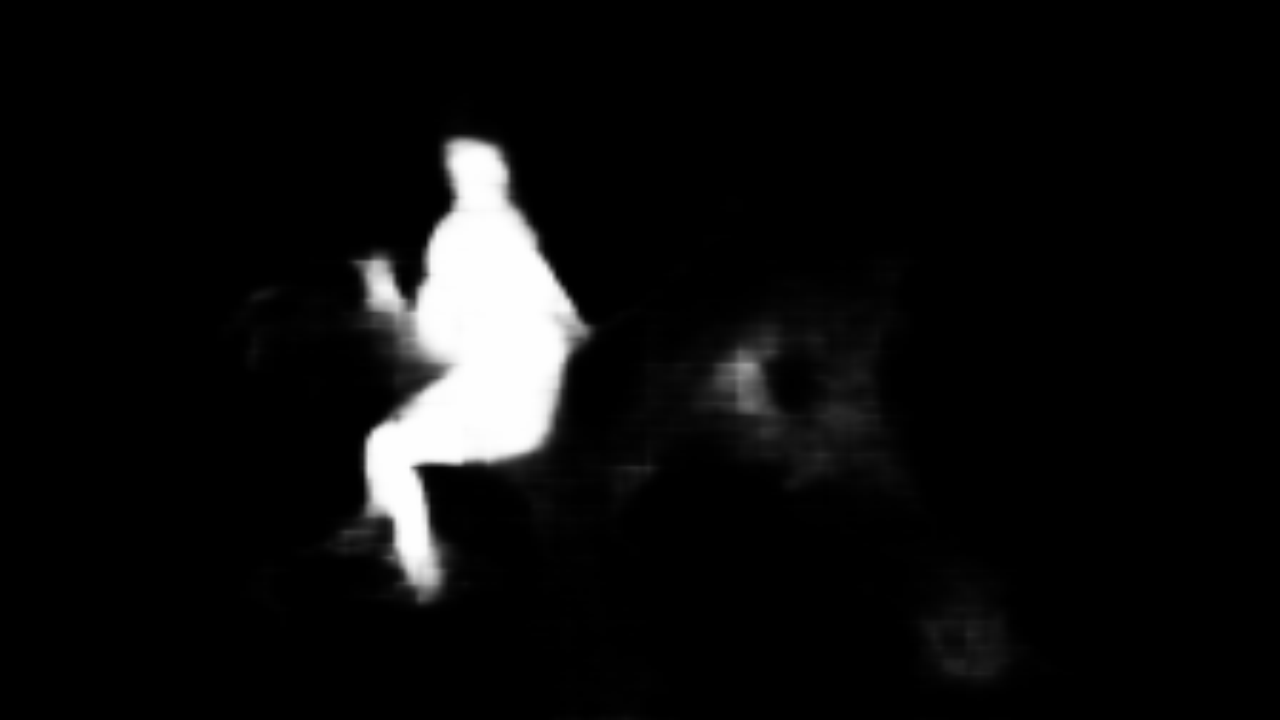}\vspace{1pt}
				\includegraphics[width=\linewidth]{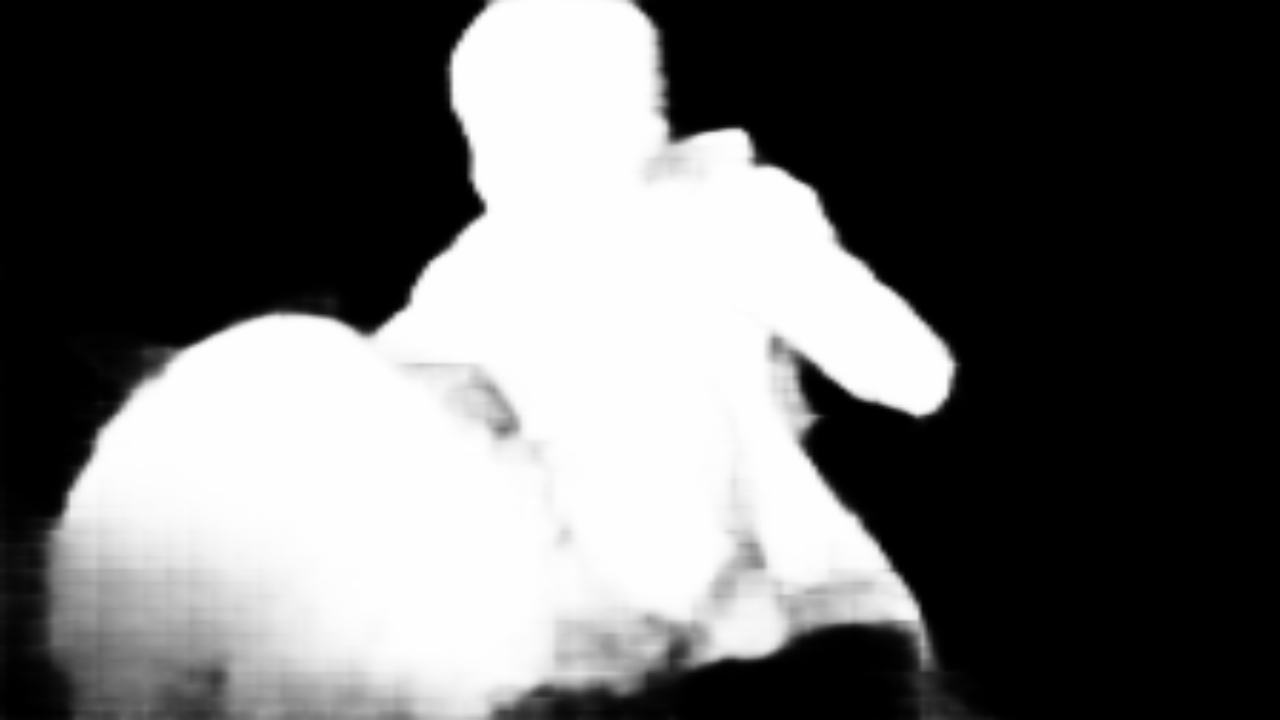}\vspace{1pt}
				\includegraphics[width=\linewidth]{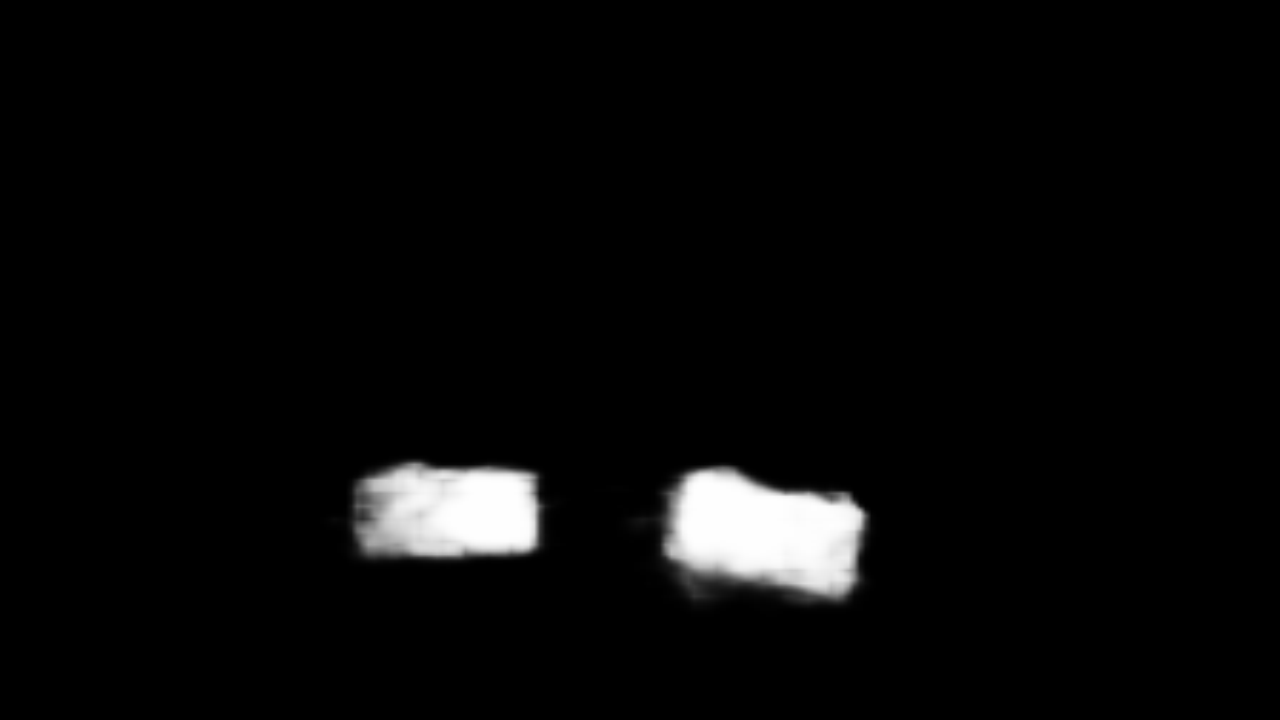}
			\end{minipage}
		}
	\end{minipage}
	\vfill
	\caption{Visualization to show the effectiveness of each component in HCA. "+GSA" means additionally adding the global self-attention shown in Fig. 3(b). "+HCA" denotes adding the whole hierarchical cross-modal attention block shown in Fig. 3(a).}
	\label{fig:HCA}
\end{figure}

\begin{figure}[!ht] 
	\centering 
        \captionsetup[subfloat]{labelsep=none,format=plain,labelformat=empty}
	\begin{minipage}[b]{0.98\linewidth} 
		\subfloat[RGB]{
			\begin{minipage}[b]{0.184\linewidth} 
				\centering
				\includegraphics[width=\linewidth]{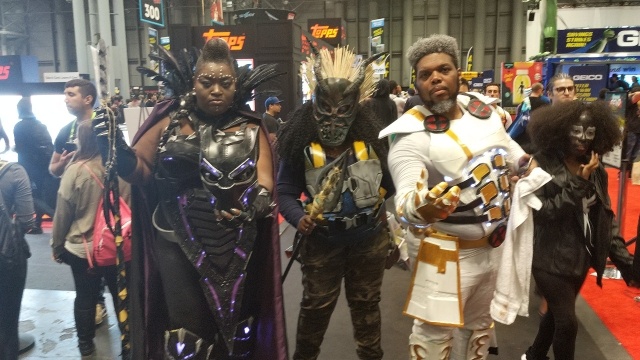}\vspace{1pt}
				\includegraphics[width=\linewidth]{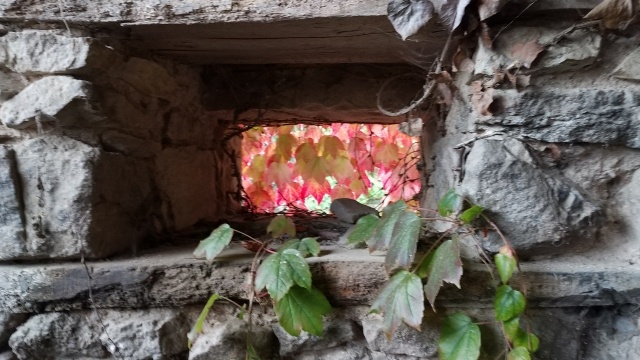}\vspace{1pt}
				\includegraphics[width=\linewidth]{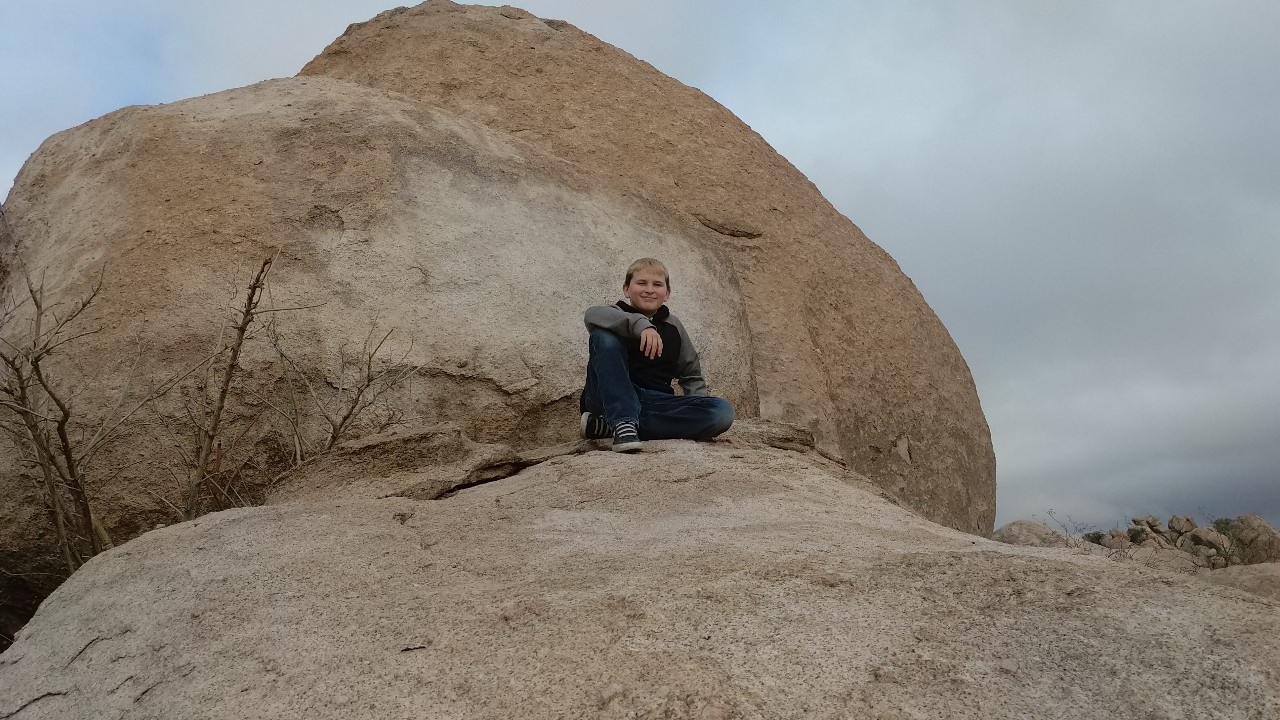}\vspace{1pt}
				\includegraphics[width=\linewidth]{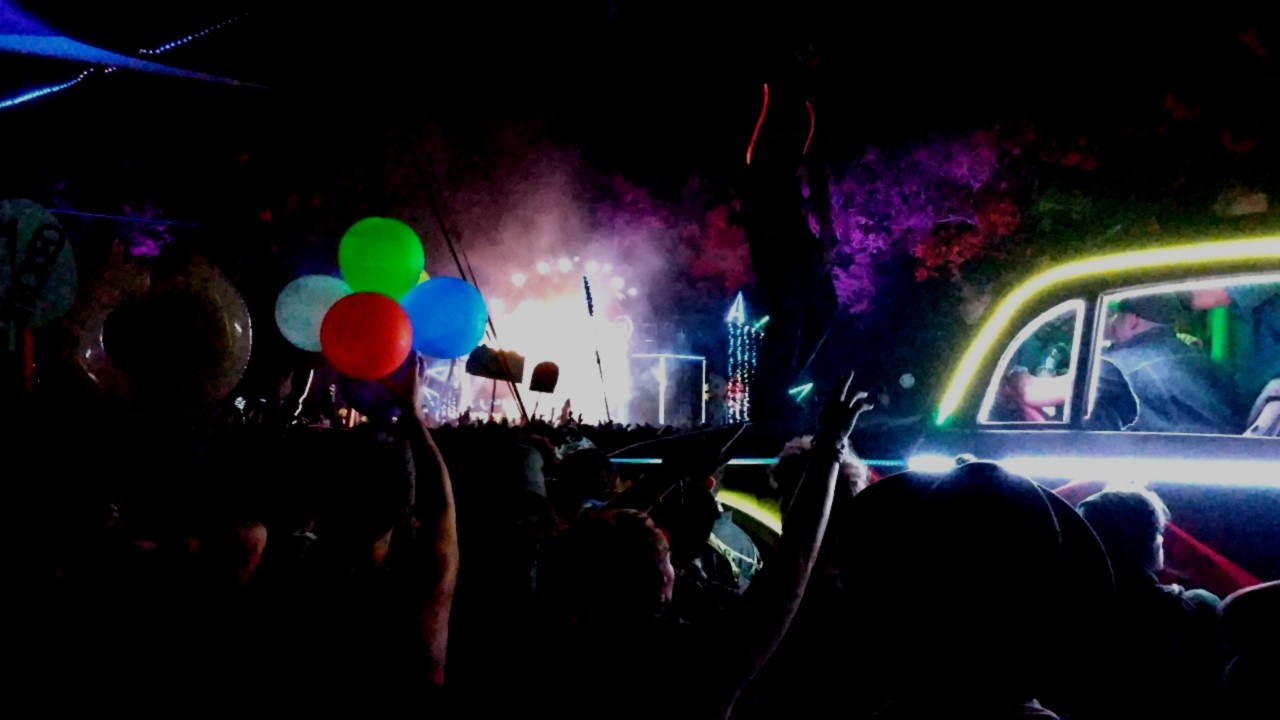}
				
			\end{minipage}
		}\hspace{-5pt}
		\hfill
		\subfloat[Depth]{
			\begin{minipage}[b]{0.184\linewidth}
				\centering
				\includegraphics[width=\linewidth]{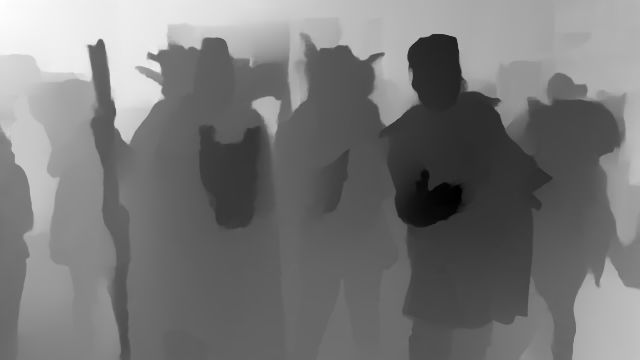}\vspace{1pt}
				\includegraphics[width=\linewidth]{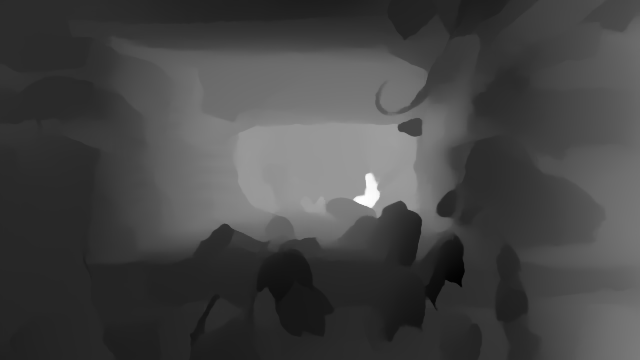}\vspace{1pt}
				\includegraphics[width=\linewidth]{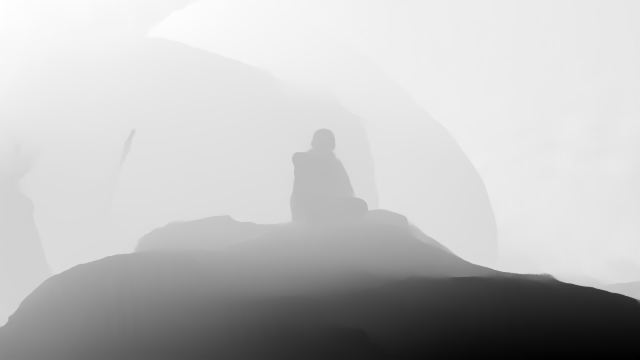}\vspace{1pt}
				\includegraphics[width=\linewidth]{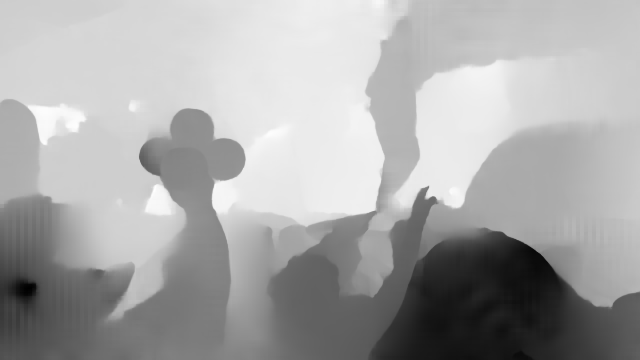}
				
			\end{minipage}
		}\hspace{-5pt}
		\hfill
		\subfloat[GT]{
			\begin{minipage}[b]{0.184\linewidth}
				\centering
				\includegraphics[width=\linewidth]{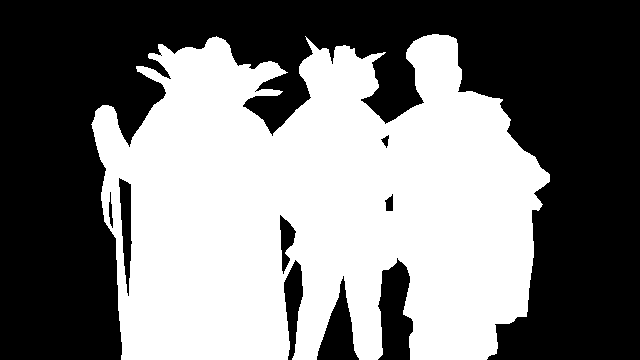}\vspace{1pt}
				\includegraphics[width=\linewidth]{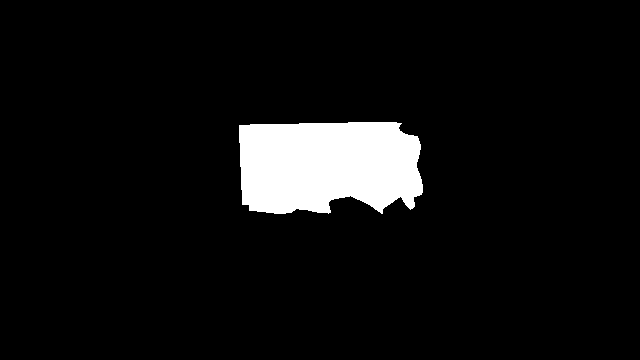}\vspace{1pt}
				\includegraphics[width=\linewidth]{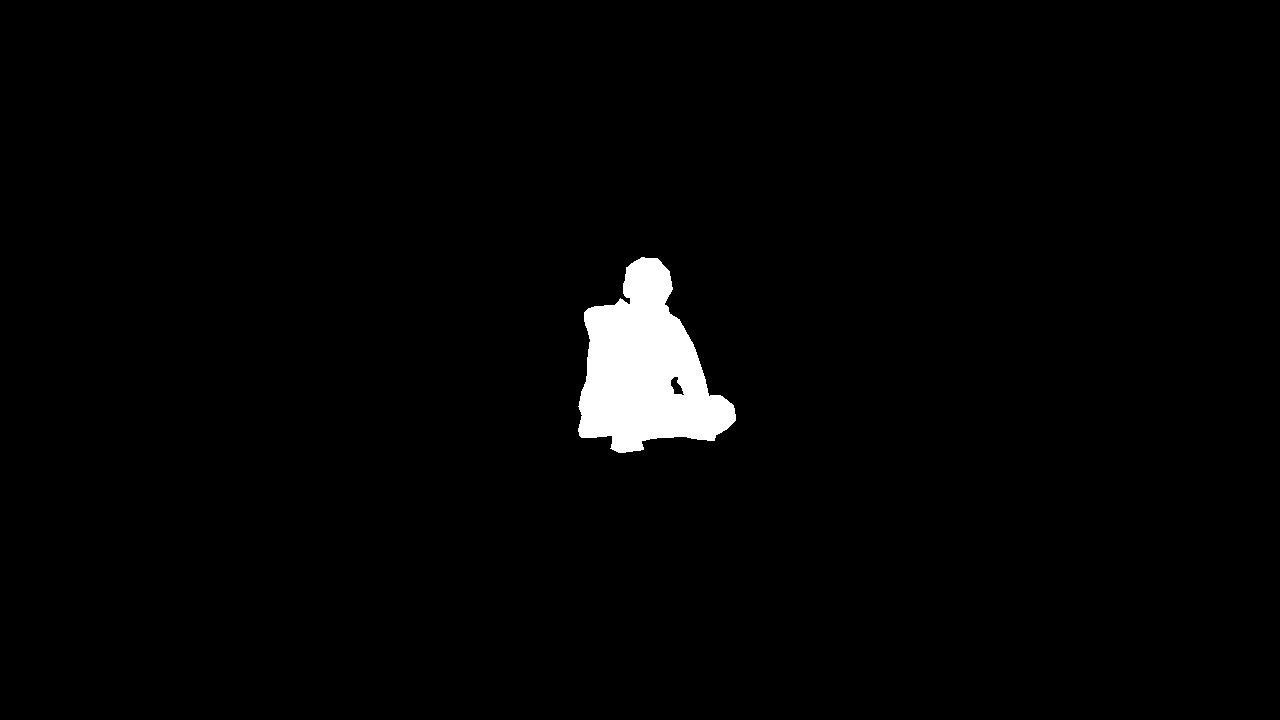}\vspace{1pt}
				\includegraphics[width=\linewidth]{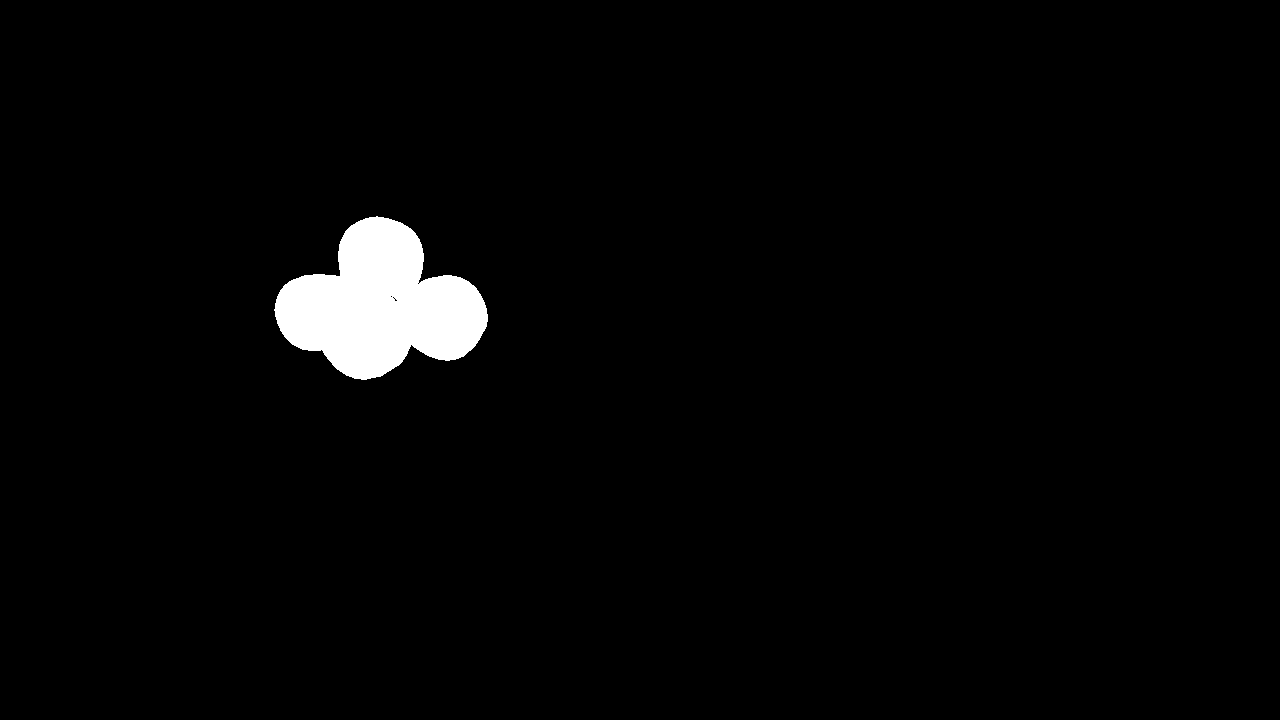}
			\end{minipage}
		}\hspace{-5pt}
		\hfill
		\subfloat[Base]{
			\begin{minipage}[b]{0.184\linewidth}
				\centering
				\includegraphics[width=\linewidth]{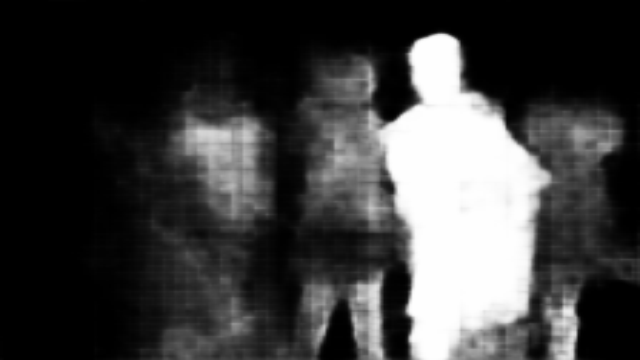}\vspace{1pt}
				\includegraphics[width=\linewidth]{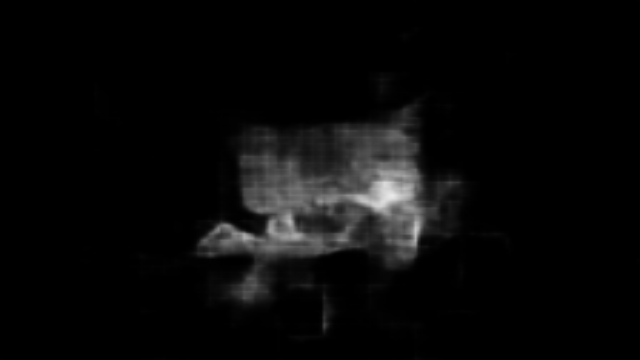}\vspace{1pt}
				\includegraphics[width=\linewidth]{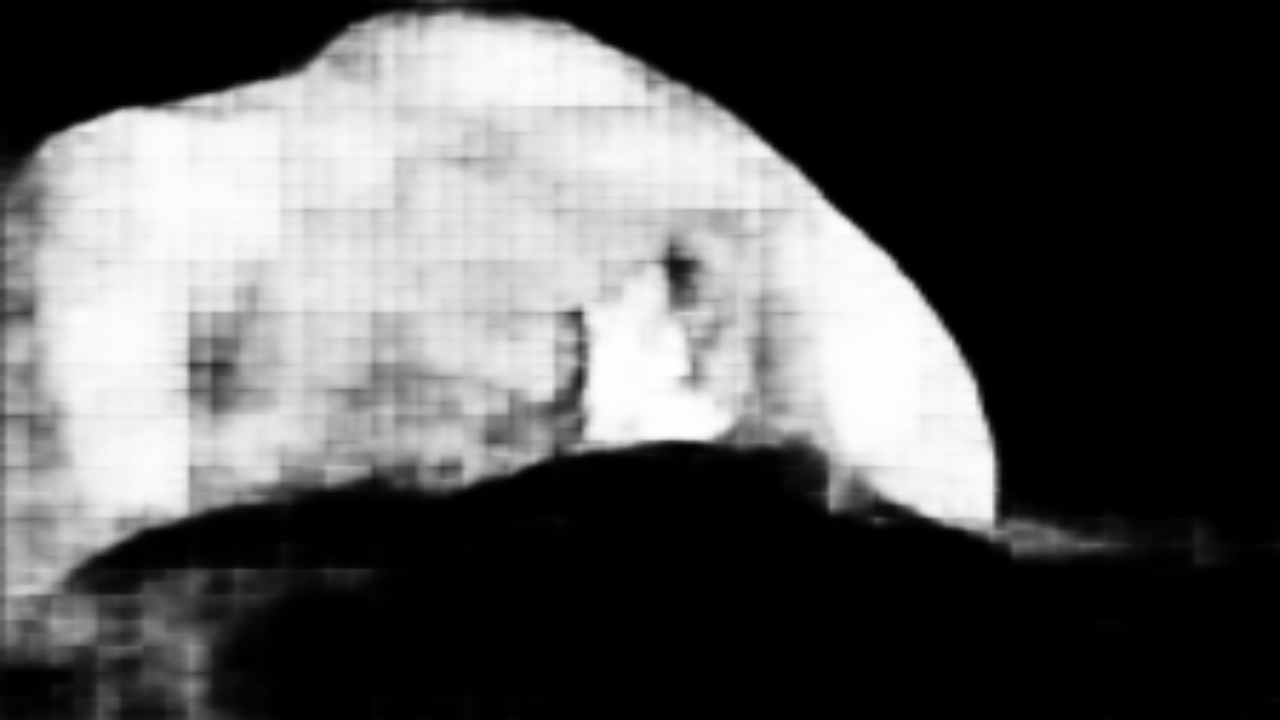}\vspace{1pt}
				\includegraphics[width=\linewidth]{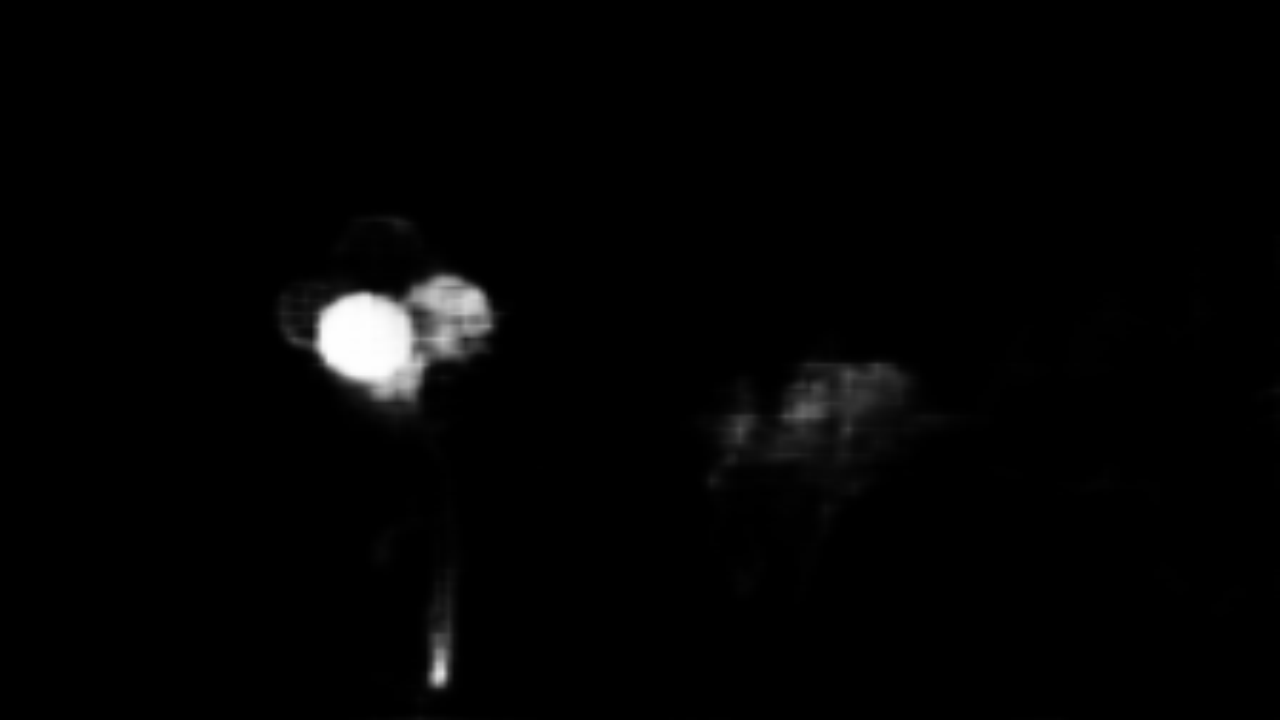}
				
			\end{minipage}
		}\hspace{-5pt}
		\hfill
		\subfloat[+FPT]{
			\begin{minipage}[b]{0.184\linewidth}
				\centering
				\includegraphics[width=\linewidth]{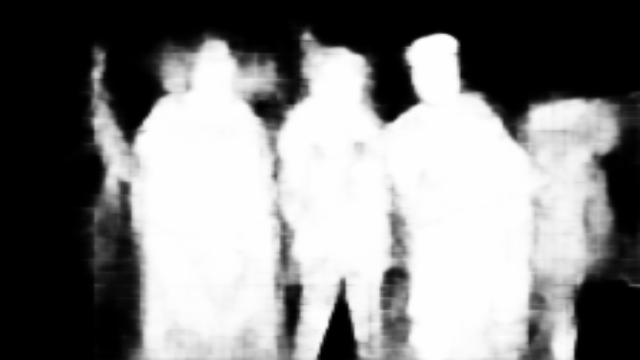}\vspace{1pt}
				\includegraphics[width=\linewidth]{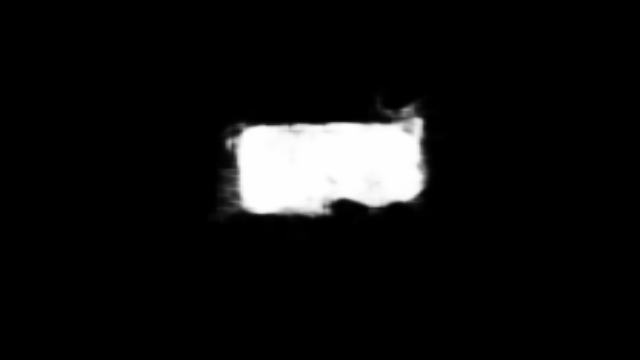}\vspace{1pt}
				\includegraphics[width=\linewidth]{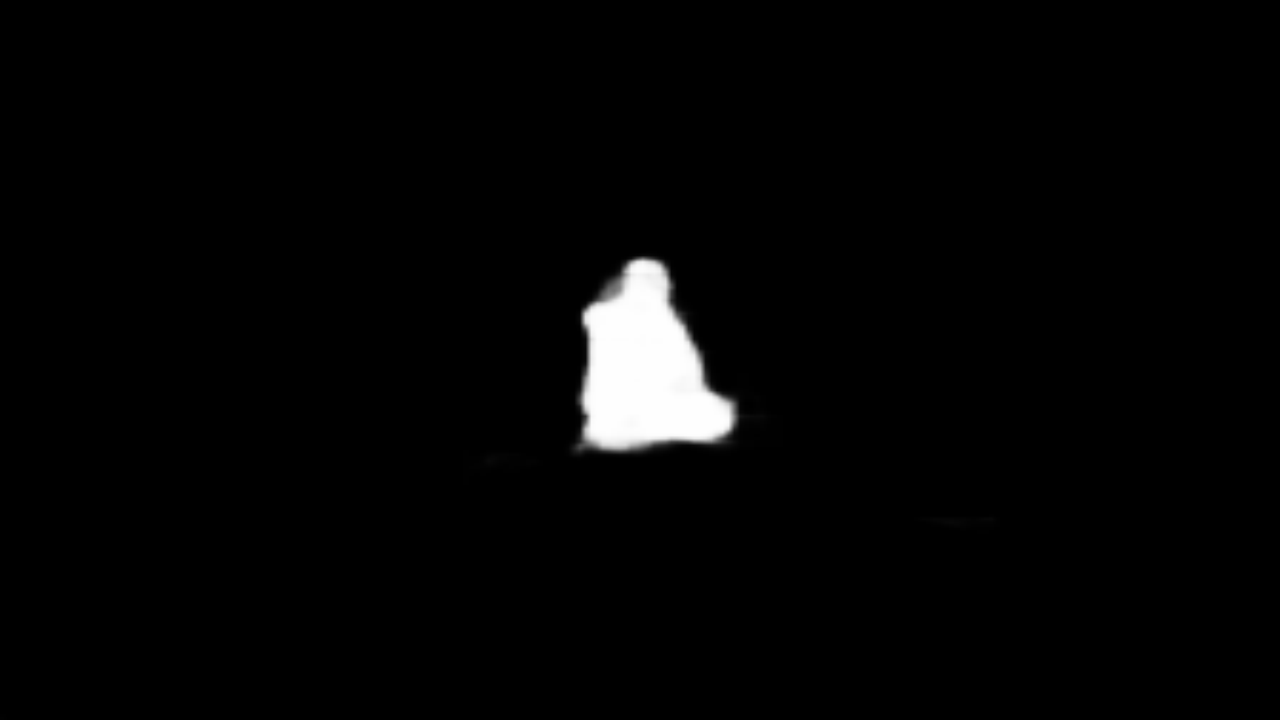}\vspace{1pt}
				\includegraphics[width=\linewidth]{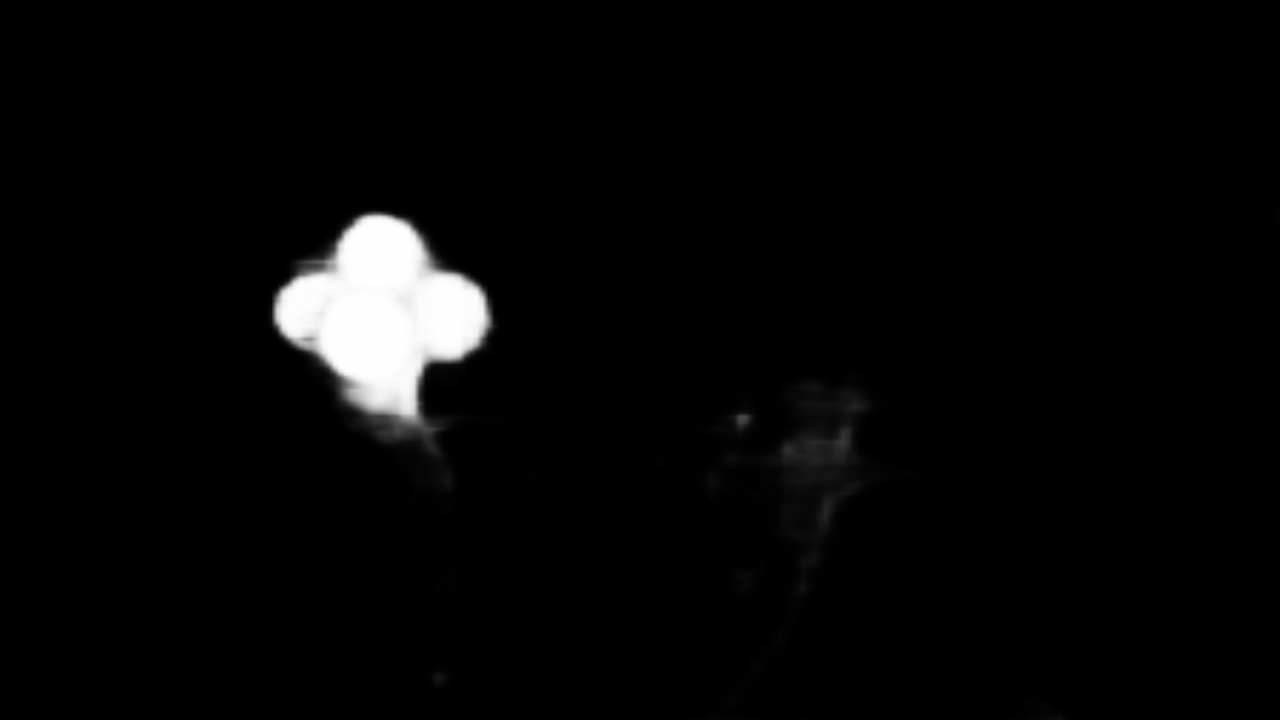}
			\end{minipage}
		}
	\end{minipage}
	\vfill
	\caption{Visualization of the effectiveness of FPT.}
	\label{fig:FPT}
\end{figure}

\par Fig. \ref{fig:HCA} shows the effectiveness of the global cross-attention (GSA) and local-aligned cross attention (LCA) components in HCA visually. As we predict, the global cross-attention in VST (see "Base" in Fig. \ref{fig:HCA}) will introduce noises from other distant areas to wrongly identify the background and foreground. With global self-attention (denoted by "+GSA") to complement the global contexts, the model has stronger global reasoning ability to localize the salient object. With the local-aligned cross attention additionally (denoted by "+HCA"), the cross-modal fusion process takes advantages of the local cross-modal correlations to remove the noises and enhance the object details.

\textbf{Effectiveness of feature pyramid for transformer.}  The comparison between "+HCA+FPT" and "+HCA" in Tab. \ref{tab:ablation} illustrates the improvement of our FPT design, indicating its contribution to boost the cross-level feature selection.  
The saliency maps in Fig. \ref{fig:FPT} visually shows the benefits of FPT.  In FPT, cross-level features are integrated progressively and selectively with the guidance from deep to shallow, thus enabling informative and adaptive multi-level combination. As a result, our model can better infer the location of the salient objects and highlight their boundaries precisely.

\begin{figure}[!ht] 
	\centering 
        \captionsetup[subfloat]{labelsep=none,format=plain,labelformat=empty}
	\begin{minipage}[b]{0.98\linewidth} 
		\subfloat[RGB]{
			\begin{minipage}[b]{0.184\linewidth} 
				\centering
				\includegraphics[width=\linewidth]{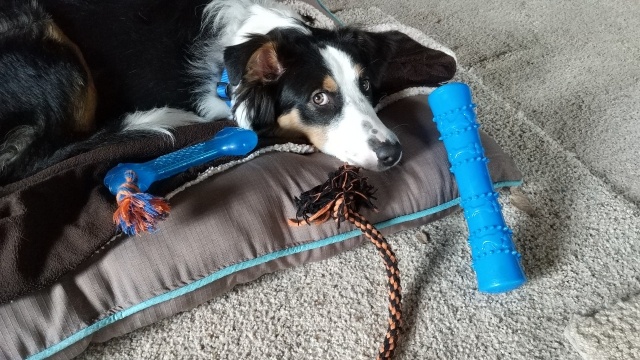}\vspace{1pt}
				\includegraphics[width=\linewidth]{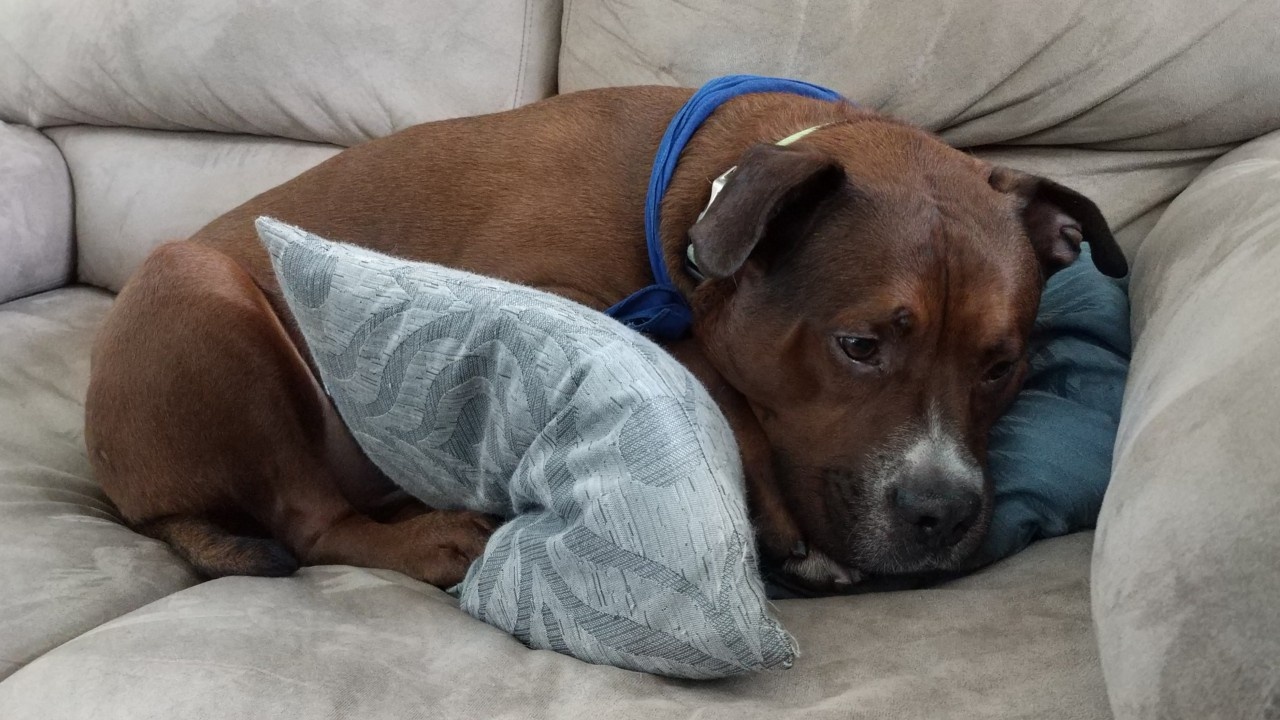}\vspace{1pt}
				\includegraphics[width=\linewidth]{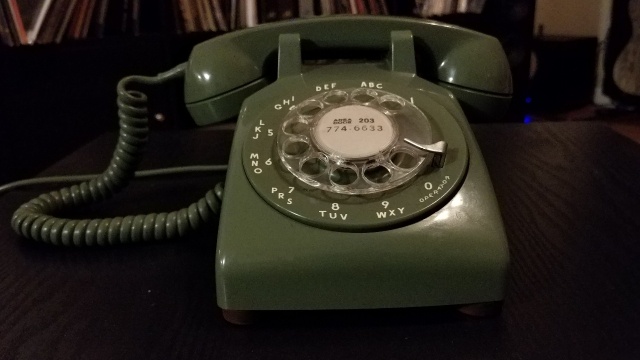}\vspace{1pt}
				\includegraphics[width=\linewidth]{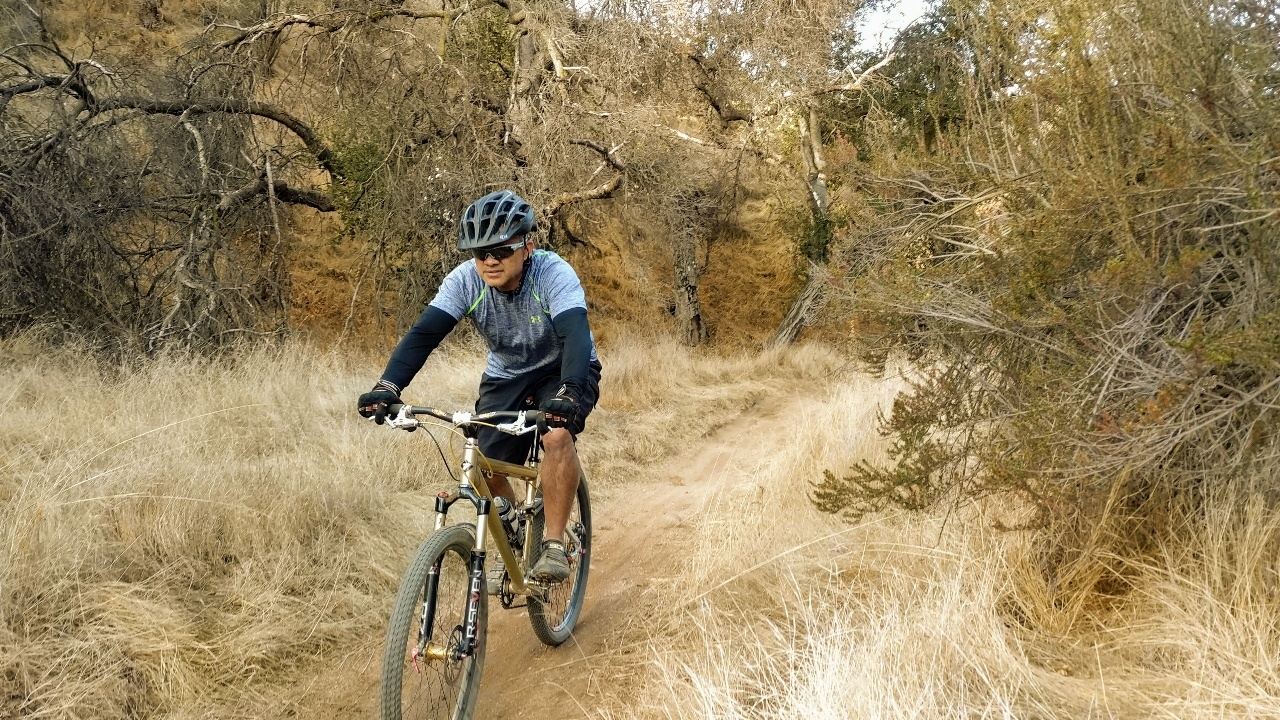}

			\end{minipage}
		}\hspace{-5pt}
		\hfill
		\subfloat[Depth]{
			\begin{minipage}[b]{0.184\linewidth}
				\centering
				\includegraphics[width=\linewidth]{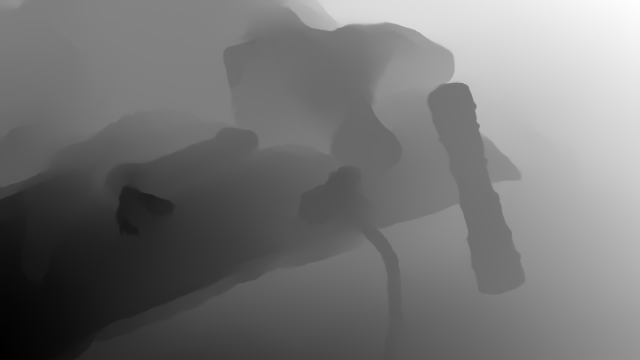}\vspace{1pt}
				\includegraphics[width=\linewidth]{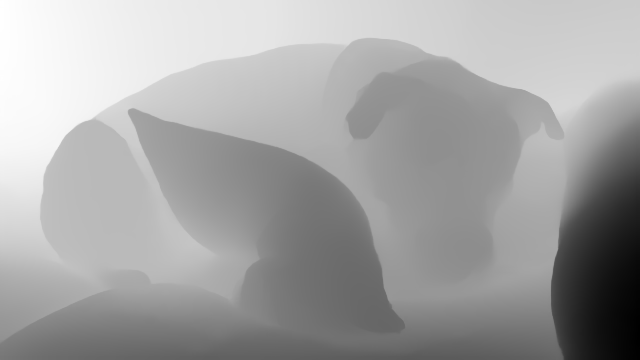}\vspace{1pt}
				\includegraphics[width=\linewidth]{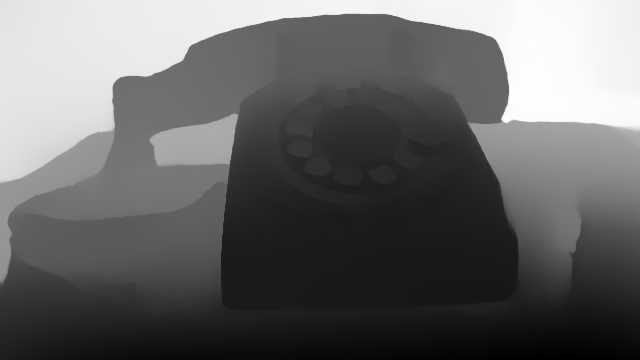}\vspace{1pt}
				\includegraphics[width=\linewidth]{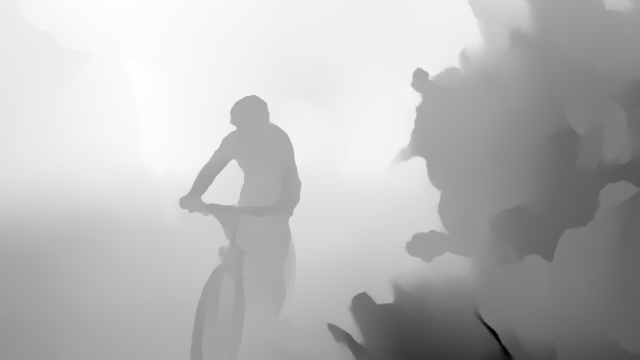}

			\end{minipage}
		}\hspace{-5pt}
		\hfill
		\subfloat[GT]{
			\begin{minipage}[b]{0.184\linewidth}
				\centering
				\includegraphics[width=\linewidth]{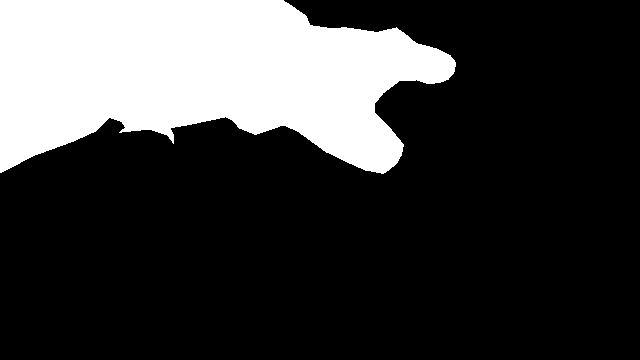}\vspace{1pt}
				\includegraphics[width=\linewidth]{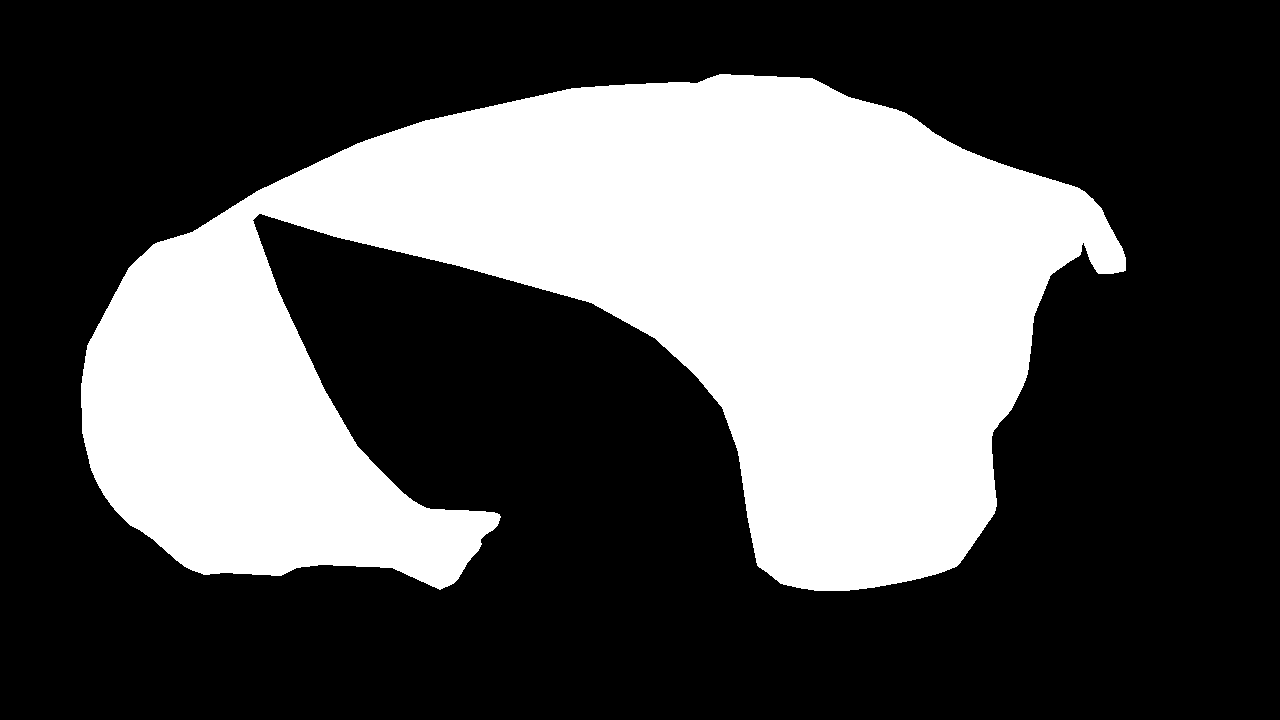}\vspace{1pt}
				\includegraphics[width=\linewidth]{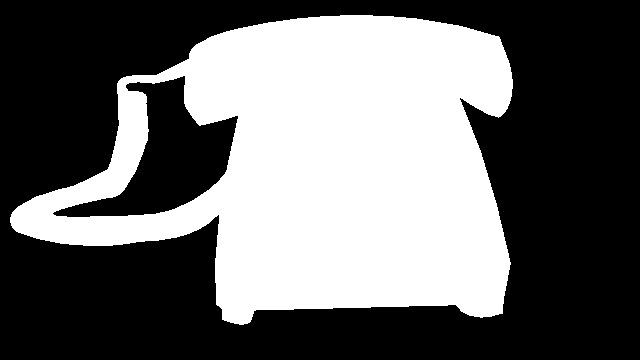}\vspace{1pt}
				\includegraphics[width=\linewidth]{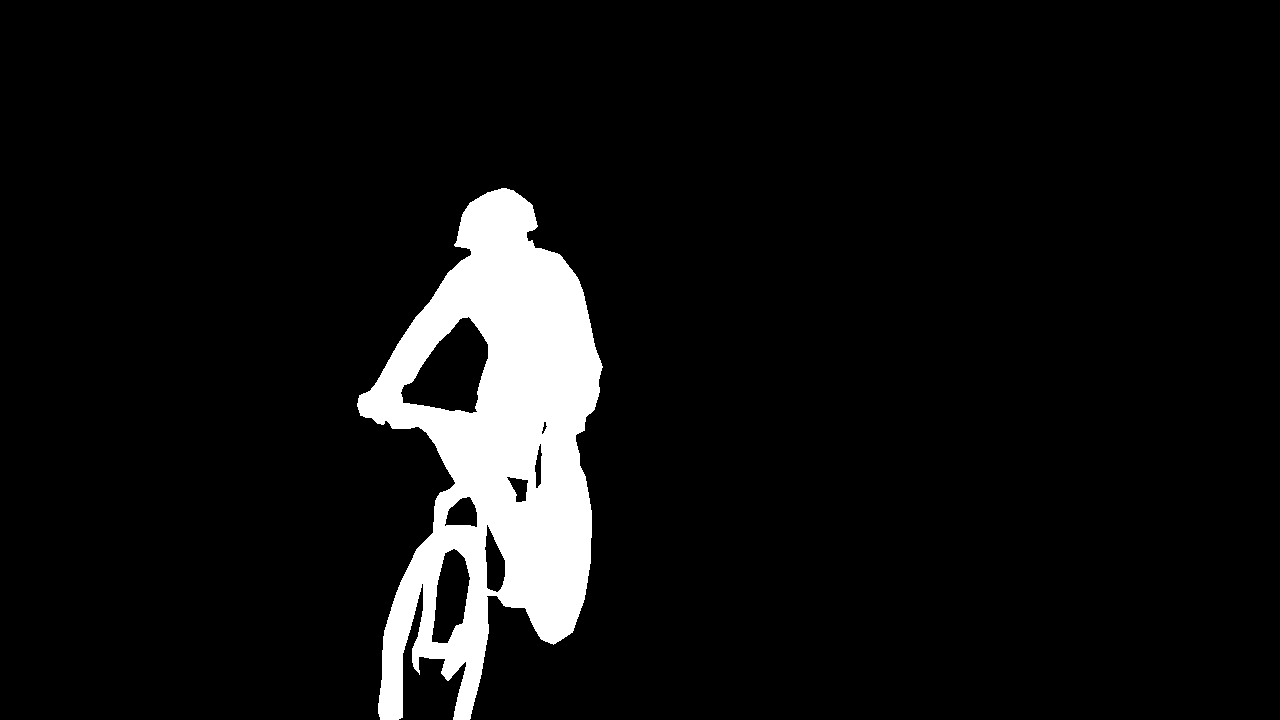}
			\end{minipage}
		}\hspace{-5pt}
		\hfill
		\subfloat[Concat]{
			\begin{minipage}[b]{0.184\linewidth}
				\centering
				\includegraphics[width=\linewidth]{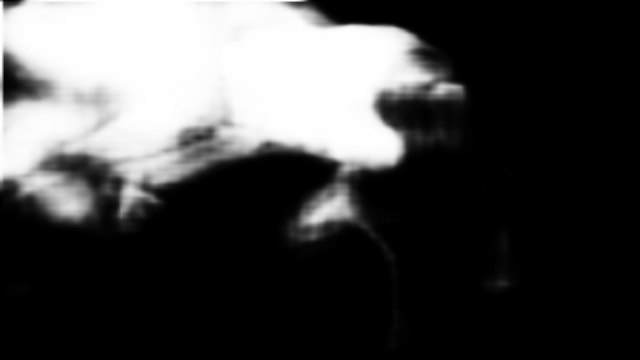}\vspace{1pt}
				\includegraphics[width=\linewidth]{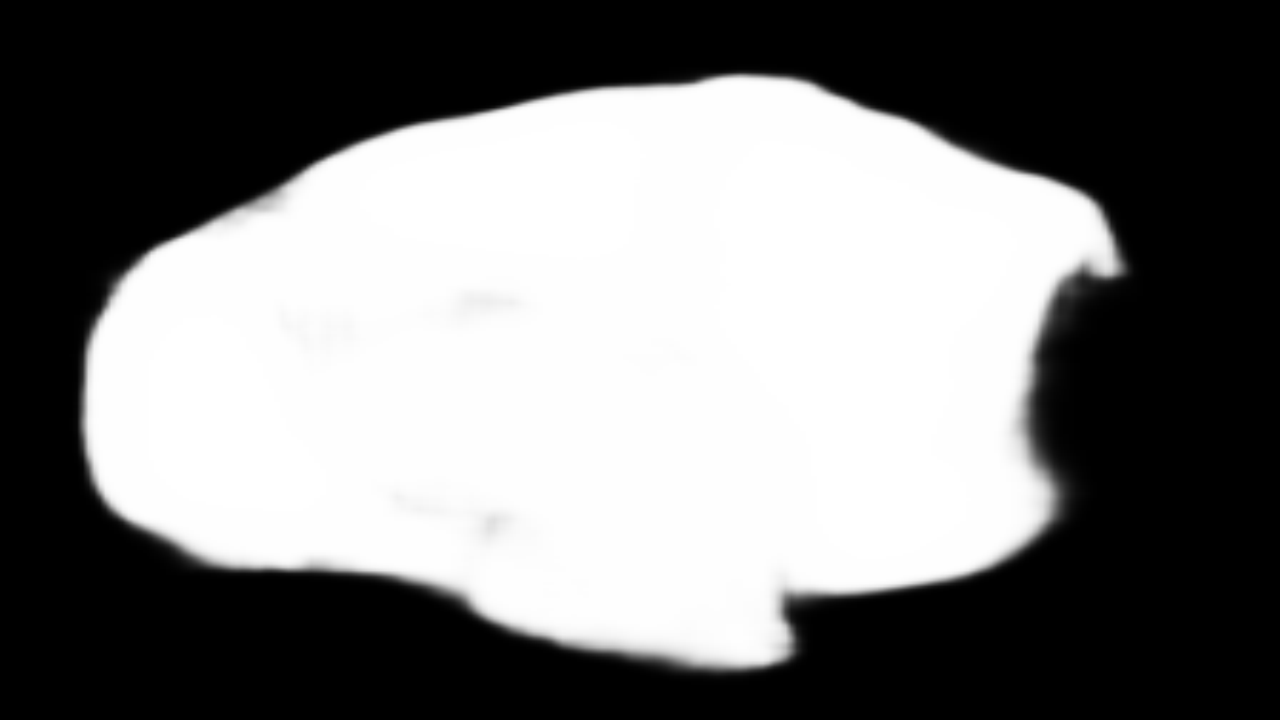}\vspace{1pt}
				\includegraphics[width=\linewidth]{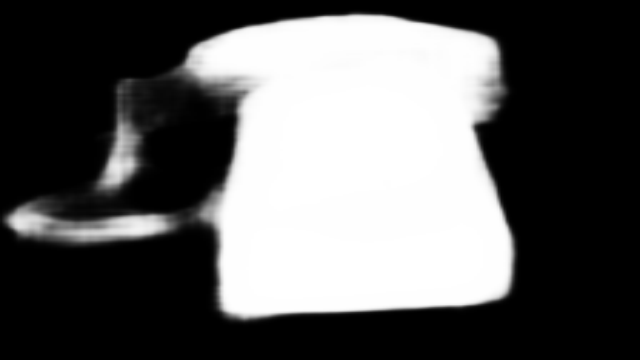}\vspace{1pt}
				\includegraphics[width=\linewidth]{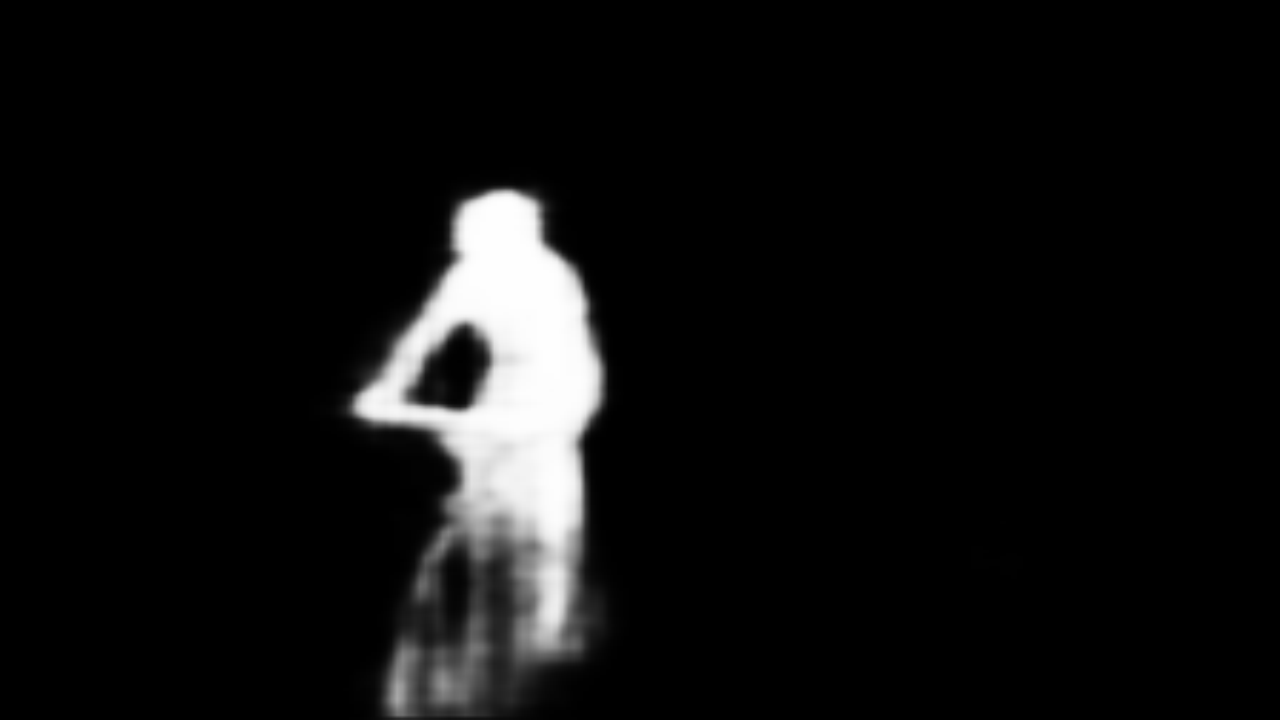}

			\end{minipage}
		}\hspace{-5pt}
		\hfill
		\subfloat[+DCM]{
			\begin{minipage}[b]{0.184\linewidth}
				\centering
				\includegraphics[width=\linewidth]{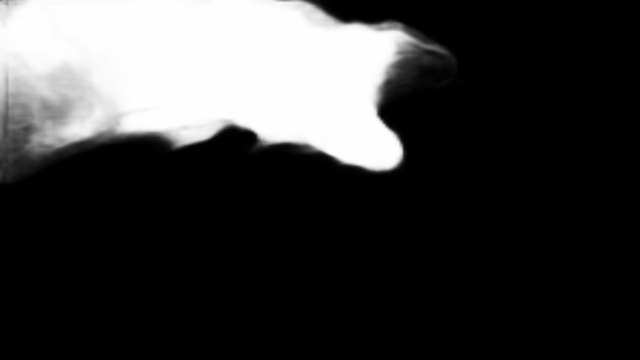}\vspace{1pt}
				\includegraphics[width=\linewidth]{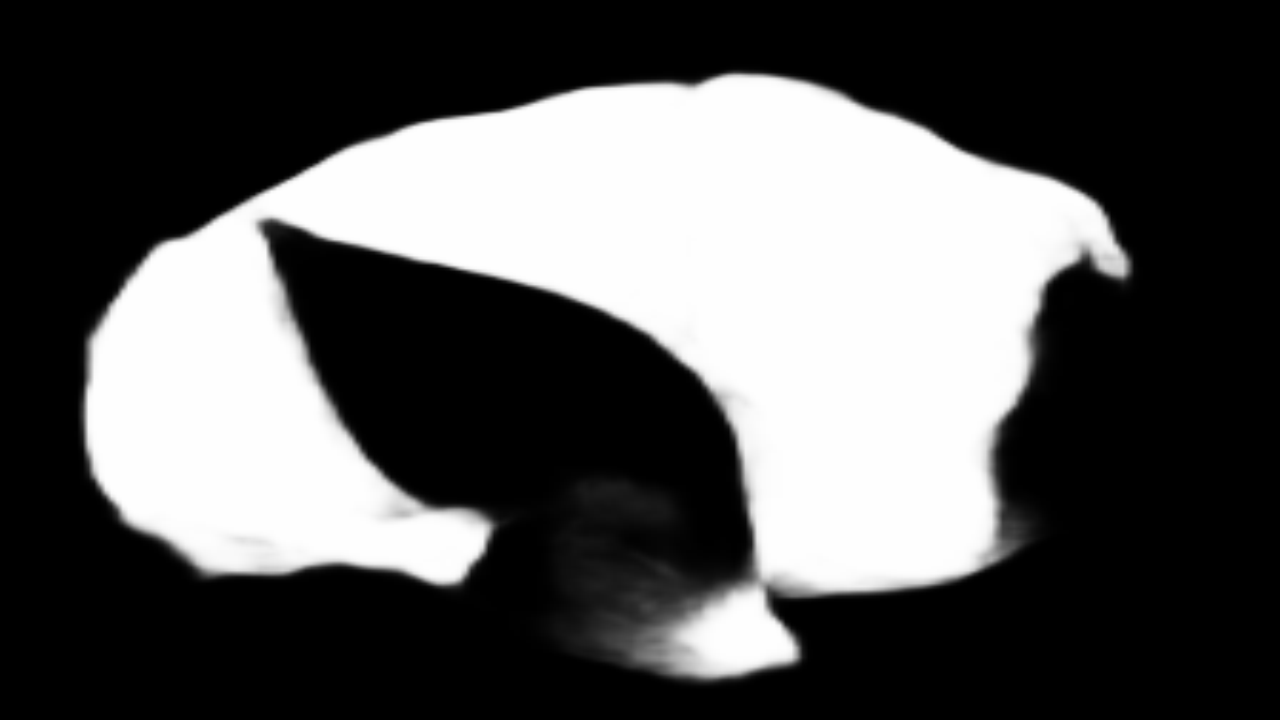}\vspace{1pt}
				\includegraphics[width=\linewidth]{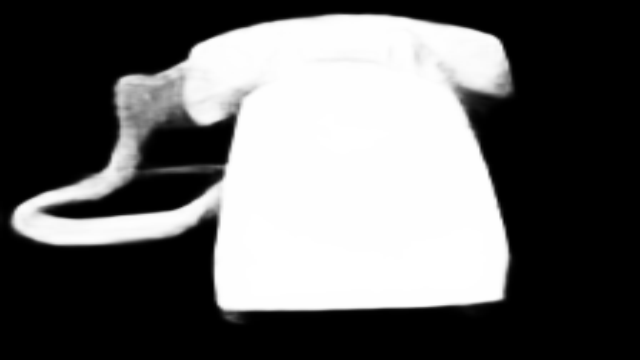}\vspace{1pt}
				\includegraphics[width=\linewidth]{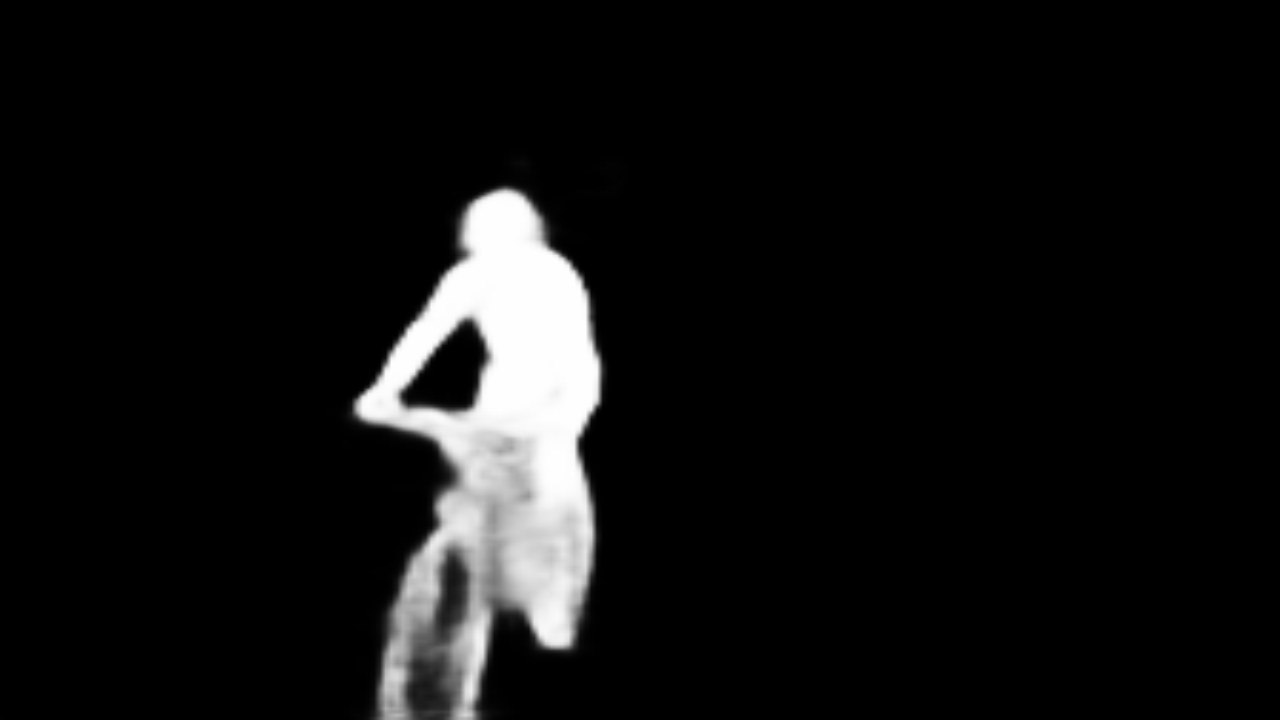}
			\end{minipage}
		}
	\end{minipage}
	\vfill
	\caption{Visualization of the effectiveness of DCM. Concat means feature concatenation.}
	\label{fig:CCM}
\end{figure}

\textbf{Effectiveness of the disentangled complementing module.} 
The comparison between line "+HCA+FPT" and "+HCA+FPT+DCM" in Tab. \ref{tab:ablation} shows the considerable improvement of DCM, indicating that diversifying the complex cross-modal complements into consistent and complementary ones will boost the fusion adaptivity.

The visualization results are shown in Fig. \ref{fig:CCM}. Undifferentiated concatenation is too ambiguous to explore the detailed consistent information between modalities, thus the predicted maps are bad in details, especially around the object boundaries (see the $1^{st}$, $3^{rd}$ and $4^{th}$ rows). Instead, DCM explicitly decouples the difference and consistency between modalities and combine them adaptively for varying scenes to localize the salient objects accurately and highlight salient regions uniformly.

\section{Conclusion}
In this paper, we propose a new transformer-based architecture for RGB-D salient object detection. To tackle the modality gap and spatial discrepancy when combining cross-modal transformer features, we tailor a hierarchical cross-modal attention to explore the cross-modal complements in terms of the global contexts and local correlations successively. Also, a feature pyramid for transformer is designed to achieve adaptive cross-level feature selection, and a disentangled complementing module is introduced to disentangle complex complements into consistent and complementary ones to boost the cross-modal fusion adaptivity. Extensive experiments verify the advantages of our multi-modal transformer and the efficacy of our designs.

\newpage

\nocite{*}
\bibliographystyle{gbt7714-numerical}
\bibliography{aaai22.bib}

\end{document}